\documentclass[10pt,twocolumn,letterpaper]{article}

\usepackage{wacv}
\usepackage{times}
\usepackage{epsfig}
\usepackage{graphicx}
\usepackage{amsmath}
\usepackage{amssymb}

\makeatletter
\@namedef{ver@everyshi.sty}{}
\makeatother
\usepackage{tikz}
\usepackage{multirow}
\usepackage{booktabs}
\usepackage{colortbl}

\usepackage{pifont}

\newcommand{\skipping}[1]{}

\usepackage[accsupp]{axessibility}  % Improves PDF readability for those with disabilities.

\newcommand*\circled[1]{\tikz[baseline=(char.base)]{\node[shape=circle,draw,inner sep=0.5pt] (char) {#1};}}

\def\wacvPaperID{345} % Enter the WACV Paper ID here

\wacvfinalcopy % *** Uncomment this line for the final submission

\ifwacvfinal
\def\assignedStartPage{9876} % *** Enter the assigned starting page number (instead of 9876)
\fi

\ifwacvfinal
\usepackage[breaklinks=true,bookmarks=false]{hyperref}
\else
\usepackage[pagebackref=true,breaklinks=true,colorlinks,bookmarks=false]{hyperref}
\fi

\ifwacvfinal
\else
\fi

\definecolor{ao}{rgb}{0.0, 0.5, 0.0}
\definecolor{applegreen}{rgb}{0.55, 0.71, 0.0}

\begin{document}

%%%%%%%%% TITLE
\title{M-FUSE: Multi-frame Fusion for Scene Flow Estimation}

\author{Lukas Mehl \hspace{15mm} Azin Jahedi \hspace{15mm} Jenny Schmalfuss \hspace{15mm} Andr\'{e}s Bruhn\\
Institute for Visualization and Interactive Systems, University of Stuttgart\\
{\tt\small \{lukas.mehl,azin.jahedi,jenny.schmalfuss,andres.bruhn\}@vis.uni-stuttgart.de}
% For a paper whose authors are all at the same institution,
% omit the following lines up until the closing ``}''.
% Additional authors and addresses can be added with ``\and'',
% just like the second author.
% To save space, use either the email address or home page, not both
%\and
%Second Author\\
%Institution2\\
%First line of institution2 address\\
%{\tt\small secondauthor@i2.org}
}

\maketitle
%\thispagestyle{empty}

%%%%%%%%% ABSTRACT
\begin{abstract}
Recently, neural network for scene flow estimation show impressive results on automotive data such as the KITTI benchmark. 
However, despite of using sophisticated rigidity assumptions and parametrizations, such networks are typically limited to only two frame pairs which does not allow them to exploit temporal information.
In our paper we address this shortcoming by proposing a novel multi-frame approach that considers an additional preceding stereo pair.
To this end, we proceed in two steps: 
Firstly, building upon the recent RAFT-3D approach, we develop an improved two-frame baseline by incorporating an advanced stereo method.
Secondly, and even more importantly, exploiting the specific modeling concepts of RAFT-3D, we propose a U-Net architecture that performs a fusion of forward and backward flow estimates and hence allows to integrate temporal information on demand. 
Experiments on the KITTI benchmark do not only show that the advantages of the improved baseline
and the temporal fusion approach complement each other, they also demonstrate that the computed scene flow is highly accurate. 
More precisely, our approach ranks second overall and first for the even more challenging foreground objects, in total outperforming the original RAFT-3D method by more than 16\%.
% Code is available at \mbox{\url{https://in/v.final}}.
Code is available at \url{https://github.com/cv-stuttgart/M-FUSE}.%
% Code is available in the supplementary material and will be published online.
\end{abstract}

%%%%%%%%% BODY TEXT

\section{Introduction}

Estimating the 3D motion field of objects in the 3D world
%from image sequences,
from stereo or RGBD image sequences, 
the so-called scene flow, is one of the fundamental tasks in computer vision. Its fields of application range from robotics and automotive scenarios \cite{Menze2015_KITTI} over markerless motion capture for virtual and augmented reality
\cite{Valgaerts2012_PerformanceCapture} %\cite{Richardt2016_3DV} 
to action recognition and intention prediction~\cite{Wang2017_ActionReg}. 

Early works go back to the seminal approach of Vedula \etal \cite{Vedula1999_SceneFlow} in the late nineties and since then variational methods have been among the leading techniques to solve this task; % for more than two decades; 
see \eg \cite{Devernay2007_SceneFlow,Vogel2015_PRSM,Wedel2011_SceneFlow}.
Only recently, four years after 
their first application to scene flow estimation~\cite{Mayer2016_SceneFlow},
% they have been considered first for estimating scene flow~\cite{Mayer2016_SceneFlow}, 
neural networks have been able take the lead in dedicated benchmarks such as KITTI \cite{Menze2015_KITTI}; see \eg the approaches in  \cite{Liu2022_camliflow,Teed2021_RAFT3D,Yang2020_opticalexpansion,Yang2021_SegmentRigid}.
This comparably~late success of neural networks, however,
%-- they have already been proposed in 2016 for scene flow estimation \cite{Mayer2016_SceneFlow} -- 
is not surprising:
Scene flow estimation has more degrees of freedom than other correspondence problems that only work in 2D or 1D, such as optical flow and stereo,
%such as optical flow and stereo (3D motion and potentially depth vs. %2D motion vs. depth), 
hence solving this task requires more sophisticated ideas and more complex network architectures. % to obtain competitive results.

One way to deal with these additional degrees of freedom
% offered by the scene flow problem
is to use semantic information.
This information can be given in terms of object models \cite{Menze2015_KITTI} or instance segmentations \cite{Behl2017_ISF,Ma2019_DRISF,Yang2021_SegmentRigid}.
Another way is to rely on point-wise \cite{Teed2021_RAFT3D} or segment-wise rigidity priors \cite{Golyanik2017_3DV,Ma2019_DRISF,Menze2015_KITTI,Vogel2015_PRSM}, or to explicitly learn segmenting rigid motions \cite{Yang2021_SegmentRigid}.
In combination with semantic information such rigidity estimates allow to assign rigid motions to all independently moving objects and to the background \cite{Ma2019_DRISF,Yang2021_SegmentRigid}.
% Finally, one can also reduce the difficulty of the problem.
And finally, it is also possible to reduce the difficulty of the problem.
This can either be done by decoupling stereo and 3D motion estimation \cite{Badki2021_binaryTTC,Liu2022_camliflow,Teed2021_RAFT3D,Wedel2011_SceneFlow,Yang2020_opticalexpansion},
% TODO: cite more? \eg \cite{Badki2021_binaryTTC,Yang2020_opticalexpansion,Liu2022_camliflow}
which also enables the use of dedicated state-of-the-art algorithms for stereo, or by directly relying on RGBD footage \cite{Golyanik2017_3DV,Gottfried_RGBDSceneFlow,Qiao2018_RGBDSceneFlow,Quiroga2014_RGBDSceneFlow}, \eg by using time-of-flight cameras or LiDAR \cite{Liu2022_camliflow}. 

%In view of the aforementioned progress in scene flow estimation it is %may seem
%surprising, that all currently leading neural net- works are limited to two time instances \cite{Badki2021_binaryTTC,Behl2017_ISF,Li2021_acosf,Ma2019_DRISF,Teed2021_RAFT3D,Yang2020_opticalexpansion,Yang2021_SegmentRigid}. However, although additional temporal information is likely to %be helpful 
%to improve the estimation, developing suitable network architectures for multi-frame scene flow estimation is %indeed 
%a difficult task; see \eg \cite{Hur2021_MonoSceneFlow}. % in the context of monocular unsupervised scene flow. 
%Hence, while for variational methods there have been successful attempts to integrate temporal information on a full model level \cite{Vogel2015_PRSM}, the few recent multi-frame neural networks for scene flow %estimation mainly focus on extending existing architectures by suitable temporal concepts. Such concepts include the propagation of motion object labels over time \cite{Neoral2017_SceneFlowTemporal}, the integration of a trainable motion inverter that allows to generate predictions from the past \cite{Schuster2021_DTF} %, the use of multiple frames to improve the estimation of the rigid background motion \cite{Taniai2017_MultiFrameSceneFlow} as well as the consideration of previous frames and previous scene flow estimates to improve %the estimation of the rigid background and to initialize the motion segmentation \cite{Taniai2017_MultiFrameSceneFlow}.

In view of the aforementioned progress in neural networks for scene flow estimation, it is remarkable that currently leading methods \cite{Badki2021_binaryTTC,Li2021_acosf,Liu2022_camliflow,Ma2019_DRISF,Teed2021_RAFT3D,Yang2020_opticalexpansion,Yang2021_SegmentRigid} 
do not exploit potentially valuable temporal information to further improve the results.
%While, there have been recently a few attempts to extend two-frame approaches to more than two stereo pairs \cite{Schuster2021_DTF,Hur2021_MonoSceneFlow} all state of the art approaches are restricted to the standard two-frame setting with two subsequent stereo pairs \cite{Devernay2007_SceneFlow}.
In fact, while differing in the actual inputs --
monocular images
\cite{Badki2021_binaryTTC,Yang2020_opticalexpansion}, stereo pairs \cite{Li2021_acosf,Liu2022_camliflow,Ma2019_DRISF,Yang2020_opticalexpansion,Yang2021_SegmentRigid,Teed2021_RAFT3D}, RGBD images \cite{Teed2021_RAFT3D}
or LiDAR point clouds \cite{Liu2022_camliflow} 
-- all leading networks are restricted to the standard two-frame setting.
%either using subsequent monocular images  \cite{Badki2021_binaryTTC,Yang2020_opticalexpansion}, stereo pairs \cite{Li2021_acosf,Ma2019_DRISF,Yang2020_opticalexpansion,Yang2021_SegmentRigid,Teed2021_RAFT3D,Liu2022_camliflow}, RGBD images \cite{Teed2021_RAFT3D} or LiDAR point clouds \cite{Liu2022_camliflow}.
%or other inputs that may be derived from stereo pairs if not directly available; \eg RGBD data \cite{Teed2021_RAFT3D} or LiDAR point clouds \cite{Liu2022_camliflow}.
%-- potentially converted to RGBD or LiDAR data in a preprocessing step. 
In this context, it is also surprising that the best multi-frame scene flow method on the KITTI benchmark is still a classical variational method which dates back to 2015 \cite{Vogel2015_PRSM}. 
This illustrates
that developing suitable multi-frame extensions of existing network 
architectures is indeed a difficult task.

\skipping{
The previously discussed observation is also reflected in
%the recent literature on multi-frame scene flow approaches
recent works on 
multi-frame scene flow networks \cite{Schuster2021_DTF,Hur2021_MonoSceneFlow}. 
While on the KITTI benchmark,
moderate improvements of 2\%-4\%\footnote{We considered from both papers the best overall results in terms of the standard SF-all outlier measure:
\cite{Schuster2021_DTF}: DTF-SENSE (9.18) vs.\ SENSE (9.55),
%\cite{Schuster2021_DTF}: DTF-PWOC (SF-all: 14.59) vs.\ PWOC-3D (SF-all: 15.63)\\
\cite{Hur2021_MonoSceneFlow}: Multi-Mono-SF-ft (33.09) vs.\ Self-Mono-SF-ft (33.88).}have been reported by those works, the underlying concepts were either not integrated into state-of-the-art
baselines \cite{Schuster2021_DTF} or they were developed for the even more challenging self-supervised monocular setting \cite{Hur2021_MonoSceneFlow}, 
not allowing the results to be competitive with those of leading supervised methods.
Moreover, in the meantime, the accuracy of currently leading two-frame networks improved by a factor two \cite{Liu2022_camliflow,Yang2021_SegmentRigid,Teed2020_RAFT} compared to the baseline in \cite{Schuster2021_DTF}. Therefore, it remains unclear, if multi-frame extensions are still capable of providing similar improvements in performance as observed for previous baselines with  significantly lower accuracy.
}

The latter observation is also reflected in the recent literature on
multi-frame scene flow networks \cite{Schuster2021_DTF,Hur2021_MonoSceneFlow}. 
On the one hand, on the KITTI benchmark,
only slight improvements of 2\%-4\% have been reported compared to the underlying two-frame baselines\footnote{We considered from both papers the best overall results in terms of the standard SF-all outlier measure:
\cite{Schuster2021_DTF}: DTF-SENSE (9.18) vs.\ SENSE (9.55),
%\cite{Schuster2021_DTF}: DTF-PWOC (SF-all: 14.59) vs.\ PWOC-3D (SF-all: 15.63)\\
\cite{Hur2021_MonoSceneFlow}: Multi-Mono-SF-ft (33.09) vs.\ Self-Mono-SF-ft (33.88).}.
Evidently,  for recent multi-frame architectures, the often much larger training gains  
%%of 15\%-28\%
do not generalize well to the actual test data.
%%\footnote{\cite{Schuster2021_DTF}: DTF-SENSE (9.57) vs.\ SENSE (11.84), \cite{Hur2021_MonoSceneFlow}: Multi-Mono-SF-ft (39.82) vs.\ Self-Mono-SF-ft (47.05).}.
On the other hand, the proposed multi-frame concepts were either not incorporated into state-of-the-art
baselines \cite{Schuster2021_DTF} or they were developed for the even more challenging self-supervised monocular setting~\cite{Hur2021_MonoSceneFlow}.
This in turn gives an explanation for the relatively poor overall performance of recent multi-frame methods 
compared to currently leading supervised two-frame approaches.
%Hence in both cases,
%the resulting overall performance  
%is far from being competitive with those of leading supervised %two-frame approaches.
And finally, as of today, the accuracy of leading two-frame %scene flow
approaches in general has improved by a factor two compared to the baseline in \cite{Schuster2021_DTF}; see \eg \cite{Liu2022_camliflow,Yang2021_SegmentRigid,Teed2020_RAFT}. This in turn raises the question
if suitable multi-frame extensions can be developed at all, if the underlying baseline already provides a sufficiently high accuracy.

\medskip
\noindent
\textbf{Contributions.}
% In our paper we will answer this question.
In our paper, we show that multi-frame ideas are still valuable in the context of recent high-accuracy networks.
%Based on 
Building upon the RAFT-3D method~\cite{Teed2021_RAFT3D}, we present a novel multi-frame approach that allows to leverage the performance of current two-frame techniques.
In this context we make the following contributions:
%our contributions are fivefold:
% (i) Initially, we show that the original RAFT-3D approach can be significantly improved by substituting the underlying stereo approach with a more suitable counterpart and by retraining the entire model.
% (ii) Then, building upon this improved baseline, we propose a novel multi-frame approach that particularly exploits the advantages of the underlying RAFT-3D architecture by combining a $SE(3)$ based prediction step with a U-Net based fusion architecture.
% (iii) Moreover, performing ablation studies and further experiments based on fourfold cross validation,  
% we illustrate the benefits of the different architectural components and identify a fusion strategy that generalizes well to the test data.
% (iv) And finally, with improvements of $9\%$ for the baseline and $16\%$ for the overall approach, we do not only report much larger performance gains than %those of 
% existing multi-frame scene flow networks, but also achieve competitive results,
% eventually ranking second in the KITTI scene flow benchmark.
(i) We propose a multi-frame architecture that particularly exploits the advantages of the underlying RAFT-3D architecture by combining a $SE(3)$ based prediction step with a U-Net based fusion architecture. In this context, we also improve the underlying two-frame baseline by substituting the employed stereo approach.
(ii) Performing ablation studies and further experiments based on fourfold cross validation, we illustrate the benefits of the different architectural components of our method. In this way, we identify a fusion strategy that generalizes well to the test data.
%(iii) And finally, using an improved baseline, our overall approach reaches improvements of $16\%$, reporting not only much larger performance gains than existing multi-frame scene flow networks, but also achieving competitive results, eventually ranking second in the KITTI scene flow benchmark.
(iii) With improvements of $9\%$ for the baseline and $16\%$ for the overall approach, we report much larger performance gains than existing multi-frame networks from the literature. These gains also lead to highly competitive results, eventually ranking second in the KITTI scene flow benchmark.

\section{Related work}
%Approaches related to our method can be roughly divided into two %groups: multi-frame scene flow approaches and online prediction %methods. Both groups will be discussed in the following.

%\medskip
% \noindent {\bf Multi-frame scene flow.} 
\paragraph{Multi-frame scene flow.}
Regarding the use of multiple time frames for scene flow estimation,
one can mainly distinguish three types of methods.
Like our approach, most of them rely on a three frame setting that has proven to be a good compromise between available temporal information and efficiency for both optical flow \cite{Maurer2018_ProFlow,Maurer2018,Ren2019_FlowTemporalFusion,Wulff2017_MRFlow} and scene flow \cite{Hur2021_MonoSceneFlow,Neoral2017_SceneFlowTemporal,Schuster2021_DTF,Schuster2020_SceneFlowFields,Taniai2017_MultiFrameSceneFlow}. 

(i) On the one hand, there are approaches that explicitly model multi-frame scene flow in terms of an {\em energy minimization} framework. Such approaches are the method of Vogel \etal \cite{Vogel2015_PRSM} that, based on piece-wise rigidity assumption, enforces a consistent piece-wise planar segmentation over time, the method of Golyanik \etal \cite{Golyanik2017_3DV} that follows a similar idea but relies on RGBD data instead of stereo sequences, the method of Taniai \etal \cite{Taniai2017_MultiFrameSceneFlow} which fuses estimates from optical flow and multi-frame time stereo, %\ie structure from motion,
and the method of Neoral and \v{S}ochman \cite{Neoral2017_SceneFlowTemporal} that extends the two-frame scene flow approach of Menze and Geiger \cite{Menze2015_KITTI} by additionally propagating object labels over time.

(ii) On the other hand, there are {\em sparse-to-dense} methods that speed up the computation of energy-based methods by considering sparse matching strategies followed by a robust interpolation step. Such a method is the approach of Schuster \etal \cite{Schuster2020_SceneFlowFields}, which performs a sparse multi-frame matching relying on the assumption that the 3D motion in terms of the scene flow is constant over time. % (three frames). 

(iii) And finally, %most related to our work, 
%as in the two-frame case, 
also {\em neural networks}
gained recently popularity in the context of multi-frame scene flow. 
%Such methods include another approach of Schuster \etal \cite{Schuster2021_DTF} that predicts the forward flow from the backward flow with a dedicated motion inverter and fuses both flows using a convex fusion step, and the self-supervised monocular approach of Hur and Roth \cite{Hur2021_MonoSceneFlow} which proposes to use a convolutional LSTM to encourage consistency over time, while forward warping the latent variables to compensate for changes in the location due to motion. 
%Such methods include the self-supervised monocular approach of Hur and Roth \cite{Hur2021_MonoSceneFlow} which uses a convolutional LSTM to encourage consistency over time as well as another recent supervised approach of Schuster \etal \cite{Schuster2021_DTF} that predicts the forward flow from the backward flow based on a small dedicated motion inverter and subsequently fuses both flows using a convex fusion step. 
Such methods include another approach of Schuster \etal \cite{Schuster2021_DTF} that predicts the forward from the backward flow based on a small learned motion inverter and subsequently fuses both flows using a convex fusion step, and the self-supervised monocular approach of Hur and Roth \cite{Hur2021_MonoSceneFlow} that uses a convolutional LSTM to encourage consistency over time.

%While our novel approach is also based on a neural network that fuses backward and forward flows, it differs significantly from the approach of Schuster \etal \cite{Schuster2021_DTF}. Baseline-wise, it relies on RAFT-3D, a recent high accuracy two-frame method that is based on a local $SE(3)$ parametrization and performs iterative updates using a recurrent unit. This allows us to exploit both the advanced parametrization and the valuable inputs of the recurrent unit when adaptively incorporating temporal information, \ie to tailor our approach explicitly towards the underlying baseline.
%On the other hand, methodology-wise, instead of learning a motion model via a dedicated motion inverter and then explicitly having to rely on it using a convex fusion module, we predict the motion using a $SE(3)$-based extrapolation with a constant motion model and then consider a more generalized U-Net based fusion step that implicitly learns possibly required motion corrections during the fusion. 

While our multi-frame method is also based on a neural network that fuses flow
estimates,
%predictions from forward and backward flows, 
its underlying strategy differs significantly 
%both conceptually and integration-wise 
from the one in
\cite{Schuster2021_DTF}.
%Integration-wise, 
On the one hand, our method not only relies on a much more advanced baseline, \ie RAFT-3D. Its entire architecture is also specifically tailored towards this baseline; \eg our method exploits both the local $SE(3)$ parametrization as well as the valuable inputs of RAFT-3D's recurrent unit when adaptively integrating temporal information.
%Conceptually, 
On the other hand, instead of learning a motion model via a small motion inverter that is naturally limited in its generalization capabilities and subsequently restricting the fusion to a convex combination, our method predicts the motion using a SE(3)-based extrapolation and then considers a more generalized U-Net based fusion step. 
While the SE(3)-based prediction holds in many scenarios, the generalized fusion step allows to implicitly learn possibly required corrections of this prediction. 

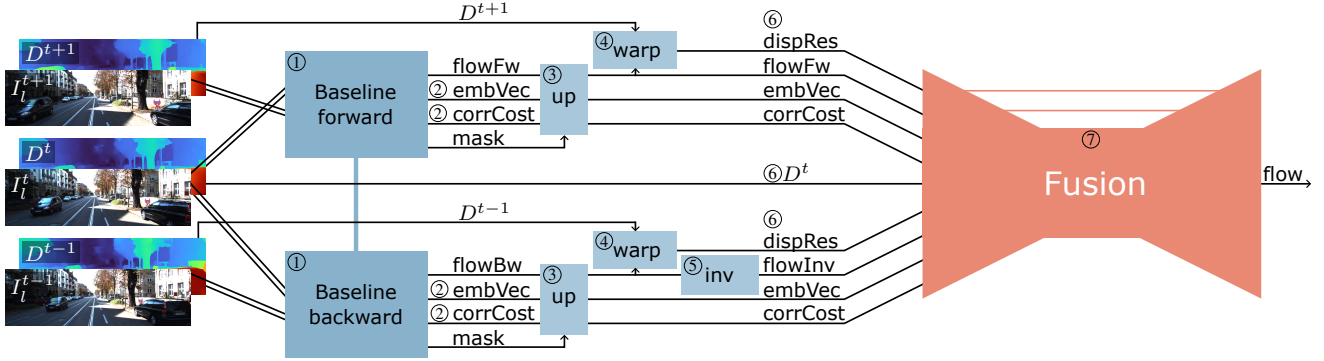
\begin{figure*}
    \centering
    \tikzset{overviewstyle/.style={anchor=north west, text=white, inner sep=0.8, text opacity=1, fill=black, opacity=0.4}}
    \begin{tikzpicture}
    \draw (0, 0) node[inner sep=0,anchor=north west] (img) {
	\includegraphics[width=\textwidth]{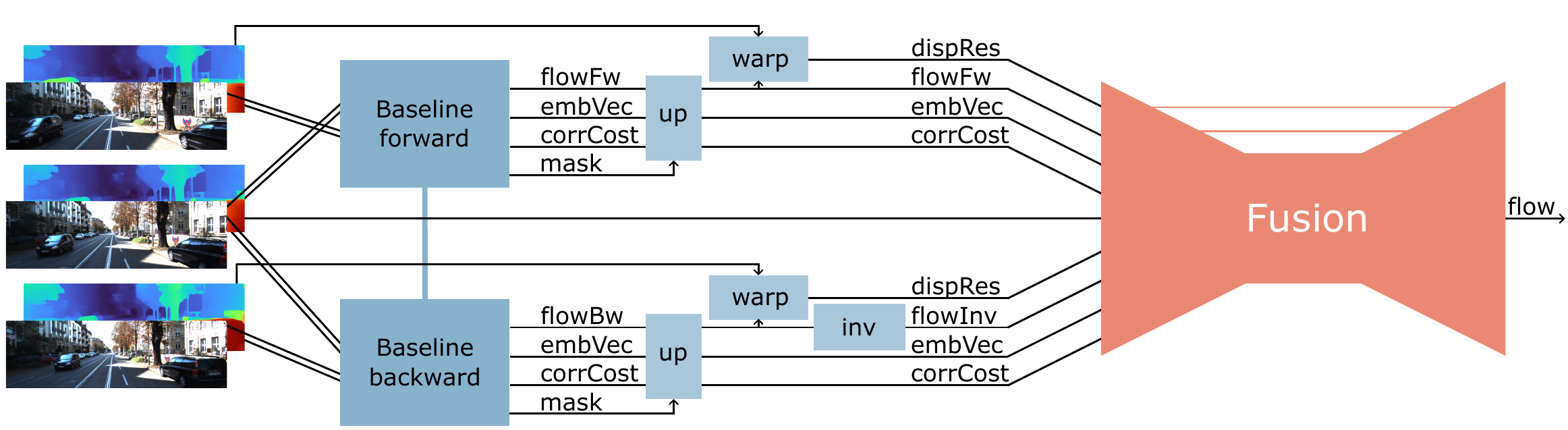}
	};
	\draw (0.30, -0.54) node[overviewstyle] {\footnotesize $D^{t+1}$};
	\draw (0.11, -0.96) node[overviewstyle] {\footnotesize $I^{t+1}_l$};
	\draw (0.30, -1.87) node[overviewstyle] {\footnotesize $D^{t}$};
	\draw (0.11, -2.28) node[overviewstyle] {\footnotesize $I^{t}_l$};
	\draw (0.30, -3.19) node[overviewstyle] {\footnotesize $D^{t-1}$};
	\draw (0.11, -3.61) node[overviewstyle] {\footnotesize $I^{t-1}_l$};
	\draw (6.09, -0.02) node[anchor=north west, inner sep=0] {\footnotesize $D^{t+1}$};
	\draw (6.09, -2.68) node[anchor=north west, inner sep=0] {\footnotesize $D^{t-1}$};
	\draw (10.4, -2.16) node[anchor=north west, inner sep=0] {\footnotesize $D^{t}$};
    \draw (3.95, -0.82) node[scale=0.8] {\footnotesize \circled{1}};
    \draw (3.95, -3.48) node[scale=0.8] {\footnotesize \circled{1}};
    \draw (5.83, -1.18) node[scale=0.78] {\footnotesize \circled{2}};
    \draw (5.83, -1.50) node[scale=0.78] {\footnotesize \circled{2}};
    \draw (5.85, -3.84) node[scale=0.78] {\footnotesize \circled{2}};
    \draw (5.85, -4.16) node[scale=0.78] {\footnotesize \circled{2}};
    \draw (7.35, -0.99) node[scale=0.8] {\footnotesize \circled{3}};
    \draw (7.35, -3.65) node[scale=0.8] {\footnotesize \circled{3}};
    \draw (8.04, -0.54) node[scale=0.78] {\footnotesize \circled{4}};
    \draw (8.04, -3.20) node[scale=0.78] {\footnotesize \circled{4}};
    \draw (9.22, -3.53) node[scale=0.8] {\footnotesize \circled{5}};
    \draw (10.28, -0.25) node[scale=0.8] {\footnotesize \circled{6}};
    \draw (10.28, -2.30) node[scale=0.8] {\footnotesize \circled{6}};
    \draw (10.28, -2.91) node[scale=0.8] {\footnotesize \circled{6}};
    \draw (14.51, -1.85) node[scale=0.8] {\footnotesize \circled{7}};
    % \draw[step=1.0,gray,thin] (0,0) grid (15,-5);
	\end{tikzpicture}
    \caption{Overview of our M-FUSE approach (see Sec.\ 3.2). We employ two shared instances of our baseline model to predict forward ($t \rightarrow t\!+\!1$) and backward ($t \rightarrow t\!-\!1$) scene flow as well as additional features used in our fusion U-Net to predict the final flow estimate.}
    \label{fig:overview}
\end{figure*}
\begin{figure*}
    \centering
    {
    \setlength\tabcolsep{1pt}
    \begin{tabular}{cccccc}
    $D^t$ & $\Delta d$ & $(u,v)$ & \emph{embVec} & \emph{corrCost} & \emph{dispRes}
    \\
    \multirow{2}{*}{
    \includegraphics[width=0.16\textwidth]{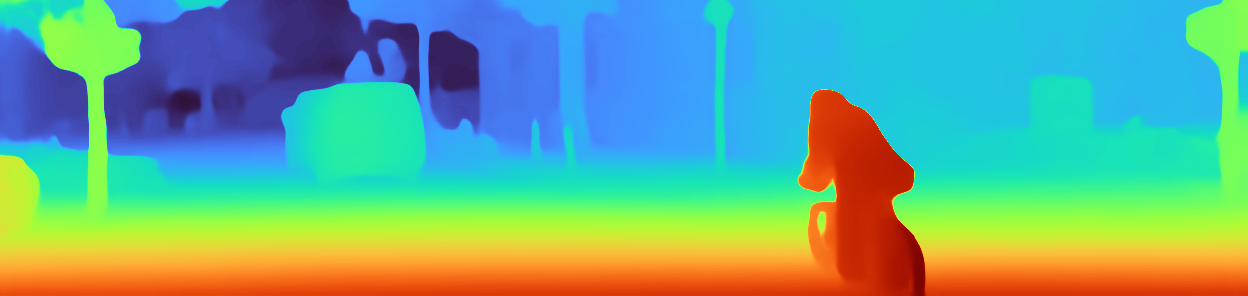}
    }\hspace{-0.8mm}
    &
    \includegraphics[width=0.16\textwidth]{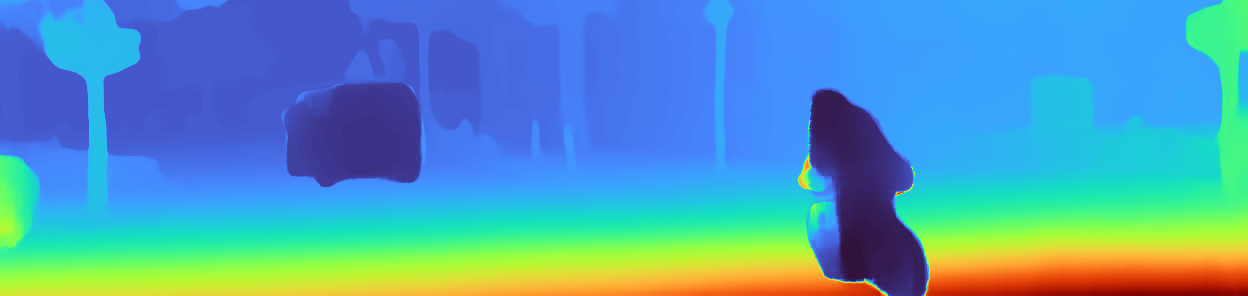}
    &
    \includegraphics[width=0.16\textwidth]{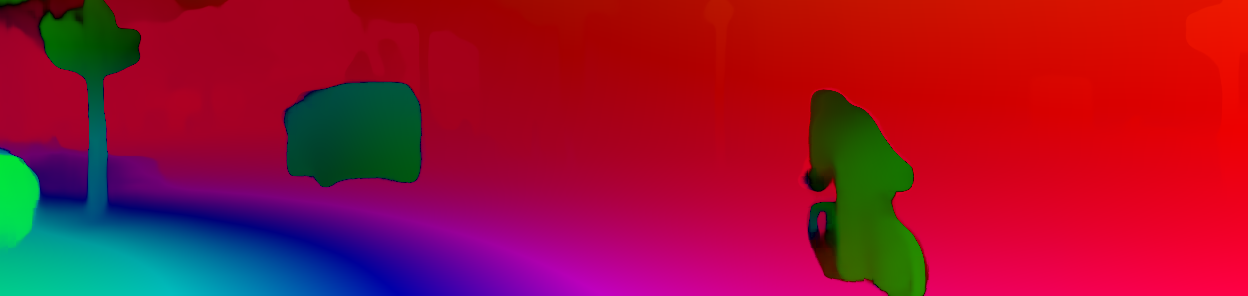}
    &
    \includegraphics[width=0.16\textwidth]{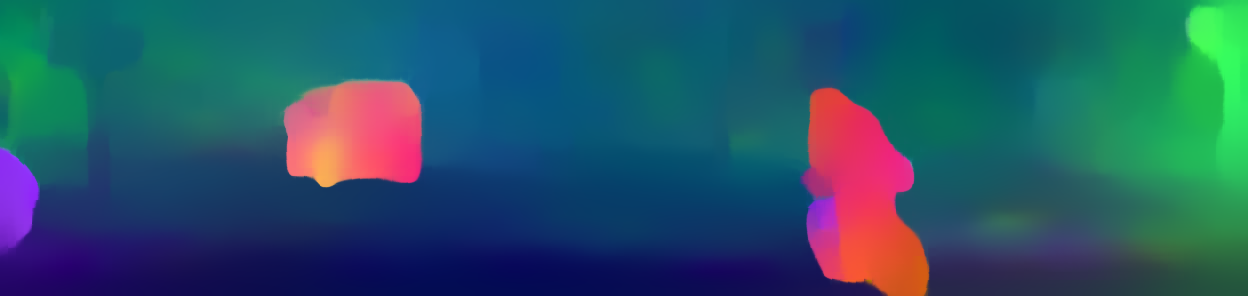}
    &
    \includegraphics[width=0.16\textwidth]{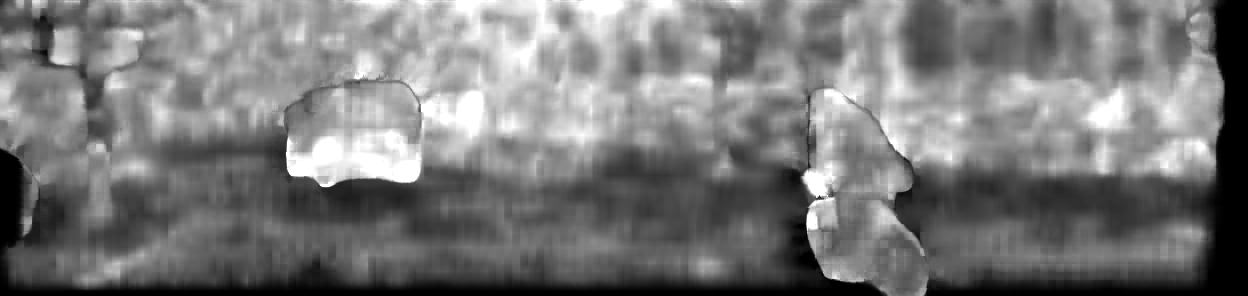}
    &
    \includegraphics[width=0.16\textwidth]{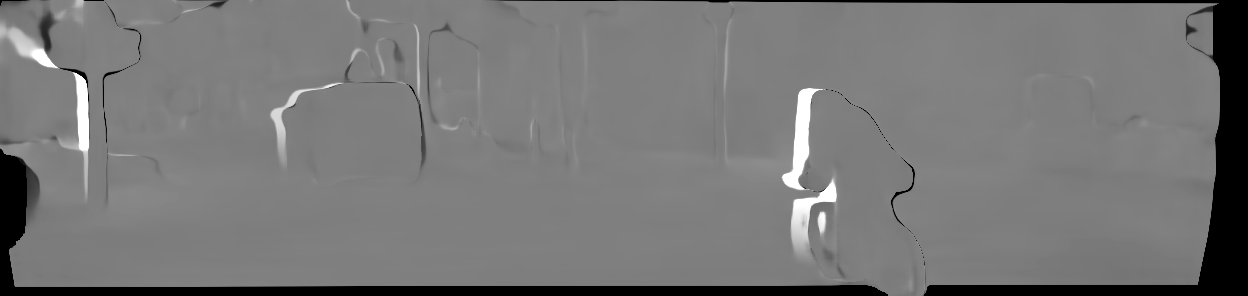}
    \\[-0.5mm]
    &
    \includegraphics[width=0.16\textwidth]{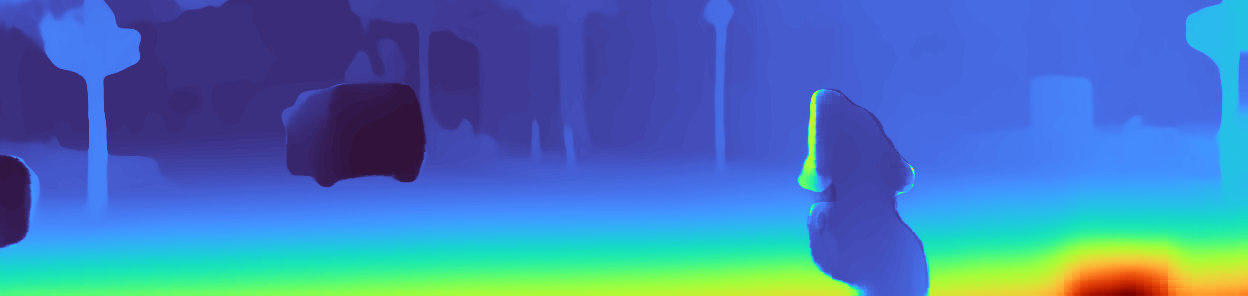}
    &
    \includegraphics[width=0.16\textwidth]{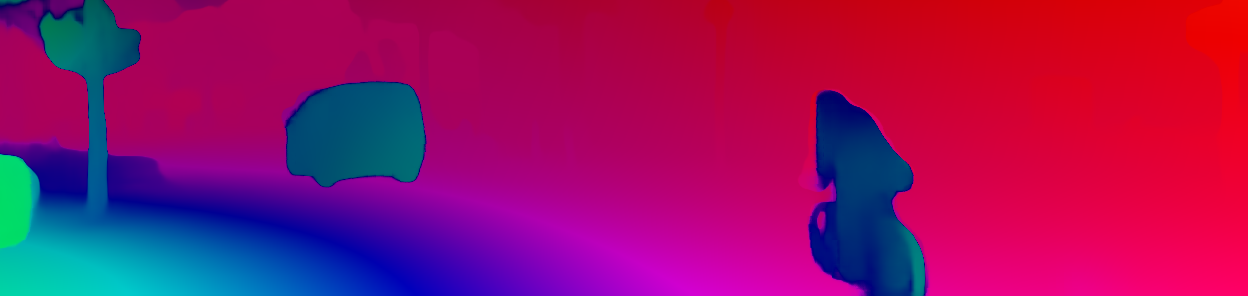}
    &
    \includegraphics[width=0.16\textwidth]{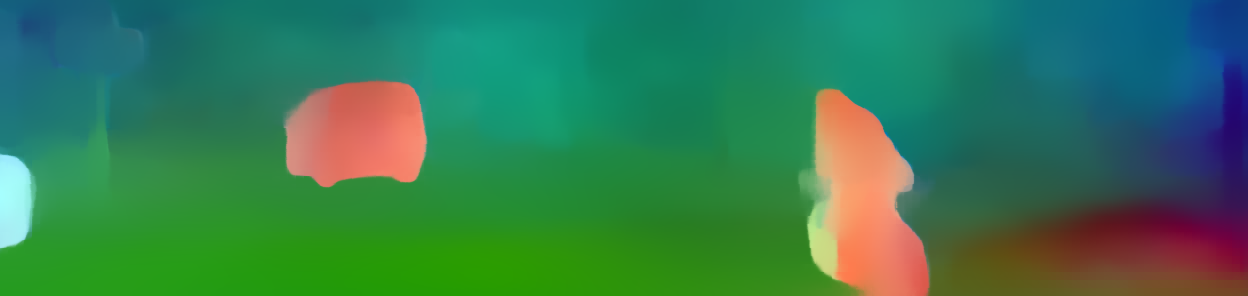}
    &
    \includegraphics[width=0.16\textwidth]{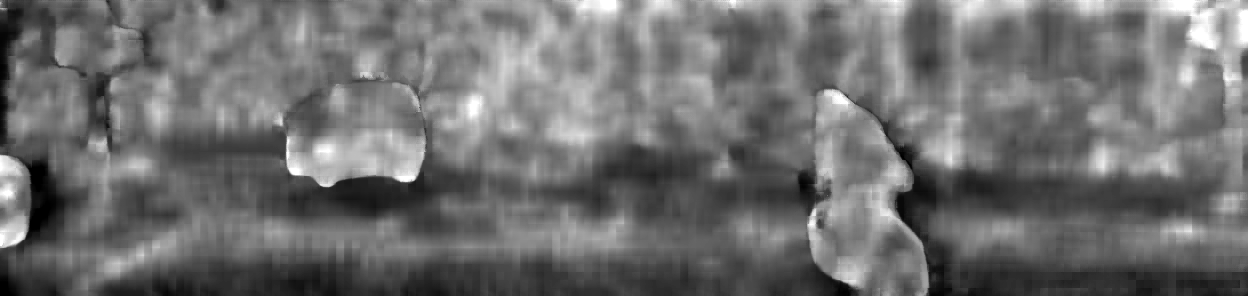}
    &
    \includegraphics[width=0.16\textwidth]{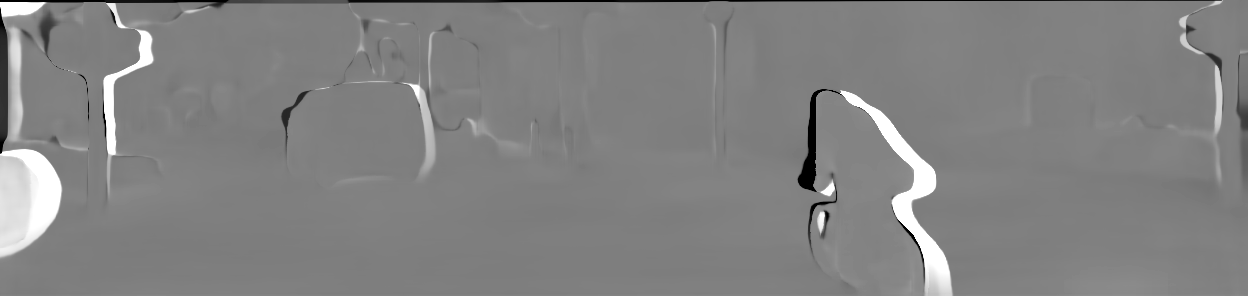}
    \end{tabular}
    \caption{Input features to our fusion module. \emph{From left to right:} Disparity in the reference frame $D^t$, disparity change $\Delta d$, optical flow $(u,v)$, rigid-motion embedding vectors, correlation cost, disparity residuals.
    \emph{Top row:} features from forward direction, \emph{bottom row:} backward direction.
    We use color visualizations for disparity/optical flow, and PCA to reduce the 16-channel embedding vectors to RGB.}
    \label{fig:inputs}
    }
\end{figure*}

\medskip
\noindent
\textbf{Multi-frame optical flow.}
In contrast to %the two approaches \cite{Schuster2021_DTF,Hur2021_MonoSceneFlow} for
scene flow estimation (3D motion), there is already an extensive literature on \emph{multi-frame neural networks} in the closely related field of optical flow estimation 
%(2D motion). %
(2D motion = projected scene flow). 
In this context, one can roughly distinguish three strategies:
%A large field in the literature that is closely related to scene flow is the field of optical flow, where several methods using more than two time instances have been proposed. %While developed for the simpler problem formulation, there are several multi-frame strategies that have the potential of being transferred to scene flow estimation.

%A first strategy explicitly estimates \emph{rigid-body motion} which is combined with unconstrained optical flow estimates.
%To this end, Wulff \etal~\cite{Wulff2017_MRFlow} propose a joint estimation of the scene structure and the rigid motion of static objects combined with unconstrained motion estimation for non-static objects.
%In a similar direction, Maurer \etal~\cite{Maurer2018} propose to fuse motion estimates computed under a static scene assumption with motion estimates from unconstrained optical flow.
(i) A first common strategy is to rely on a relaxed version of a \emph{constant motion model}. This can either be achieved by adding a temporal smoothness constraint to the loss such as the unsupervised method of Janai \etal~\cite{Janai2018_UnsupMultiframeOpticalFlow} or, even less strict, by initializing the estimation 
with the motion-compensated flow from the previous time step
% with the previous flow compensated by the corresponding change in location 
(warm start) as employed by the method of Teed and Deng \cite{Teed2020_RAFT}.

(ii) A second strategy is to estimate multiple flows in a {\em joint recurrent unit} ($\neq$ shared recurrent unit). A corresponding self-supervised approach has been proposed by Liu \etal~\cite{Liu2019_SelfSupervisedOpticalFlow} who estimate forward and backward flows %in a three-frame setting 
with a double cost volume, which allows to learn flow in occluded regions without an explicit motion model.

(iii) A third  strategy seeks to improve the estimation by incorporating a \emph{learned motion model}. 
To this end, Maurer and Bruhn~\cite{Maurer2018_ProFlow} proposed an approach that integrates predictions from a motion model that is learned with a small neural network on the fly. Similarly, Stone \etal \cite{Stone2021_UnsupOpticalFlow} incorporated this idea in an unsupervised method, where the %learned 
motion model helps to teach the flow in occluded regions. Alternatively, instead of learning the model directly, Ren \etal~\cite{Ren2019_FlowTemporalFusion} fuse a forward flow estimate based on a constant motion model, but allow corrections in a learned follow-up fusion step. To this end, they employ the %U-Net based
fusion module from FlowNet2~\cite{Ilg2017_Flownet2} originally designed for combining small and large displacements.

%with warped optical flow from the previous frame using the fusion module from FlowNet2~\cite{Ilg2017_Flownet2}.

%In contrast to these multi-frame strategies for optical flow, our approach works in the scene flow setting. There, predictions are, per constructions, much more meaningful, since they are not restricted to projected 2D space. On the one hand, this allows us to perform a more realistic $SE(3)$-based prediction. On the other hand, when fusing this prediction with the original forward flow, we can consider additionally disparity information which makes it simpler for the fusion network to learn possibly required corrections. Similar to \cite{Ren2019_FlowTemporalFusion}, we employ a U-Net architecture. However, we tightly integrate our model with the baseline approach and do not use warping of previous flow estimates.

% While our approach is similar to \cite{Ren2019_FlowTemporalFusion} in the sense that it also relies on a U-Net architecture, it generalizes the underlying ideas to scene flow, \ie an unprojected and hence more meaningful representation of motion in a 3D scene. 
While our approach is similar to \cite{Ren2019_FlowTemporalFusion} in the sense that it also relies on a U-Net architecture, it generalizes the underlying ideas to the scene flow setting, \ie a setting where the motion in a 3D scene is unprojected and hence more meaningful -- in contrast to optical flow.
This offers new possibilities such as predictions by more realistic motion models (constant scene flow, constant $SE(3)$ motion) as well as a much better guided fusion (having access to optical flow and disparity). 
Moreover, unlike \cite{Ren2019_FlowTemporalFusion, Stone2021_UnsupOpticalFlow} our approach is specifically tailored towards the baseline network, \ie RAFT-3D, explicitly exploiting the underlying network characteristics (parametrization, cost volume, convex upsampling, high resolution disparity). 
As our experiments show, such a tight integration is required to further improve state-of-the-art two-frame scene flow networks.
Finally, in contrast to \cite{Ren2019_FlowTemporalFusion} our approach predicts
%conceptually, our approach follows rather \cite{Maurer2018_ProFlow} than \cite{Ren2019_FlowTemporalFusion} by predicting 
the forward from the backward flow 
%($t \rightarrow t-1$) 
instead of using the previous forward flow. 
%($t \rightarrow t$). 
This, in turn, avoids registering the information from the previous time frame (optical flow, fusion features) via motion-based interpolation (warping).

\section{Approach}
We propose a neural network for scene flow estimation from a triplet of stereo frames.
Given the three stereo frames
$(I^{t-1}_l, I^{t-1}_r), (I^{t}_l, I^{t}_r)$ and $(I^{t+1}_l, I^{t+1}_r)$,
our goal is to estimate the four-dimensional scene flow $(u,v,d,d')$~\cite{Devernay2007_SceneFlow} between frame $t$ and frame $t\!+\!1$.
Here, $(u,v)$ denotes the optical flow and $d$ and $d'$ are the disparities at time $t$ and $t\!+\!1$ registered to the reference frame $I^{t}_l$.

Following recent works~\cite{Badki2021_binaryTTC,Liu2022_camliflow,Teed2021_RAFT3D,Yang2020_opticalexpansion}, we thereby decouple the disparity estimation from the recovery of the 3D motion.
To this end, we precompute disparities for each stereo frame using a dedicated stereo method, yielding $D^{t-1}$, $D^t$ and $D^{t+1}$.
This allows us to directly take over the final estimate for $d$ from $D^t$.
Hence, the scene flow problem reduces to estimating $(u,v,d')$.
Please note that $d'$ cannot be taken over directly from $D^{t+1}$ since $d'$ is registered to $I^{t}_l$ while $D^{t+1}$ is registered to $I^{t+1}_l$.
However, knowing $D^t$, we can easily convert between $d'$ and the change in disparity $\Delta d$, with $d' = D^t + \Delta d$ when estimating the scene flow.

Using this notation, let us now explain our two-frame baseline and subsequently our full multi-frame network.

\subsection{Improved two-frame baseline}
% We base our work on the recent two-frame approach RAFT-3D~\cite{Teed2021_RAFT3D}.
% This method employs a two-step pipeline that first uses an off-the-shelf stereo estimation network to compute left-right disparities and then estimates the scene flow while keeping the reference disparity $d$ fixed.
We base our work on the recent two-frame approach RAFT-3D~\cite{Teed2021_RAFT3D}, which first uses an off-the-shelf stereo estimation network to compute left-right disparities and then estimates the scene flow while keeping the reference disparity $D^t$ fixed.
% The approach employs a recurrent neural network which operates at 1/8th of the original resolution, where the final result is obtained by a convex upsampling strategy~\cite{Teed2020_RAFT} based on a mask predicted by the network.
The approach employs a recurrent neural network which operates at 1/8th of the original resolution, where the final result is obtained by a learned convex upsampling~\cite{Teed2020_RAFT}.
% predicts scene flow on 1/8th of the original resolution, which is brought back to high resolution by a learned mask-based convex upsampling.
Notably, RAFT-3D predicts the scene flow in terms of a field of $SE(3)$ transformation matrices, and afterwards translates them to the standard parametrization $(u,v,d')$.
Building upon this work, we proceed in two steps.
% We build upon this work in two steps.
Making use of recent progress in the field of stereo estimation, we first exchange RAFT-3D's stereo estimation network, GANet~\cite{Zhang2019_GANet} with the recently well-performing LEAStereo~\cite{Cheng2020_LEAStereo}.
Subsequently, with the improved stereo results, we fully retrain RAFT-3D using their provided code.
This way, we obtain an improved two-frame baseline serving as a building block for our multi-frame network.

\subsection{Multi-frame fusion network}

Conceptually, our multi-frame fusion network consists of two shared instances of %a baseline 
our improved two-frame baseline and a fusion module predicting the final scene flow.
More precisely, given two initial motion estimates for the forward and backward flow in low resolution, we derive low-resolution features which we adaptively upsample and subsequently combine with high-resolution features, eventually fusing forward and inverted backward flow estimates in a feature-guided high-resolution fusion module.
In the following, we discuss all steps of our approach in detail; see Figure~\ref{fig:overview} for a complete overview.

\medskip
\noindent
\circled{1} \textbf{Initial scene flow estimation.}
Initially, our improved two-frame baseline predicts forward ($t \rightarrow t\!+\!1$) and backward ($t \rightarrow \!t-\!1$) scene flow at 1/8th of the full image resolution; as in the original RAFT-3D method~\cite{Teed2021_RAFT3D}.
It predicts flow estimates in terms of a field of $SE(3)$ transformation matrices and a weighting mask for convex upsampling.

\medskip
\noindent
\circled{2} \textbf{Low-resolution features.}
In order to later guide our fusion of flow estimates, we consider two features derived from the specific architecture of our baseline.
First, rigid-motion embedding vectors (\emph{emb\-Vec}) are essential to our baseline method, as they are used for a soft-grouping of pixels that belong to objects with the same rigid motion~\cite{Teed2021_RAFT3D}.
Since this segmentation information can be valuable for the fusion of forward and backward flow estimates, we utilize these features as an input to our fusion module.
To this end, we extract the 16-channel prediction of the rigid-motion embedding vectors by the network for both the forward and the backward baselines.
Second, the cost volume is at the core of recent motion estimation algorithms~\cite{Sun2018_PWC,Teed2020_RAFT,Teed2021_RAFT3D} since it assigns matching costs to potential flow estimates.
In order to better guide our fusion module, we look up the correlation costs (\emph{corr\-Cost}) for the current flow estimates in forward and backward direction, which provides supporting information on the quality of the estimates.
Note that we omit the multi-scale pyramid and the spatially extended lookup employed by \cite{Teed2020_RAFT,Teed2021_RAFT3D} and only extract a single cost value per pixel for the central location.

\medskip
\noindent
\circled{3} \textbf{Joint convex upsampling.}
So far, the flow predictions as well as the extracted features are given on 1/8th of the original resolution.
As the next step, we will hence exploit the convex upsampling mask predicted by the baseline networks in order to obtain flow predictions and features on the original high resolution.
This proceeding offers three advantages:
We can utilize disparity maps, which are given at the original resolution, we can perform backward-to-forward prediction at the original resolution and we can ultimately fuse flows at the original resolution.

\medskip
\noindent
\circled{4} \textbf{High-resolution features.}
With the correlation cost at hand, we have matching information based on image features, but so far we do not make use of any disparity cues to guide the fusion.
In order to create meaningful features, we first convert the upsampled forward and backward transformation fields to optical flow and disparity estimates which yields $(u_\text{fw},v_\text{fw}, d'_\text{fw})$ and $(u_\text{bw},v_\text{bw}, d'_\text{bw})$, respectively.
Then, we warp the initial high-resolution disparity estimates $D^{t+1}$ and $D^{t-1}$ using these optical flows such that they are aligned with the reference frame and subtract the corresponding disparity estimates in order to compute disparity residuals (\emph{dispRes}) for both directions as
\begin{align}
    \mathcal{W}(D^{t+1}, u_\text{fw},v_\text{fw}) &- d'_\text{fw} \;,
    \\
    \mathcal{W}(D^{t-1}, u_\text{bw},v_\text{bw}) &- d'_\text{bw} \;,
\end{align}
where $\mathcal{W}(D, u,v)$ denotes backward-warping of $D$ using the optical flow $(u,v)$.
If the correct scene flow is given, the residuals are 0 for non-occluded pixels~\cite{Teed2021_RAFT3D}.

\medskip
\noindent
\circled{5} \textbf{Backward-to-forward prediction.}
In our initial flow estimation, we predicted backward flow pointing towards the previous frame.
In order to obtain a meaningful prediction in forward direction, we utilize the $SE(3)$ motion parametrization of the upsampled scene flow and invert the backward transformations with a differentiable matrix inversion~\cite{Teed2021_Lie}.
Note that this inversion in matrix space is ca-\linebreak pable of performing true inversion of rotational motion ra- ther than simple linear inversion in the standard scene flow representation, which only flips the sign.
Subsequently, we convert the matrix representation of the forward and the inverted backward flow to optical flow and disparity change, which is the parametrization we employ for the fusion.

\medskip
\noindent
\circled{6} \textbf{Fusion inputs.}
As a final step, we concatenate the forward and backward flow and all features which we provide to the fusion module, yielding 43 channels that are visualized in Figure~\ref{fig:inputs}.
Summarizing, we employ the disparity in the reference frame $D^t$ and for forward and backward direction scene flow estimates $(u ,v, \Delta d)$, rigid-motion embedding vectors, correlation costs and disparity residuals.

\medskip
\noindent
\circled{7} \textbf{Fusion module.}
With a rich set of inputs at hand, we apply our fusion module to predict a final scene flow estimate.
The fusion module is a CNN that uses a U-Net architecture~\cite{Ronneberger2015_UNet}, employing three depth levels with channel sizes 64, 128 and 256, where each level in the downsampling as well as in the upsampling branch consists of two $3 \times 3$ convolutional layers with stride 1 and zero-padding, to preserve image dimensions.
The downsampling uses the same convolutional layers with a stride of 2 and upsampling employs transposed convolutions with kernel size 4, stride 2 and zero-padding of 1.
Similar to the original U-Net, we use residual connections between the downsampling and the upsampling branch, however, instead of concatenation, we add the upsampled tensor to the skip connection.
After each convolutional layer, a LeakyReLU activation~\cite{Maas2013_leakyrelu} with slope $0.1$ is applied.
Finally, the three-channel output is predicted with one $3 \times 3$ convolution without activation.
Please note that in contrast to~\cite{Schuster2021_DTF} our generalized fusion is not restricted to a convex combination, \ie a linear blending of predictions and forward flow.
Hence, it implicitly allows to perform corrections when performing the fusion, in case predictions or forward flow are not accurate.

\begin{table*}
\caption{Top ranking non-anonymous submissions to the KITTI benchmark.}
\label{tab:KITTI_results}
\begin{center}
\begin{tabular}{lcc>{\columncolor[gray]{0.95}[6pt][5pt]}c
                 cc>{\columncolor[gray]{0.95}[6pt][5pt]}c
                 cc>{\columncolor[gray]{0.95}[6pt][5pt]}c
                 cc>{\columncolor[gray]{0.95}[6pt][5pt]}c}
\toprule
Method & \!\!D1-bg\!\! & \!\!D1-fg\!\! & \!\!D1-all\!\! & \!\!D2-bg\!\! & \!\!D2-fg\!\! & \!\!D2-all\!\! & \!\!Fl-bg\!\! & \!\!Fl-fg\!\! & \!\!Fl-all\!\! & \!\!SF-bg\!\! & \!\!SF-fg\!\! & \!\!SF-all\\
\midrule
DTF\_SENSE~\cite{Schuster2021_DTF} & 2.08 & \underline{3.13} & 2.25 & 4.82 & 9.02 & 5.52 & 7.31 & 9.48 & 7.67 & 8.21 & 14.08 & 9.18
\\
PRSM~\cite{Vogel2015_PRSM} & 3.02 & 10.52 & 4.27 & 5.13 & 15.11 & 6.79 & 5.33 & 13.40 & 6.68 & 6.61 & 20.79 & 8.97
\\
Binary TTC~\cite{Badki2021_binaryTTC} & \underline{1.48} & 3.46 & \underline{1.81} & 3.84 & 9.39 & 4.76 & 5.84 & 8.67 & 6.31 & 7.45 & 13.74 & 8.50
\\
Stereo expansion~\cite{Yang2020_opticalexpansion}\!\!\! & \underline{1.48} & 3.46 & \underline{1.81} & 3.39 & 8.54 & 4.25 & 5.83 & 8.66 & 6.30 & 7.06 & 13.44 & 8.12
\\
ISF~\cite{Behl2017_ISF} & 4.12 & 6.17 & 4.46 & 4.88 & 11.34 & 5.95 & 5.40 & 10.29 & 6.22 & 6.58 & 15.63 & 8.08
\\
ACOSF~\cite{Li2021_acosf} & 2.79 & 7.56 & 3.58 & 3.82 & 12.74 & 5.31 & 4.56 & 12.00 & 5.79 & 5.61 & 19.38 & 7.90
\\
UberATG-DRISF~\cite{Ma2019_DRISF}\!\!\!\! & 2.16 & 4.49 & 2.55 & 2.90 & 9.73 & 4.04 & 3.59 & 10.40 & 4.73 & 4.39 & 15.94 & 6.31
\\
RAFT-3D~\cite{Teed2021_RAFT3D} & \underline{1.48} & 3.46 & \underline{1.81} & 2.51 & 9.46 & 3.67 & 3.39 & 8.79 & 4.29 & 4.27 & 13.27 & 5.77
\\
RigidMask+ISF~\cite{Yang2021_SegmentRigid}\!\!\! & 1.53 & 3.65 & 1.89 & 2.09 & 8.92 & 3.23 & \underline{2.63} & 7.85 & 3.50 & \underline{3.25} & 13.08 & 4.89
\\
CamLiFlow~\cite{Liu2022_camliflow} & \underline{1.48} & 3.46 & \underline{1.81} & \textbf{1.92} & \underline{8.14} & \textbf{2.95} & \textbf{2.31} & \textbf{7.04} & \textbf{3.10} & \textbf{2.87} & \underline{12.23} & \textbf{4.43}
\\
\midrule
\textbf{M-FUSE (ours)} & \textbf{1.40} & \textbf{2.91} & \textbf{1.65} & 2.14 & \textbf{8.10} & \underline{3.13} & 2.66 & \underline{7.47} & \underline{3.46} & 3.43 & \textbf{11.84} & \underline{4.83}
\\
\textbf{Baseline}\!\!\! & \textbf{1.40} & \textbf{2.91} & \textbf{1.65} & \underline{1.97} & 9.22 & 3.17 & 2.98 & 9.51 & 4.06 & 3.53 & 13.57 & 5.20
\\
\midrule
Baseline Improvements &&&&&&&&&&&&\\
(Baseline vs. RAFT-3D) & \!\!\!5.4\%\!\!\! & \!\!\!15.9\%\!\!\! & \!\!\!8.8\%\!\!\! & \!\!\!21.5\%\!\!\! & \!\!\!2.5\%\!\!\! & \!\!\!13.6\%\!\!\! & \!\!\!12.1\%\!\!\! & \!\!\!-8.2\%\!\!\! & \!\!\!5.4\%\!\!\! & \!\!\!17.3\%\!\!\! & \!\!\!-2.3\%\!\!\! & \!\!\!9.9\%\!\!\!
\\
\midrule
Multi-frame Improvements\!\!\!\!    &        &         &         &         &           &         &         &         &         &         &         &\\
(M-FUSE vs. Baseline) &      - &       - &       - &  \!\!\!-8.6\%\!\! &   \!\!\!12.1\%\!\!  &  \!\!\!1.3\%\!\! &  \!\!\!10.7\%\!\! &  \!\!\!21.5\%\!\! &  \!\!\!14.8\%\!\! &  \!\!\!2.8\%\!\! &  \!\!\!12.7\%\!\! &  \!\!\!7.1\%\!\!
\\
\midrule
Overall Improvements   &        &         &         &         &           &         &         &         &         &         &         &\\
(M-FUSE vs. RAFT-3D) &  \!\!\!5.4\%\!\! &  \!\!\!15.9\%\!\! &   \!\!\!8.8\%\!\! &  \!\!\!14.7\%\!\! &  \!\!\!14.4\%\!\!  &  \!\!\!14.7\%\!\! &  \!\!\!21.5\%\!\! &  \!\!\!15.0\%\!\! &  \!\!\!19.3\%\!\! &  \!\!\!19.7\%\!\! &  \!\!\!10.8\%\!\! &  \!\!\!16.3\%\!\!
\\
\bottomrule
\end{tabular}
\end{center}
\end{table*}

\subsection{Supervision}
Our network predicts the scene flow as triplet $(u,v,\Delta d)$.
We compute the target disparity as $d' \!=\! d \!+\! \Delta d$ and supervise our training using a robustified sublinear L1 loss, reading
\begin{equation}
\mathcal{L}_\text{fuse} \!=\! \sum_\mathbf{x} \left( \alpha \cdot | d' \!-\! d'_\text{gt} | + | u \!-\! u_\text{gt} | + | v \!-\! v_\text{gt} | + \epsilon \right)^\gamma.
\end{equation}
In all our experiments, we chose $\epsilon=0.01$ and $\gamma=0.4$.
We additionally introduce a weighting parameter $\alpha$ to balance the loss components for disparity and optical flow and set it to 2.
%
% Finally, we also utilize the original RAFT-3D loss $\mathcal{L}_\text{R3D}$~\cite{Teed2021_RAFT3D} and apply it directly to the output of the forward baseline.
% The total loss then reads $\mathcal{L} = \mathcal{L}_\text{fuse} + \mu \cdot \mathcal{L}_\text{R3D}$, with $\mu=0.1$.
Finally, we also utilize the multi-iteration loss from RAFT-3D~\cite{Teed2021_RAFT3D} $\mathcal{L}_\text{R3D}$ that computes the $L1$ norm of optical flow and disparity change with
a per-iteration weight.
% with basis $\gamma = 0.9$.
We apply it directly to the output of the forward baseline and obtain a total loss of $\mathcal{L} = \mathcal{L}_\text{fuse} + \mu \cdot \mathcal{L}_\text{R3D}$, with $\mu=0.1$.

\section{Experiments}
We implemented our model in PyTorch~\cite{Paszke2019_pytorch} and initialized the fusion module's weights with the normal distributed initialization by He \etal~\cite{He2015_init} for convolutions, and zero-initialization for biases.
For the two-frame baseline, we used code provided by the authors~\cite{Teed2021_RAFT3D}.

\medskip
\noindent
\textbf{Training details.}
For the two-frame baseline, we followed the original training of RAFT-3D~\cite{Teed2021_RAFT3D} with 200K steps pretraining on FlyingThings3D~\cite{Mayer2016_SceneFlow} and 50K steps finetuning on the KITTI \emph{train} split~\cite{Menze2015_KITTI}; the latter using our improved disparity estimates~\cite{Cheng2020_LEAStereo}.
For training our multi-frame method, we initialized our shared forward and backward model with the pretrained two-frame baseline and also finetuned for 50K steps on the KITTI \emph{train} split -- this time, however, dividing the 50K steps in two stages.
% of 10K and 40K steps.
First, for 10K steps, we kept the parameters of the shared baseline models fixed in order to pretrain the fusion module.
Then, for the remaining 40K steps, we trained our entire model end-to-end.
Thereby, we used the Adam optimizer~\cite{Kingma2015_Adam} with the same linear-decay learning rate strategy~\cite{Smith2019_OneCycleLR} as RAFT-3D~\cite{Teed2021_RAFT3D}, employing maximum learning rates of $5\cdot 10^{-4}$ and $1\cdot 10^{-4}$ for the two finetuning stages.
During all stages, we trained on a single NVIDIA A100 GPU with batch size 4.
Moreover, we utilized spatial and photometric augmentations~\cite{Teed2021_RAFT3D} with crop size $256 \times 960$.
% Reason for R3D loss

\begin{figure*}
\centering
{
\tikzset{labelstyle/.style={anchor=north west, text=white, inner sep=1.5, text opacity=1, scale=0.66, yshift=-1, xshift=1, fill=black, opacity=0.6}}
\tikzset{imgstyle/.style={inner sep=0,anchor=north west,outer sep=0,draw=none,line width=0}}
\setlength\tabcolsep{1pt}
\begin{tabular}{cccccc}
&
\includegraphics[trim={0 0 0 2.85cm},clip,width=0.241\textwidth]{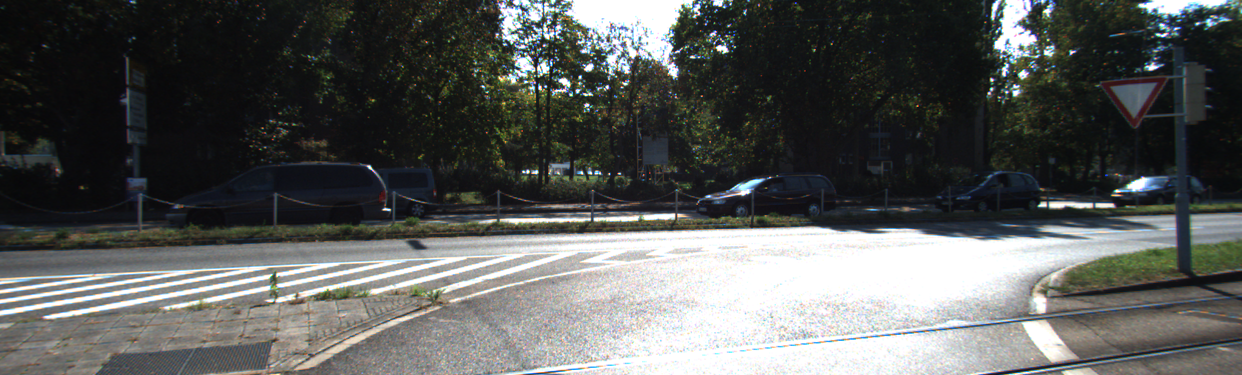}
&
\includegraphics[trim={0 0 0 2.85cm},clip,width=0.241\textwidth]{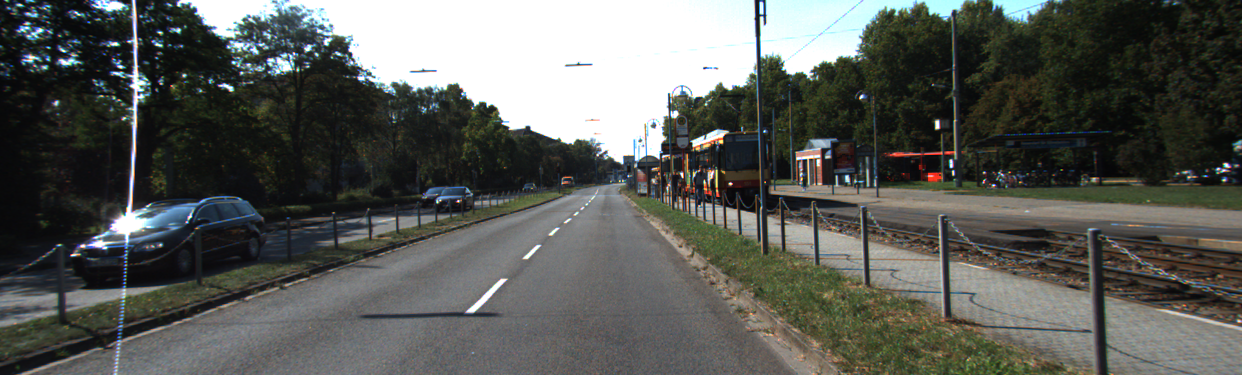}
&
\includegraphics[trim={0 0 0 2.85cm},clip,width=0.241\textwidth]{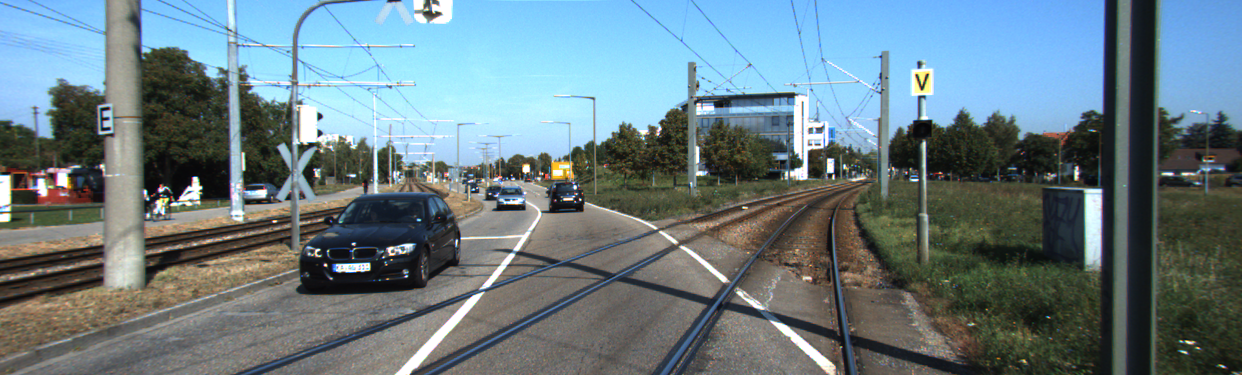}
&
\includegraphics[trim={0 0 0 2.85cm},clip,width=0.241\textwidth]{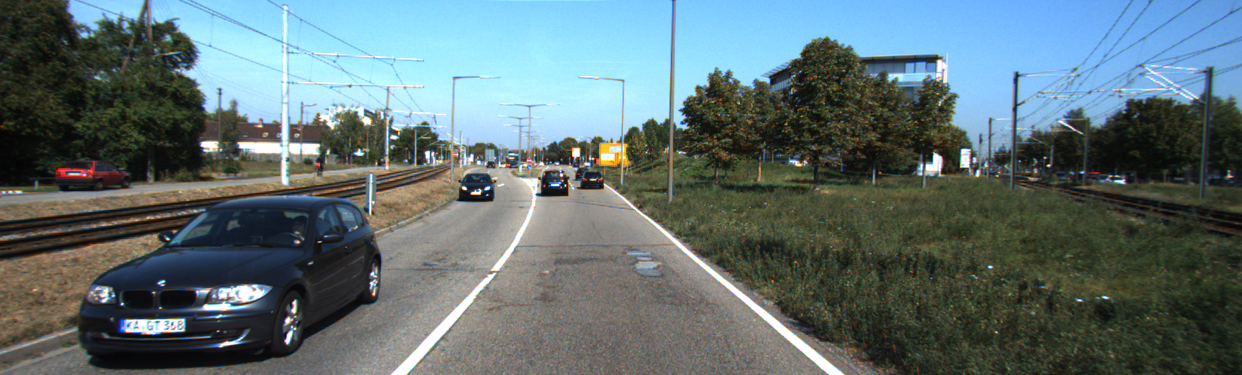}
\\[-1.9mm]
\begin{tikzpicture}
\draw[fill=none,draw=none] (0, 0) rectangle (0.4,1.03) node[pos=.5, inner sep=0] {\small \emph{D2}};
\end{tikzpicture}
&
\begin{tikzpicture}
\draw (0, 0) node[imgstyle] (img) {
\includegraphics[trim={0 0 0 2.85cm},clip,width=0.241\textwidth]{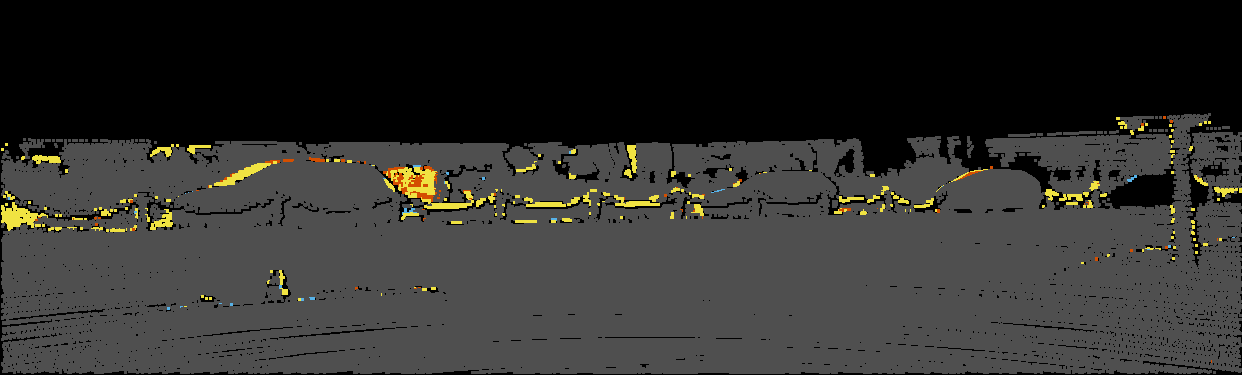}
};
\draw (img.north west) node[labelstyle] {Baseline: 1.49\;\; M-FUSE: 1.71\;\; -14.8\%};
\end{tikzpicture}
&
\begin{tikzpicture}
\draw (0, 0) node[imgstyle] (img) {
\includegraphics[trim={0 0 0 2.85cm},clip,width=0.241\textwidth]{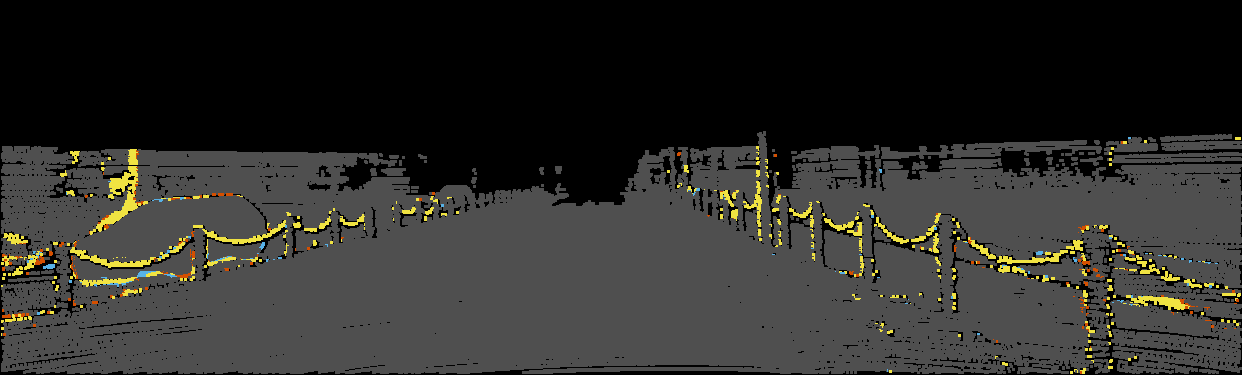}
};
\draw (img.north west) node[labelstyle] {Baseline: 2.14\;\; M-FUSE: 2.28\;\; -6.5\%};
\end{tikzpicture}
&
\begin{tikzpicture}
\draw (0, 0) node[imgstyle] (img) {
\includegraphics[trim={0 0 0 2.85cm},clip,width=0.241\textwidth]{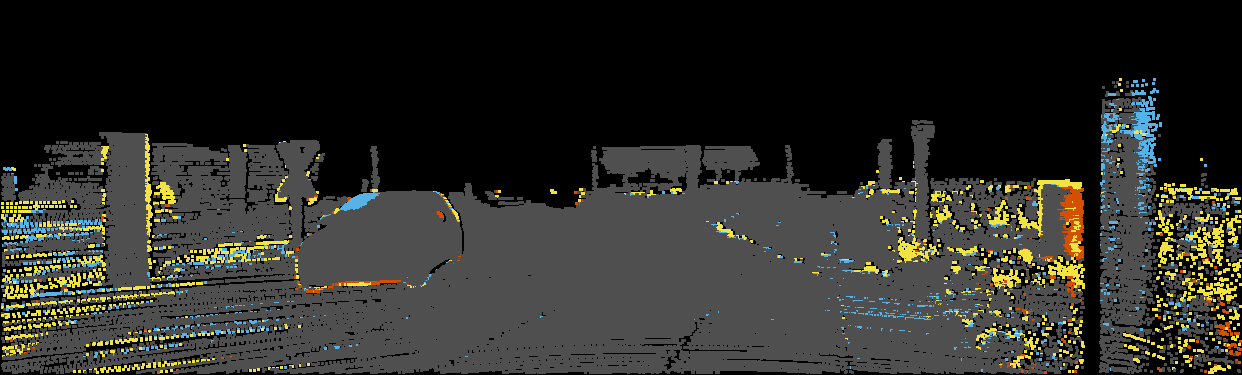}
};
\draw (img.north west) node[labelstyle] {Baseline: 6.18\;\; M-FUSE: 4.78\;\; +22.7\%};
\end{tikzpicture}
&
\begin{tikzpicture}
\draw (0, 0) node[imgstyle] (img) {
\includegraphics[trim={0 0 0 2.85cm},clip,width=0.241\textwidth]{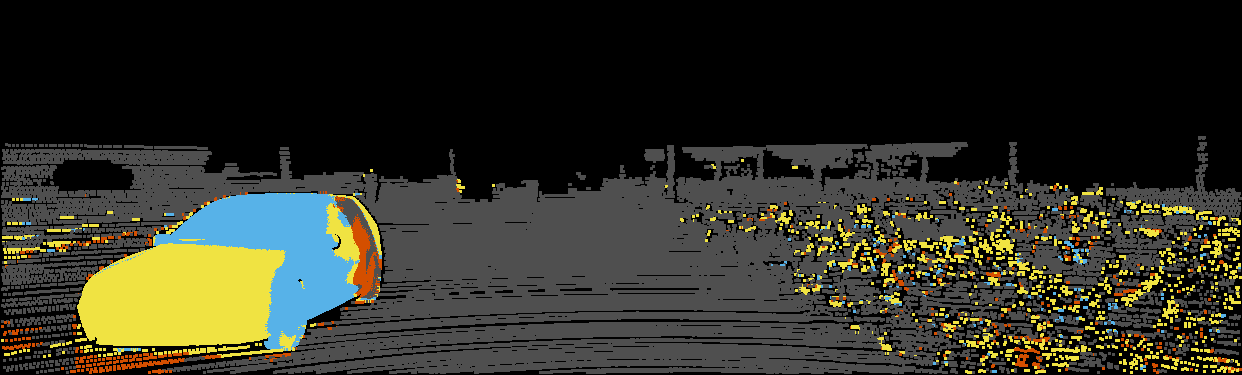}
};
\draw (img.north west) node[labelstyle] {Baseline: 47.81\,\; M-FUSE: 30.93\,\; +35.3\%};
\end{tikzpicture}
\\[-1.9mm]
\begin{tikzpicture}
\draw[fill=none,draw=none] (0, 0) rectangle (0.4,1.03) node[pos=.5, inner sep=0] {\small \emph{Fl}};
\end{tikzpicture}
&
\begin{tikzpicture}
\draw (0, 0) node[imgstyle] (img) {
\includegraphics[trim={0 0 0 2.85cm},clip,width=0.241\textwidth]{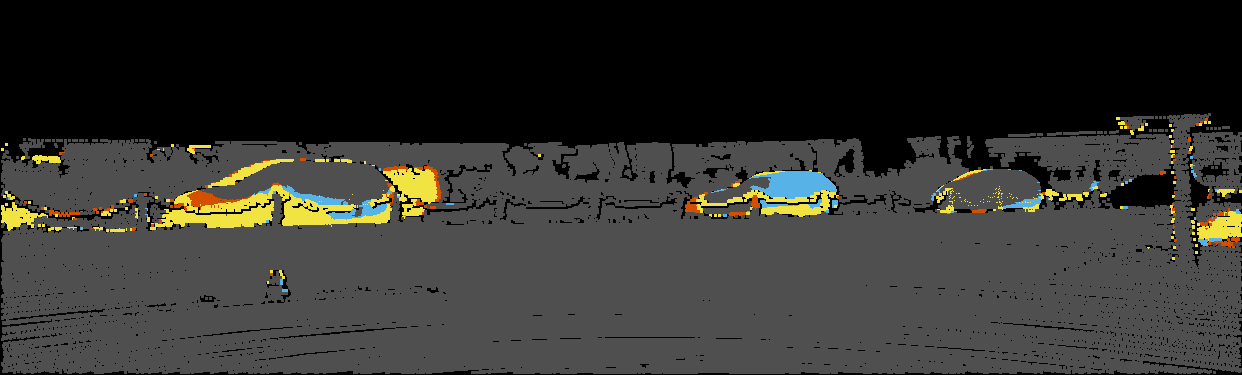}
};
\draw (img.north west) node[labelstyle] {Baseline: 6.91\;\; M-FUSE: 5.29\;\; +23.4\%};
\end{tikzpicture}
&
\begin{tikzpicture}
\draw (0, 0) node[imgstyle] (img) {
\includegraphics[trim={0 0 0 2.85cm},clip,width=0.241\textwidth]{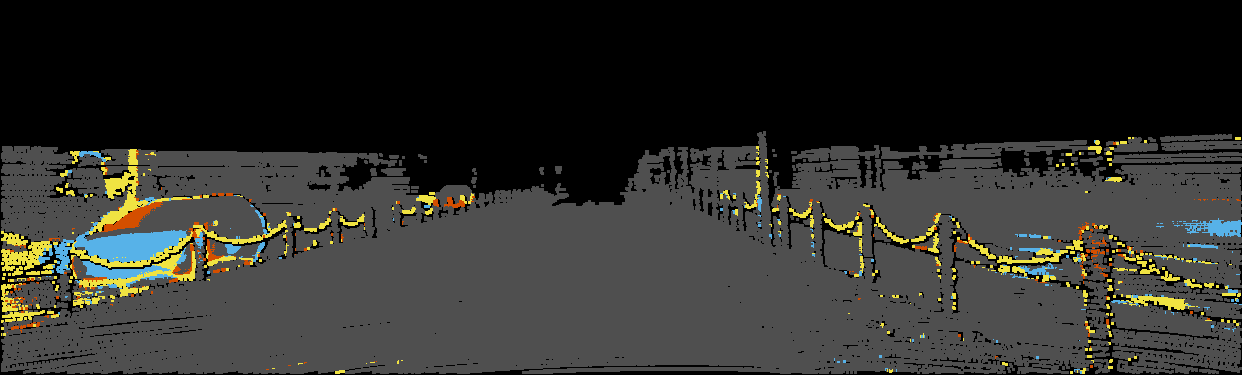}
};
\draw (img.north west) node[labelstyle] {Baseline: 6.47\;\; M-FUSE: 4.23\;\; +34.6\%};
\end{tikzpicture}
&
\begin{tikzpicture}
\draw (0, 0) node[imgstyle] (img) {
\includegraphics[trim={0 0 0 2.85cm},clip,width=0.241\textwidth]{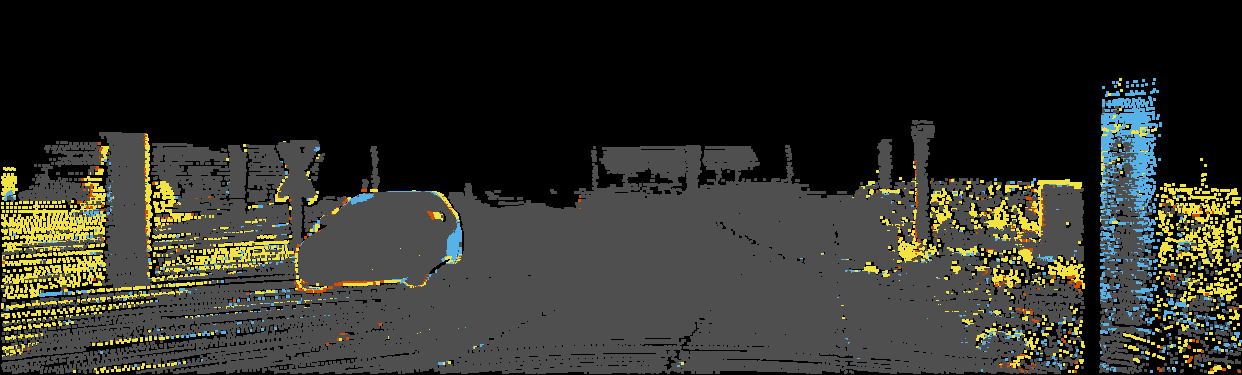}
};
\draw (img.north west) node[labelstyle] {Baseline: 8.52\;\; M-FUSE: 6.09\;\; +28.5\%};
\end{tikzpicture}
&
\begin{tikzpicture}
\draw (0, 0) node[imgstyle] (img) {
\includegraphics[trim={0 0 0 2.85cm},clip,width=0.241\textwidth]{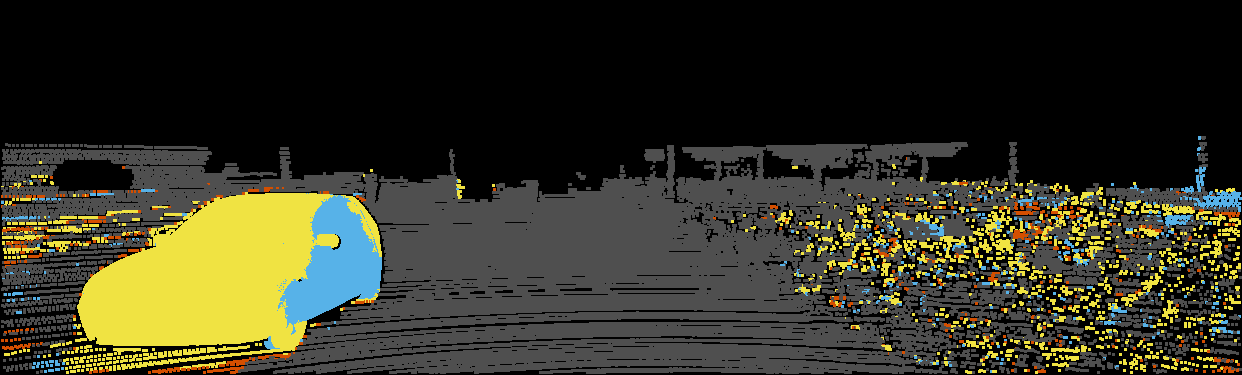}
};
\draw (img.north west) node[labelstyle] {Baseline: 52.36\,\; M-FUSE: 42.31\,\; +19.2\%};
\end{tikzpicture}
\\[-1.9mm]
\begin{tikzpicture}
\draw[fill=none,draw=none] (0, 0) rectangle (0.4,1.03) node[pos=.5, inner sep=0] {\small \emph{SF}};
\end{tikzpicture}
&
\begin{tikzpicture}
\draw (0, 0) node[imgstyle] (img) {
\includegraphics[trim={0 0 0 2.85cm},clip,width=0.241\textwidth]{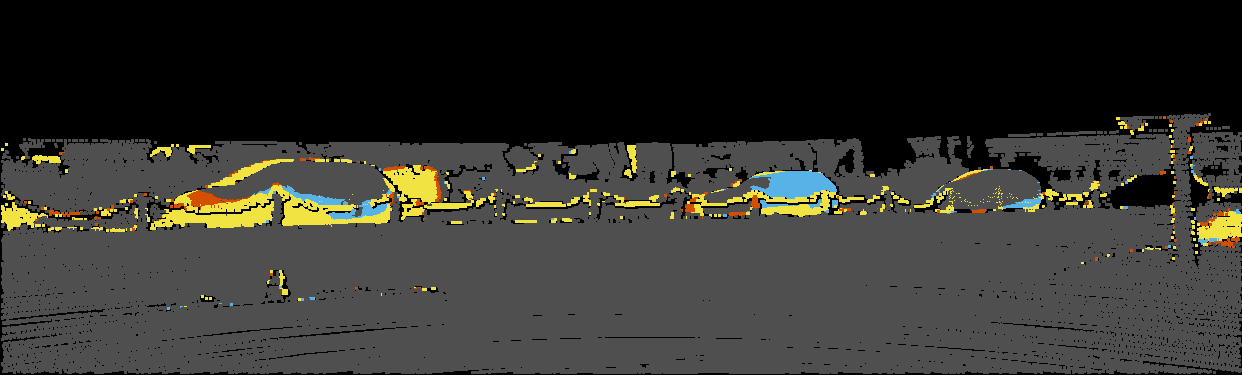}
};
\draw (img.north west) node[labelstyle] {Baseline: 7.55\;\; M-FUSE: 5.96\;\; +21.1\%};
\end{tikzpicture}
&
\begin{tikzpicture}
\draw (0, 0) node[imgstyle] (img) {
\includegraphics[trim={0 0 0 2.85cm},clip,width=0.241\textwidth]{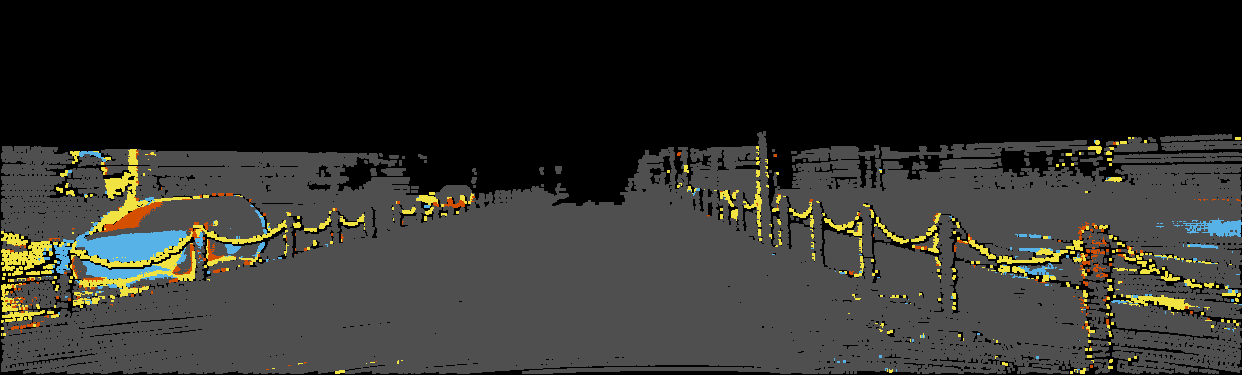}
};
\draw (img.north west) node[labelstyle] {Baseline: 6.64\;\; M-FUSE: 4.50\;\; +32.2\%};
\end{tikzpicture}
&
\begin{tikzpicture}
\draw (0, 0) node[imgstyle] (img) {
\includegraphics[trim={0 0 0 2.85cm},clip,width=0.241\textwidth]{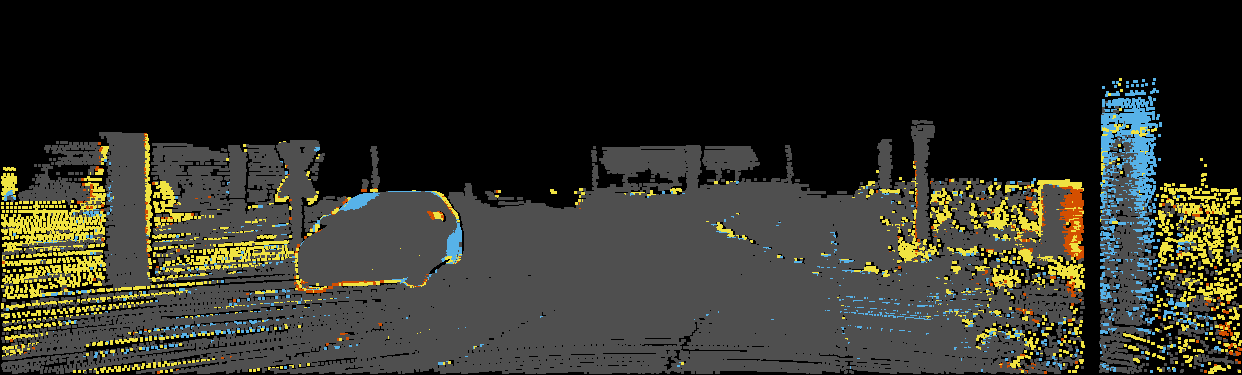}
};
\draw (img.north west) node[labelstyle] {Baseline: 9.70\;\; M-FUSE: 7.21\;\; +25.7\%};
\end{tikzpicture}
&
\begin{tikzpicture}
\draw (0, 0) node[imgstyle] (img) {
\includegraphics[trim={0 0 0 2.85cm},clip,width=0.241\textwidth]{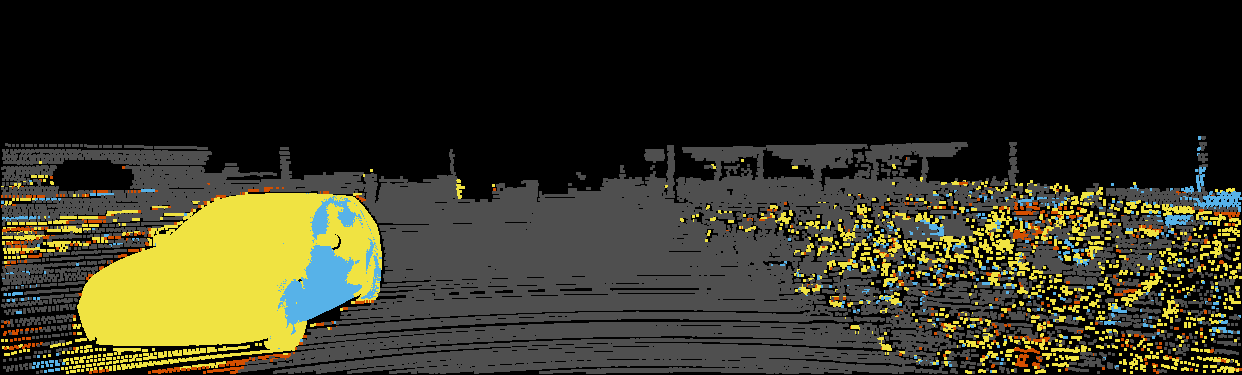}
};
\draw (img.north west) node[labelstyle] {Baseline: 52.77\,\; M-FUSE: 46.16\,\; +12.5\%};
\end{tikzpicture}
\end{tabular}
\caption{Qualitative evaluation of multi-frame improvements for four sequences of the KITTI benchmark (M-FUSE vs.\ baseline). \emph{From top to bottom:} reference frame, change in the outlier errors \emph{D2}, \emph{Fl} and \emph{SF}. \emph{Grey:} Both methods are inliers, \emph{blue:} M-FUSE is inlier and two-frame baseline is outlier, \emph{red:} two-frame baseline is inlier and M-FUSE is outlier, \emph{yellow:} both methods are outliers.}
\label{fig:multframe_improv}
}
\end{figure*}
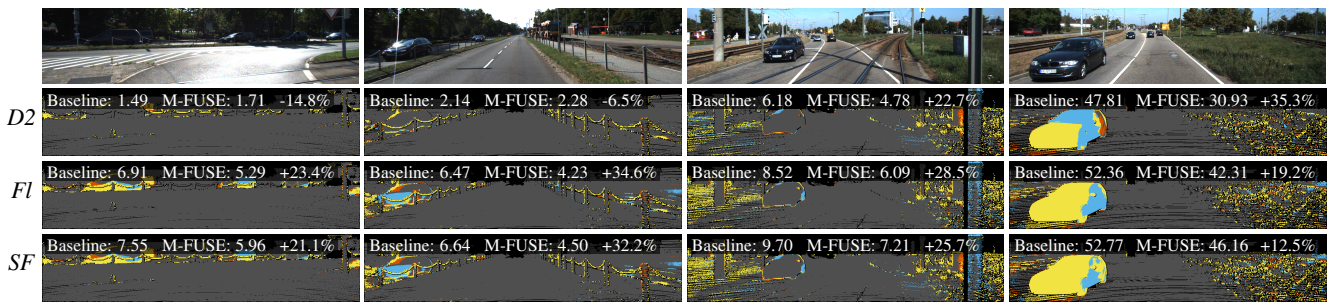

\begin{figure*}
\newcommand*\imgtrimtop{3cm}
\centering{
\tikzset{labelstyle/.style={anchor=north west, text=white, inner sep=2, text opacity=1, scale=0.7, yshift=-1, xshift=1, fill=black, opacity=0.6}}
\tikzset{imgstyle/.style={inner sep=0,anchor=north west,outer sep=0,draw=none,line width=0}}
\setlength\tabcolsep{1pt}
\begin{tabular}{ccccc}
\begin{tikzpicture}
\draw (0, 0) node[imgstyle] (img) {
\includegraphics[trim={0 0 0 \imgtrimtop},clip,width=0.197\textwidth]{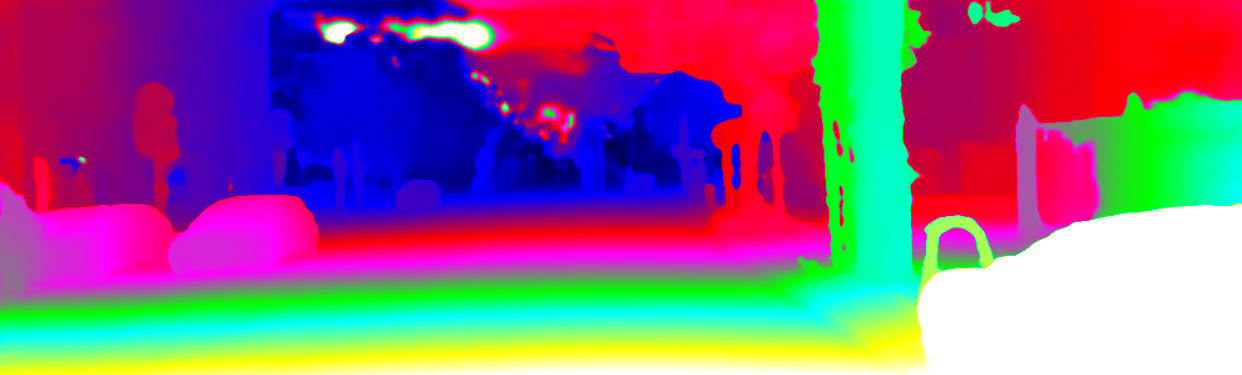}};
\draw (img.north west) node[labelstyle] {RigidMask+ISF};
\end{tikzpicture} &
\begin{tikzpicture}
\draw (0, 0) node[imgstyle] (img) {
\includegraphics[trim={0 0 0 \imgtrimtop},clip,width=0.197\textwidth]{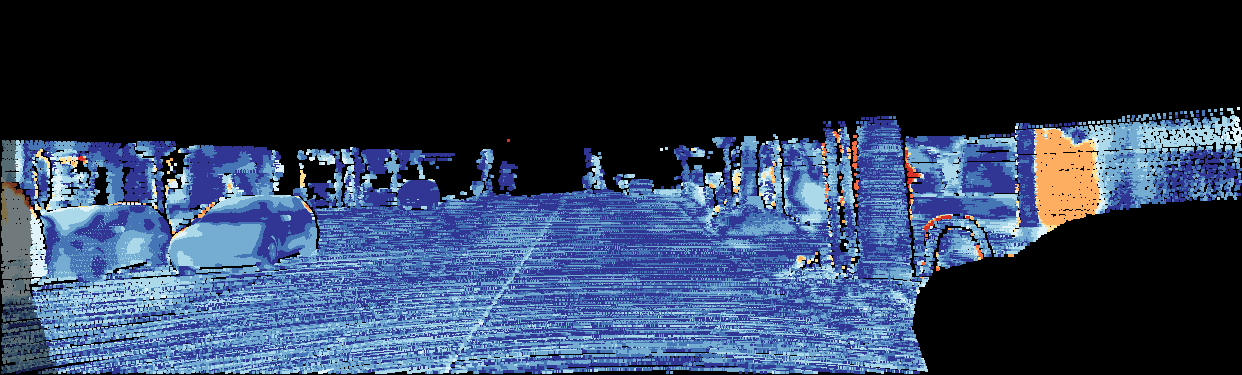}};
\draw (img.north west) node[labelstyle] {D2: 2.80};
\end{tikzpicture} &
\includegraphics[trim={0 0 0 \imgtrimtop},clip,width=0.197\textwidth]{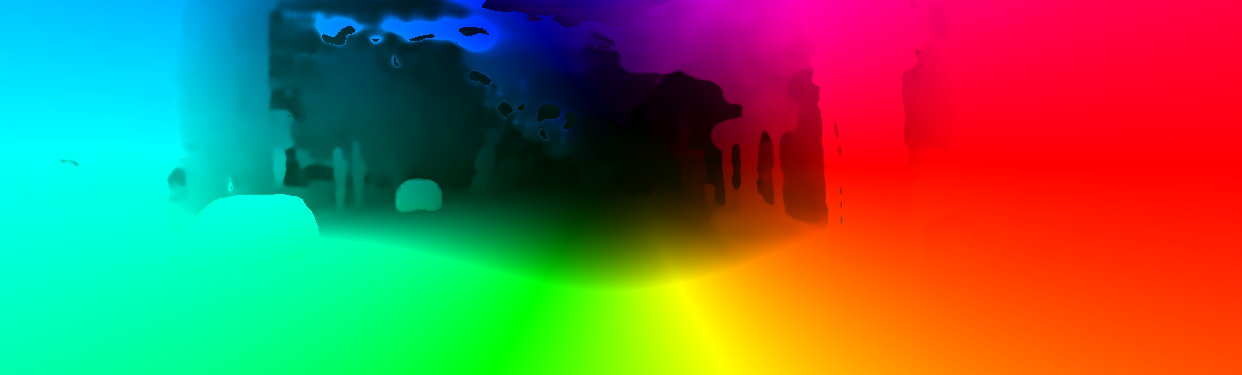} &
\begin{tikzpicture}
\draw (0, 0) node[imgstyle] (img) {
\includegraphics[trim={0 0 0 \imgtrimtop},clip,width=0.197\textwidth]{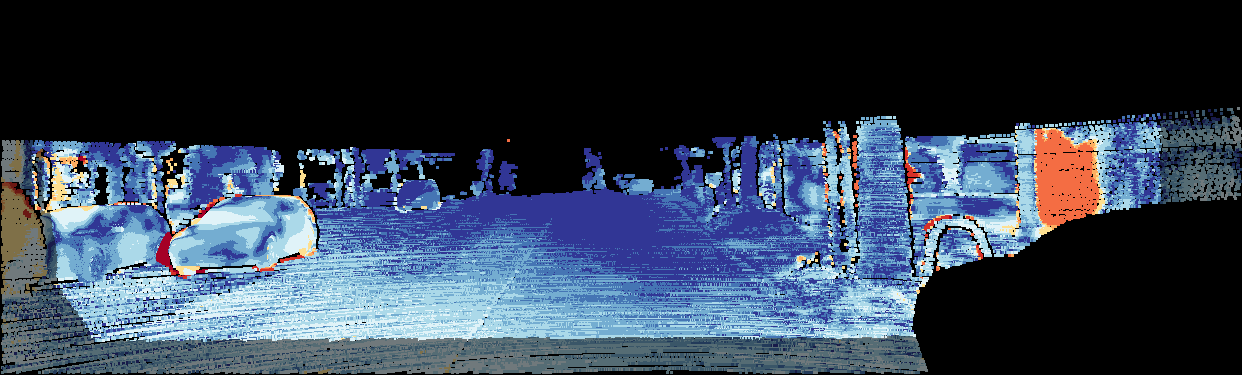}};
\draw (img.north west) node[labelstyle] {Fl: 3.96};
\end{tikzpicture} &
\begin{tikzpicture}
\draw (0, 0) node[imgstyle] (img) {
\includegraphics[trim={0 0 0 \imgtrimtop},clip,width=0.197\textwidth]{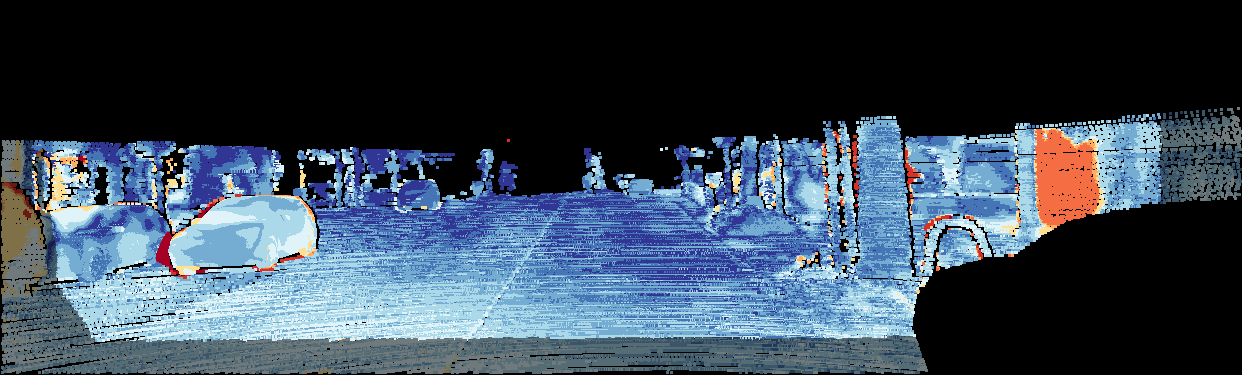}};
\draw (img.north west) node[labelstyle] {SF: 4.02};
\end{tikzpicture}
\\[-0.6mm]
\begin{tikzpicture}
\draw (0, 0) node[imgstyle] (img) {
\includegraphics[trim={0 0 0 \imgtrimtop},clip,width=0.197\textwidth]{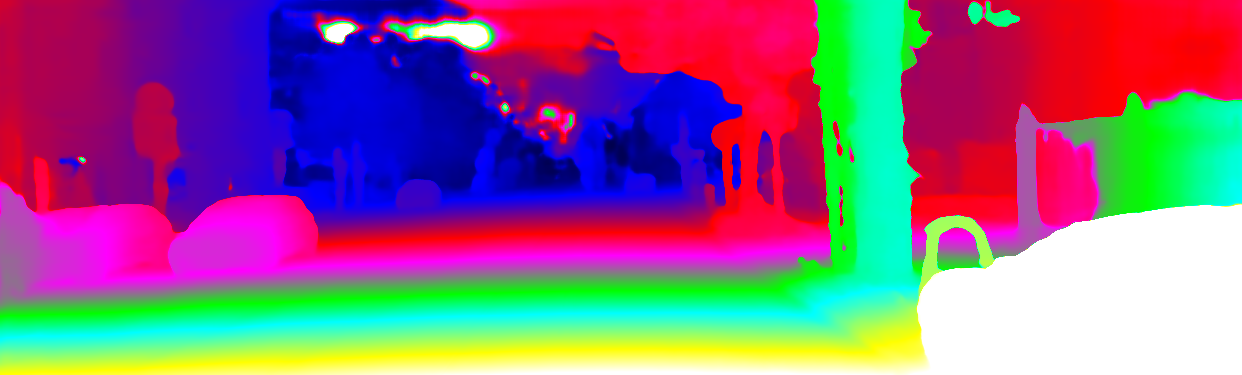}};
\draw (img.north west) node[labelstyle] {CamLiFlow};
\end{tikzpicture} &
\begin{tikzpicture}
\draw (0, 0) node[imgstyle] (img) {
\includegraphics[trim={0 0 0 \imgtrimtop},clip,width=0.197\textwidth]{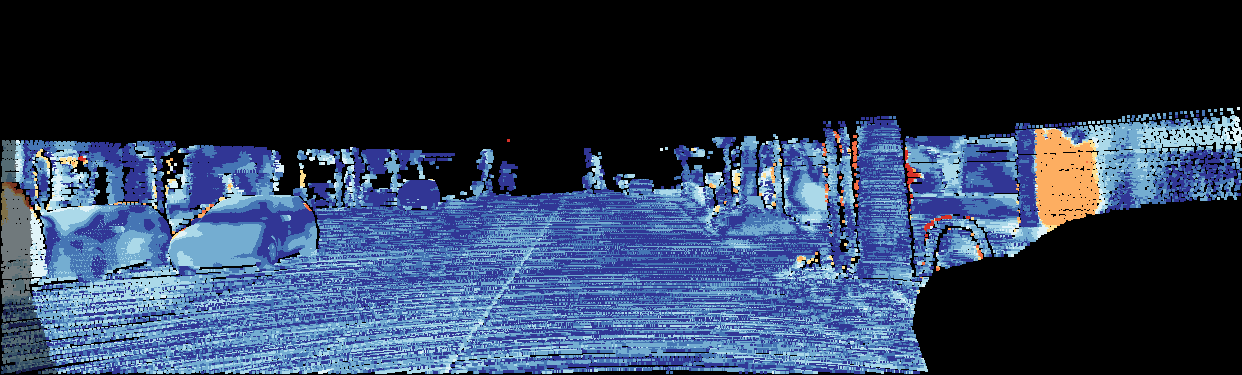}};
\draw (img.north west) node[labelstyle] {D2: 2.38};
\end{tikzpicture} &
\includegraphics[trim={0 0 0 \imgtrimtop},clip,width=0.197\textwidth]{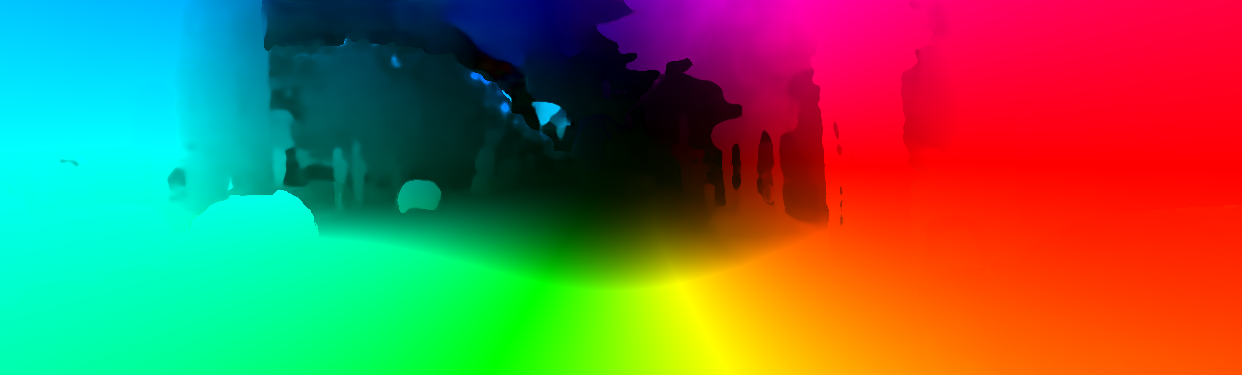} &
\begin{tikzpicture}
\draw (0, 0) node[imgstyle] (img) {
\includegraphics[trim={0 0 0 \imgtrimtop},clip,width=0.197\textwidth]{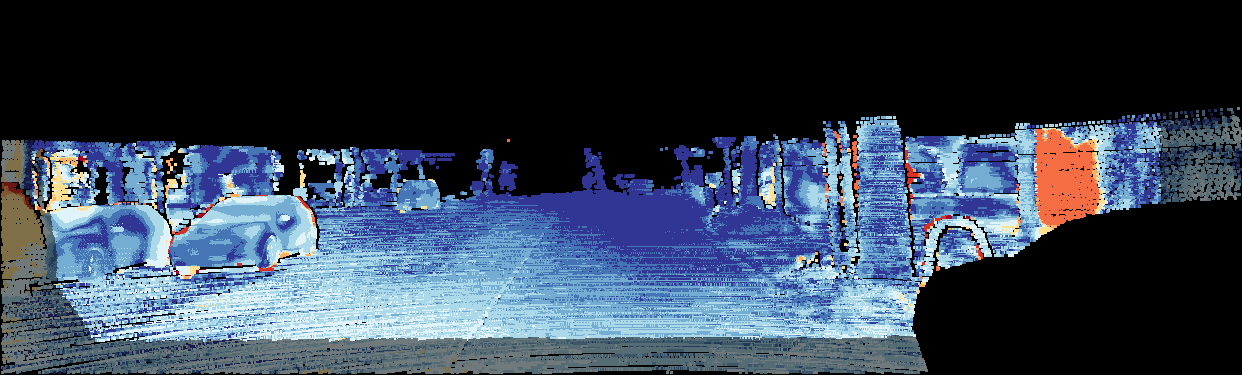}};
\draw (img.north west) node[labelstyle] {Fl: 3.64};
\end{tikzpicture} &
\begin{tikzpicture}
\draw (0, 0) node[imgstyle] (img) {
\includegraphics[trim={0 0 0 \imgtrimtop},clip,width=0.197\textwidth]{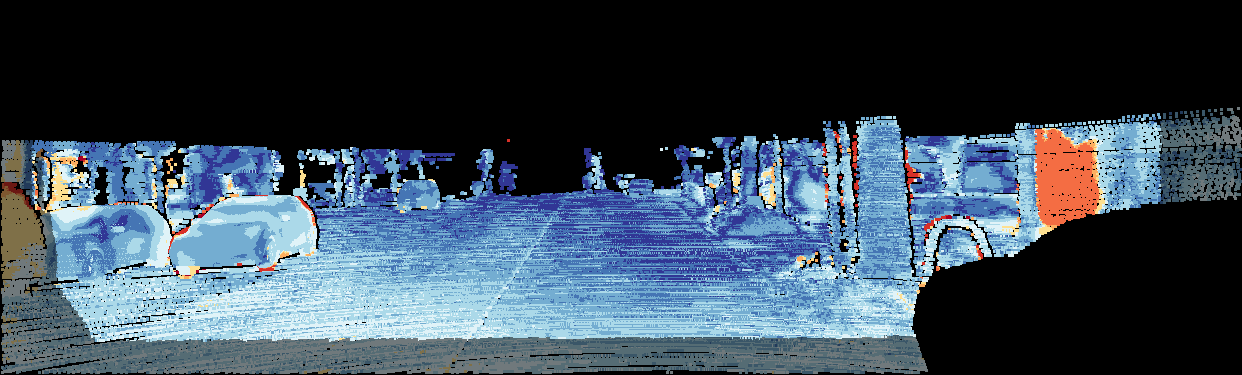}};
\draw (img.north west) node[labelstyle] {SF: 3.72};
\end{tikzpicture}
\\[-0.6mm]
\begin{tikzpicture}
\draw (0, 0) node[imgstyle] (img) {
\includegraphics[trim={0 0 0 \imgtrimtop},clip,width=0.197\textwidth]{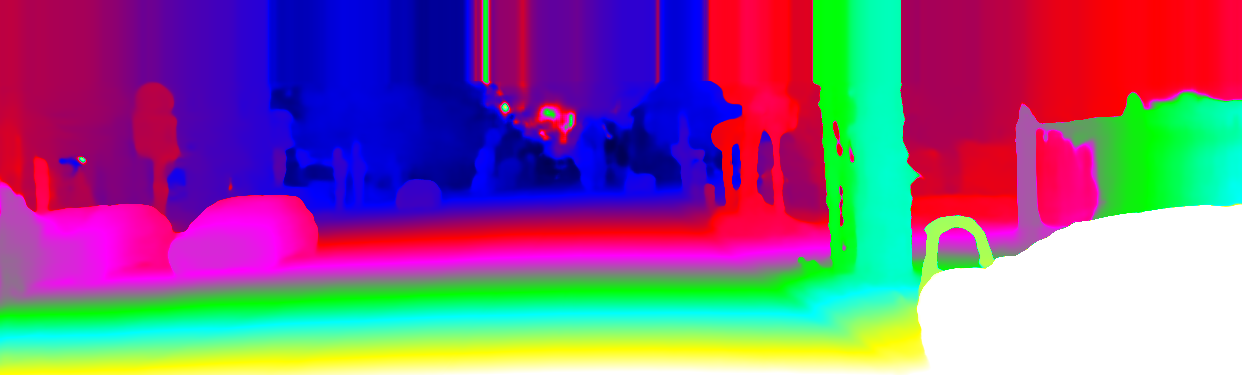}};
\draw (img.north west) node[labelstyle] {RAFT-3D};
\end{tikzpicture} &
\begin{tikzpicture}
\draw (0, 0) node[imgstyle] (img) {
\includegraphics[trim={0 0 0 \imgtrimtop},clip,width=0.197\textwidth]{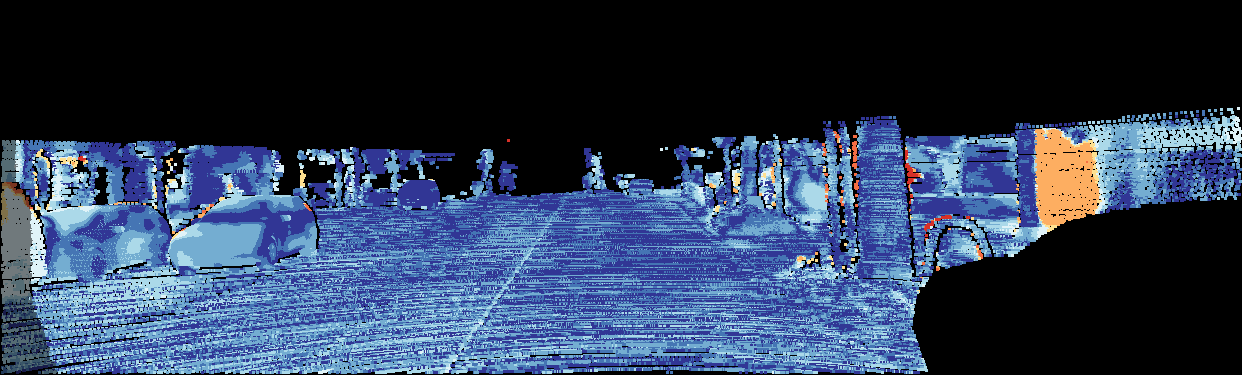}};
\draw (img.north west) node[labelstyle] {D2: 2.61};
\end{tikzpicture} &
\includegraphics[trim={0 0 0 \imgtrimtop},clip,width=0.197\textwidth]{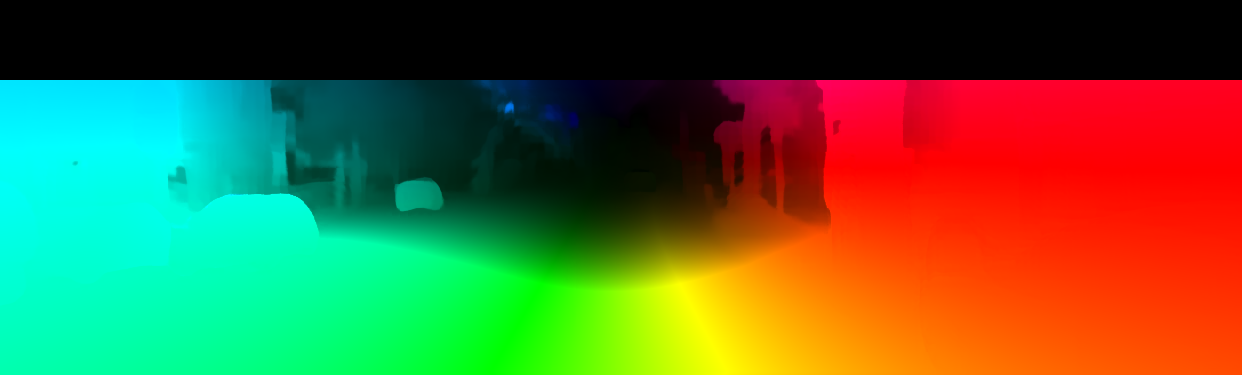} &
\begin{tikzpicture}
\draw (0, 0) node[imgstyle] (img) {
\includegraphics[trim={0 0 0 \imgtrimtop},clip,width=0.197\textwidth]{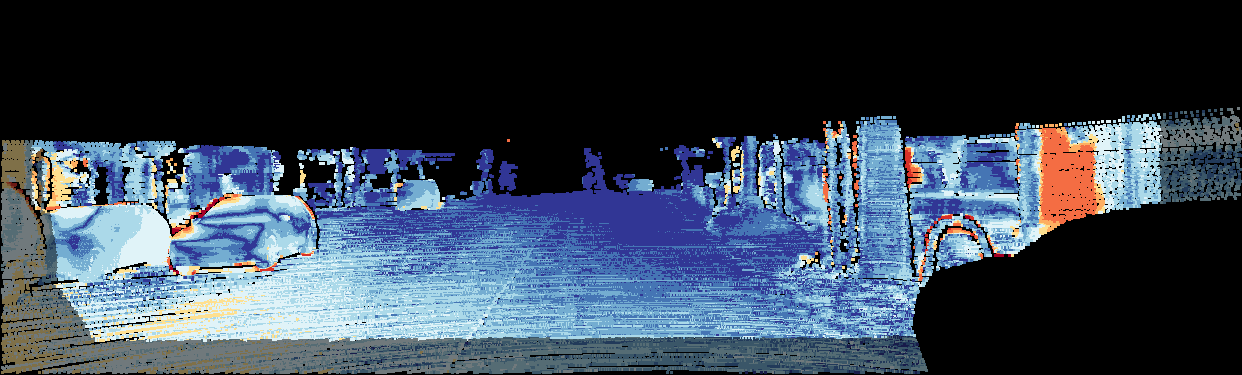}};
\draw (img.north west) node[labelstyle] {Fl: 4.95};
\end{tikzpicture} &
\begin{tikzpicture}
\draw (0, 0) node[imgstyle] (img) {
\includegraphics[trim={0 0 0 \imgtrimtop},clip,width=0.197\textwidth]{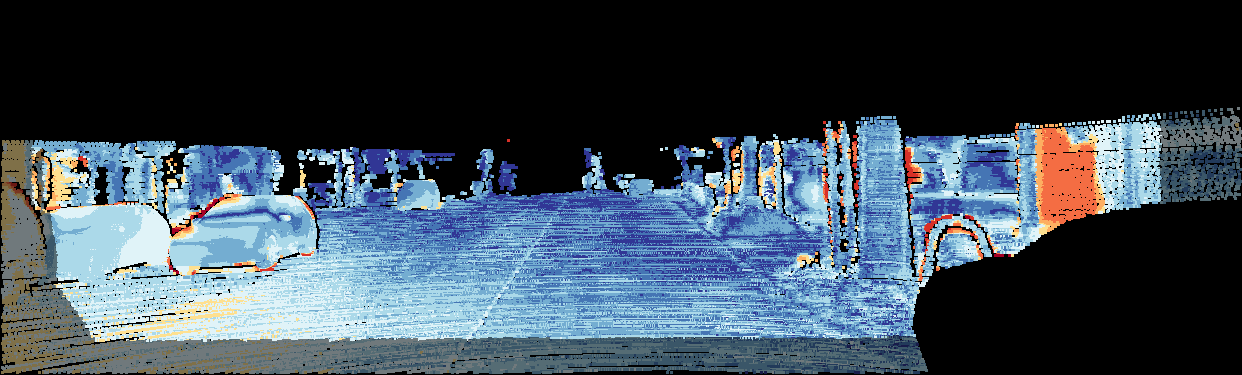}};
\draw (img.north west) node[labelstyle] {SF: 5.54};
\end{tikzpicture}
\\[-0.6mm]
\begin{tikzpicture}
\draw (0, 0) node[imgstyle] (img) {
\includegraphics[trim={0 0 0 \imgtrimtop},clip,width=0.197\textwidth]{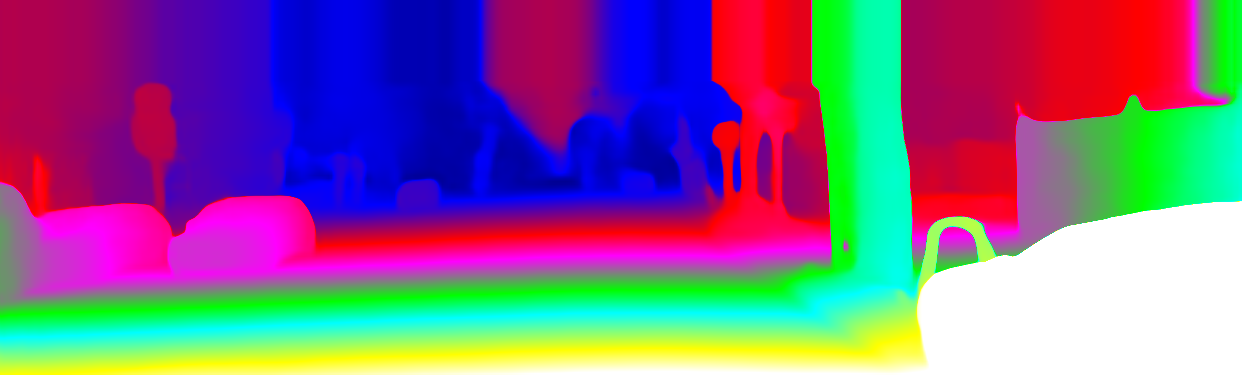}};
\draw (img.north west) node[labelstyle] {M-FUSE};
\end{tikzpicture} &
\begin{tikzpicture}
\draw (0, 0) node[imgstyle] (img) {
\includegraphics[trim={0 0 0 \imgtrimtop},clip,width=0.197\textwidth]{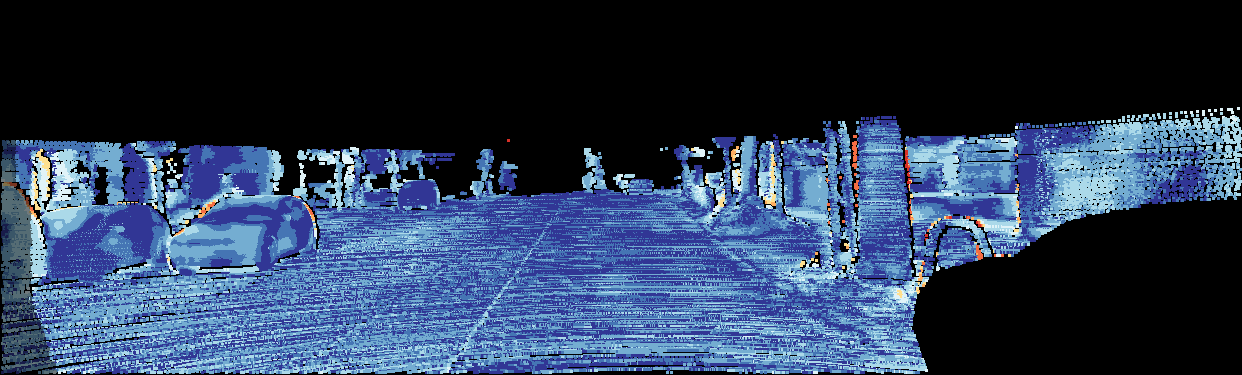}};
\draw (img.north west) node[labelstyle] {D2: 0.60};
\end{tikzpicture} &
\includegraphics[trim={0 0 0 \imgtrimtop},clip,width=0.197\textwidth]{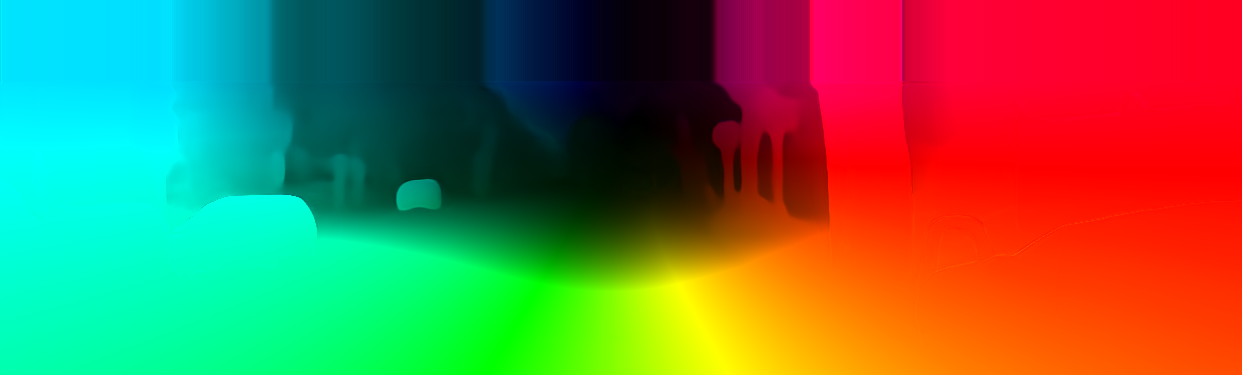} &
\begin{tikzpicture}
\draw (0, 0) node[imgstyle] (img) {
\includegraphics[trim={0 0 0 \imgtrimtop},clip,width=0.197\textwidth]{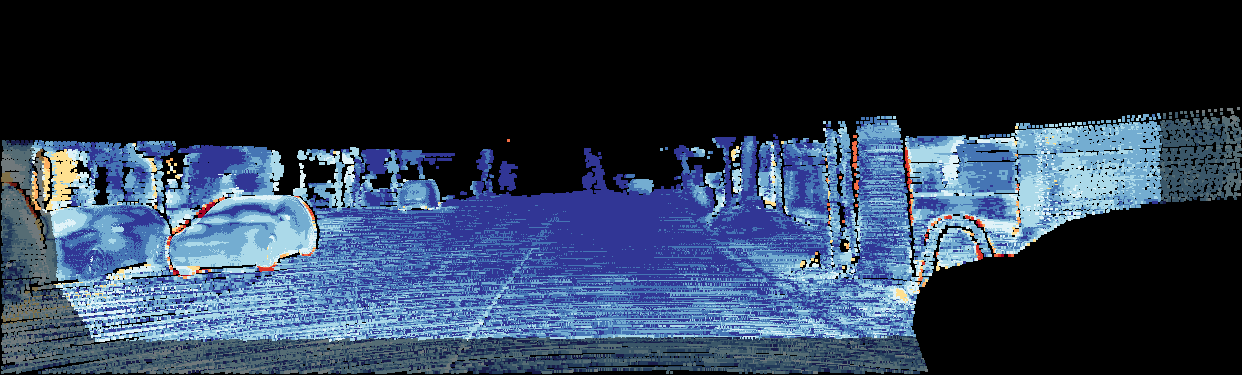}};
\draw (img.north west) node[labelstyle] {Fl: 1.15};
\end{tikzpicture} &
\begin{tikzpicture}
\draw (0, 0) node[imgstyle] (img) {
\includegraphics[trim={0 0 0 \imgtrimtop},clip,width=0.197\textwidth]{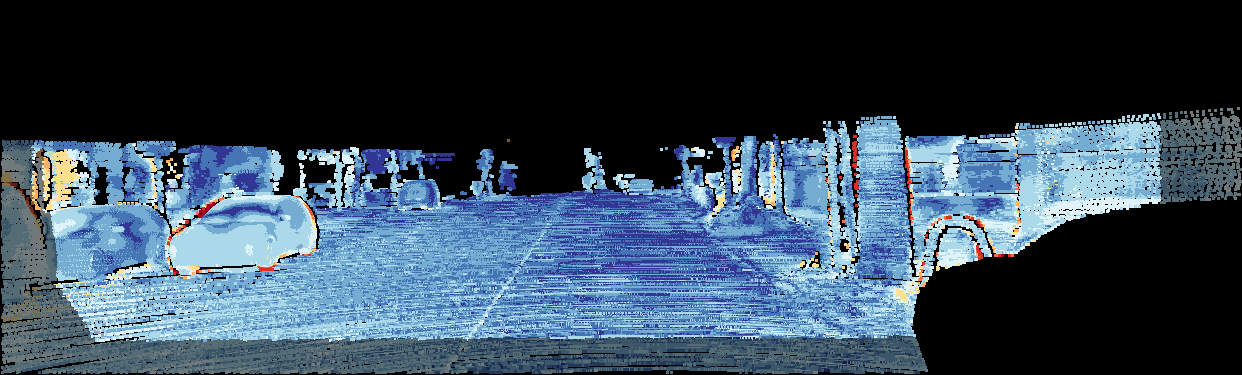}};
\draw (img.north west) node[labelstyle] {SF: 1.31};
\end{tikzpicture}
\\
\begin{tikzpicture}
\draw (0, 0) node[imgstyle] (img) {
\includegraphics[trim={0 0 0 \imgtrimtop},clip,width=0.197\textwidth]{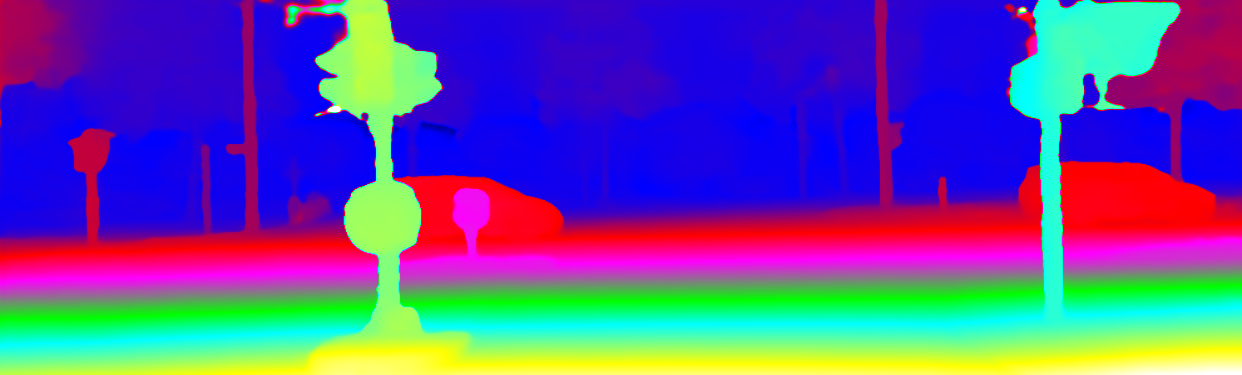}};
\draw (img.north west) node[labelstyle] {RigidMask+ISF};
\end{tikzpicture} &
\begin{tikzpicture}
\draw (0, 0) node[imgstyle] (img) {
\includegraphics[trim={0 0 0 \imgtrimtop},clip,width=0.197\textwidth]{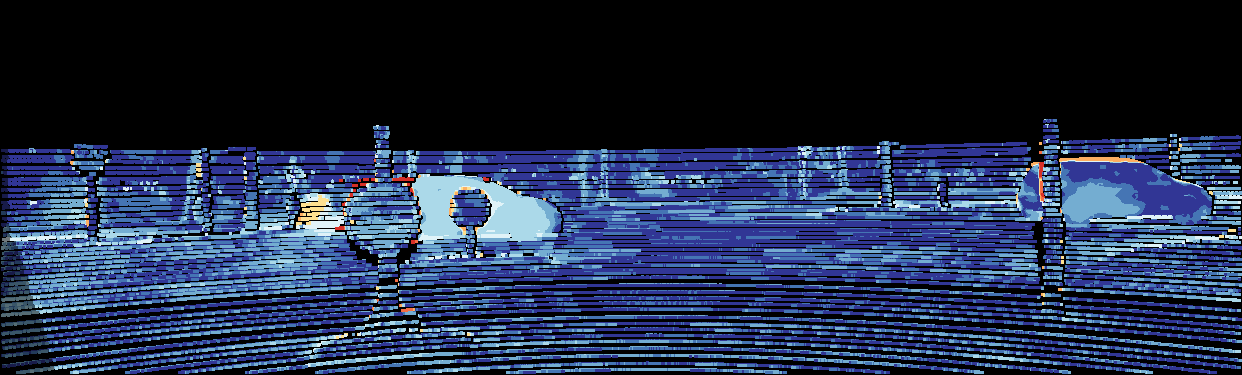}};
\draw (img.north west) node[labelstyle] {D2: 1.20};
\end{tikzpicture} &
\includegraphics[trim={0 0 0 \imgtrimtop},clip,width=0.197\textwidth]{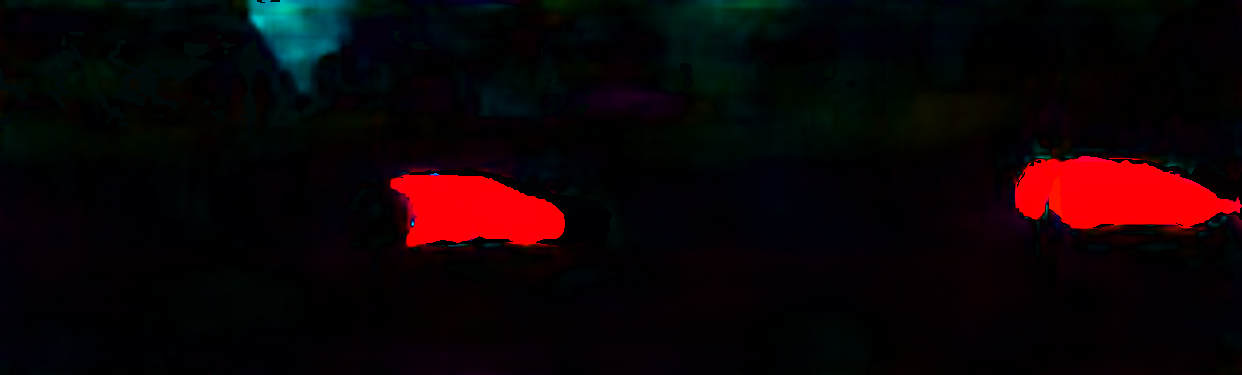} &
\begin{tikzpicture}
\draw (0, 0) node[imgstyle] (img) {
\includegraphics[trim={0 0 0 \imgtrimtop},clip,width=0.197\textwidth]{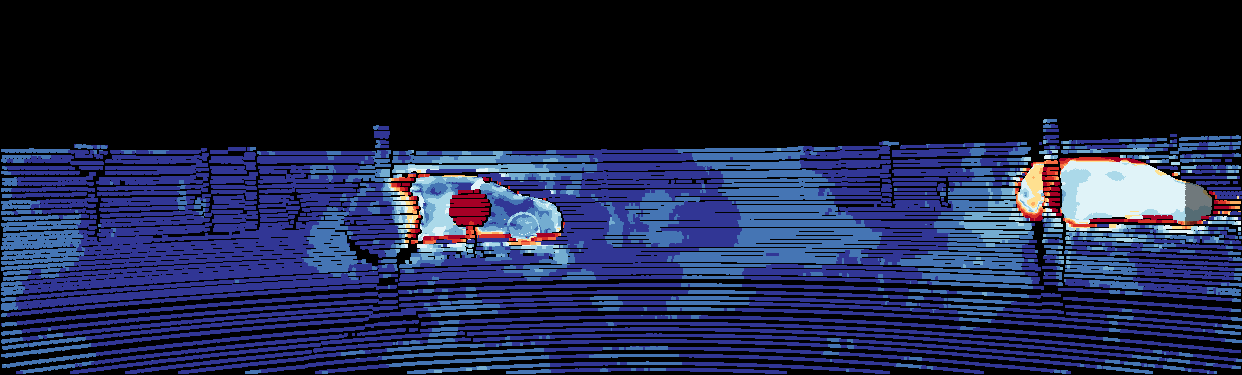}};
\draw (img.north west) node[labelstyle] {Fl: 3.60};
\end{tikzpicture} &
\begin{tikzpicture}
\draw (0, 0) node[imgstyle] (img) {
\includegraphics[trim={0 0 0 \imgtrimtop},clip,width=0.197\textwidth]{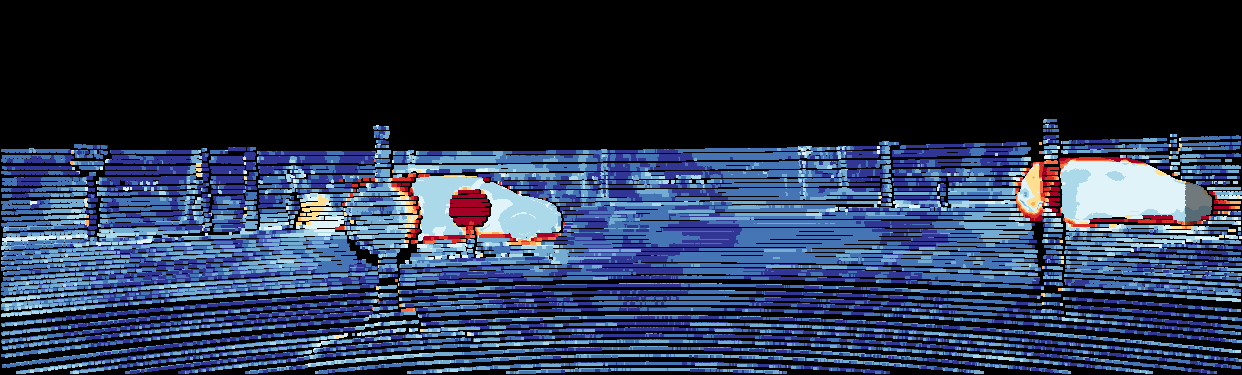}};
\draw (img.north west) node[labelstyle] {SF: 4.08};
\end{tikzpicture}
\\[-0.6mm]
\begin{tikzpicture}
\draw (0, 0) node[imgstyle] (img) {
\includegraphics[trim={0 0 0 \imgtrimtop},clip,width=0.197\textwidth]{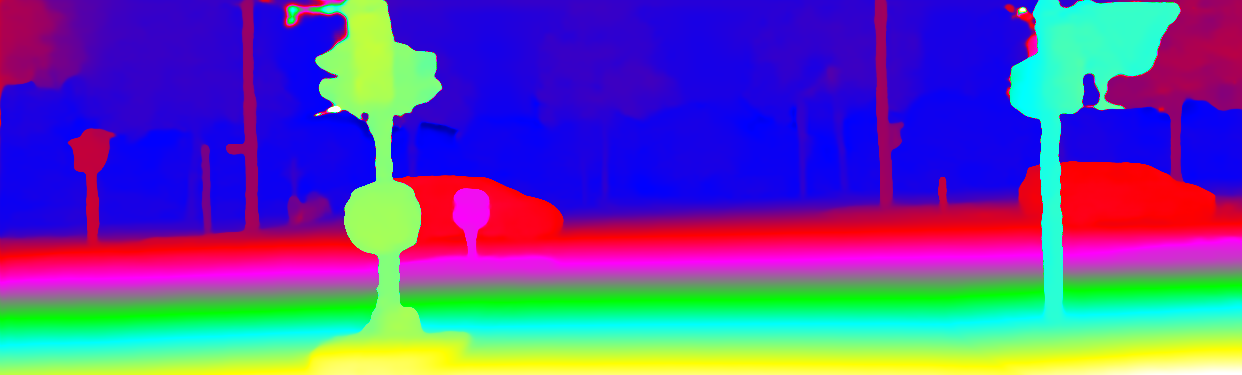}};
\draw (img.north west) node[labelstyle] {CamLiFlow};
\end{tikzpicture} &
\begin{tikzpicture}
\draw (0, 0) node[imgstyle] (img) {
\includegraphics[trim={0 0 0 \imgtrimtop},clip,width=0.197\textwidth]{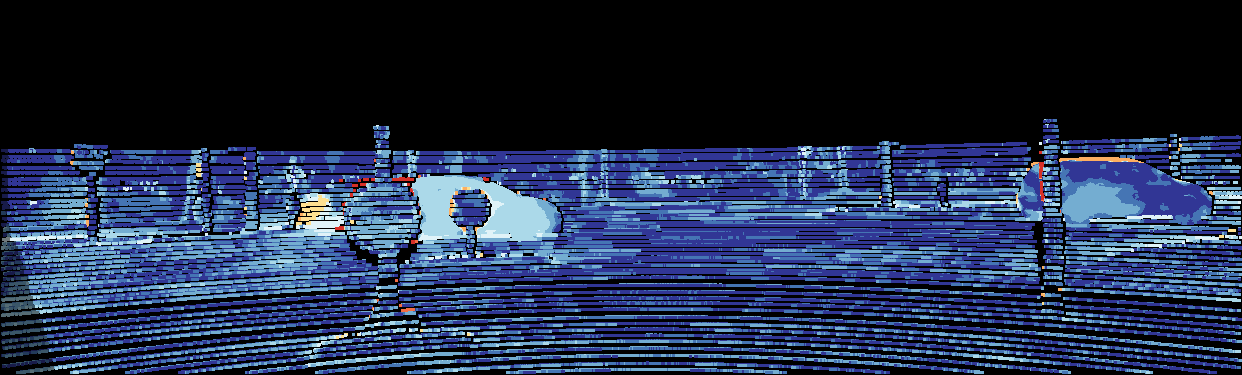}};
\draw (img.north west) node[labelstyle] {D2: 1.00};
\end{tikzpicture} &
\includegraphics[trim={0 0 0 \imgtrimtop},clip,width=0.197\textwidth]{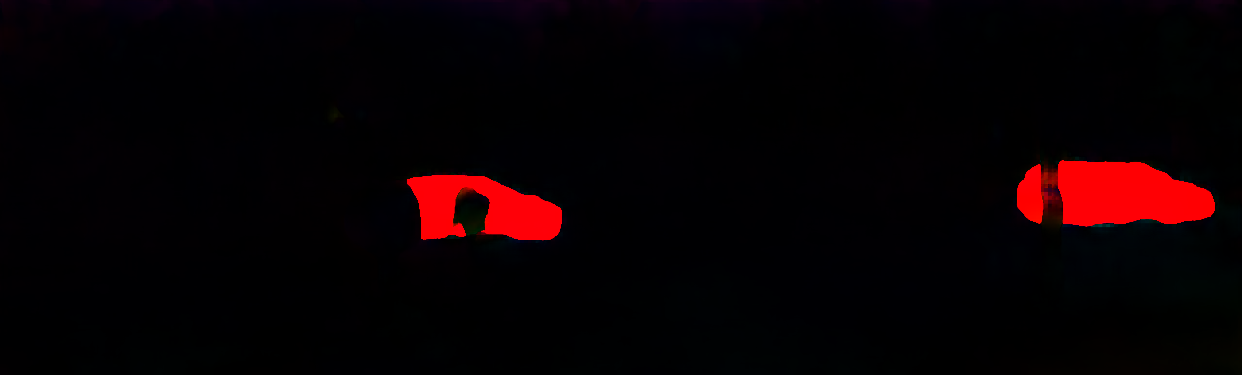} &
\begin{tikzpicture}
\draw (0, 0) node[imgstyle] (img) {
\includegraphics[trim={0 0 0 \imgtrimtop},clip,width=0.197\textwidth]{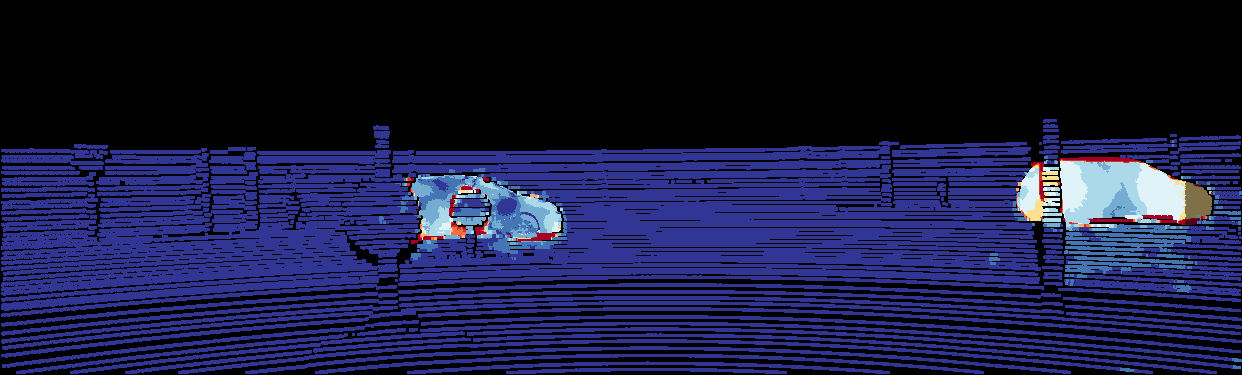}};
\draw (img.north west) node[labelstyle] {Fl: 3.03};
\end{tikzpicture} &
\begin{tikzpicture}
\draw (0, 0) node[imgstyle] (img) {
\includegraphics[trim={0 0 0 \imgtrimtop},clip,width=0.197\textwidth]{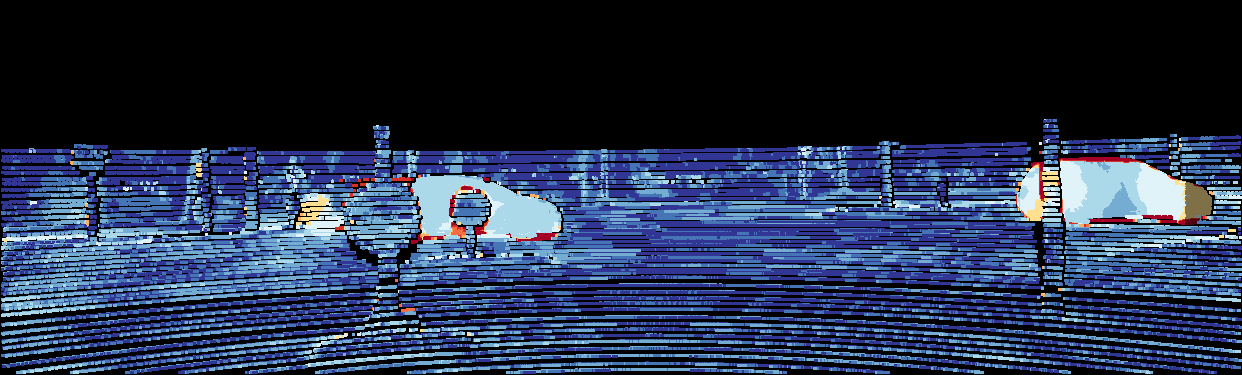}};
\draw (img.north west) node[labelstyle] {SF: 3.42};
\end{tikzpicture}
\\[-0.6mm]
\begin{tikzpicture}
\draw (0, 0) node[imgstyle] (img) {
\includegraphics[trim={0 0 0 \imgtrimtop},clip,width=0.197\textwidth]{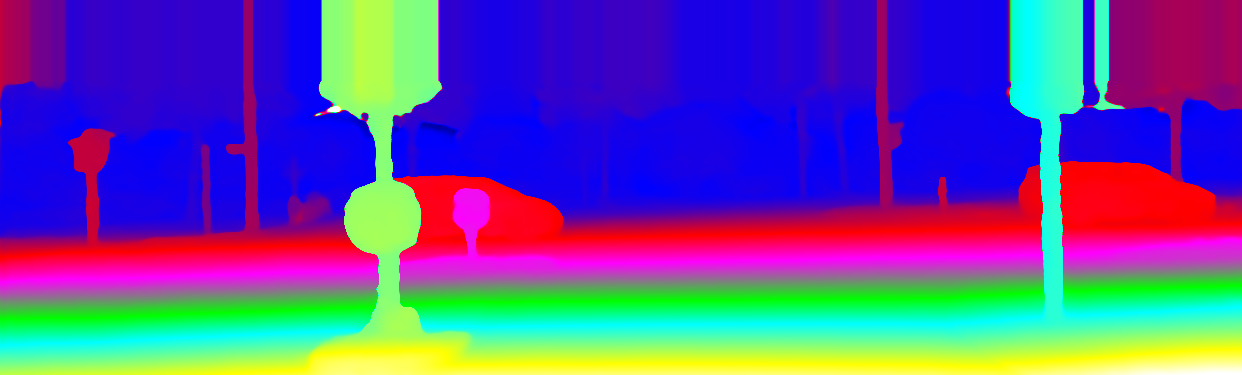}};
\draw (img.north west) node[labelstyle] {RAFT-3D};
\end{tikzpicture} &
\begin{tikzpicture}
\draw (0, 0) node[imgstyle] (img) {
\includegraphics[trim={0 0 0 \imgtrimtop},clip,width=0.197\textwidth]{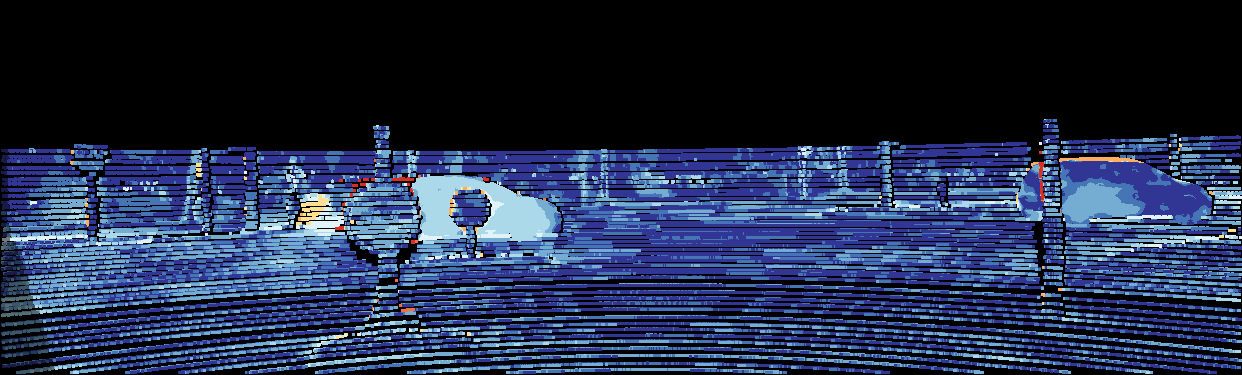}};
\draw (img.north west) node[labelstyle] {D2: 1.07};
\end{tikzpicture} &
\includegraphics[trim={0 0 0 \imgtrimtop},clip,width=0.197\textwidth]{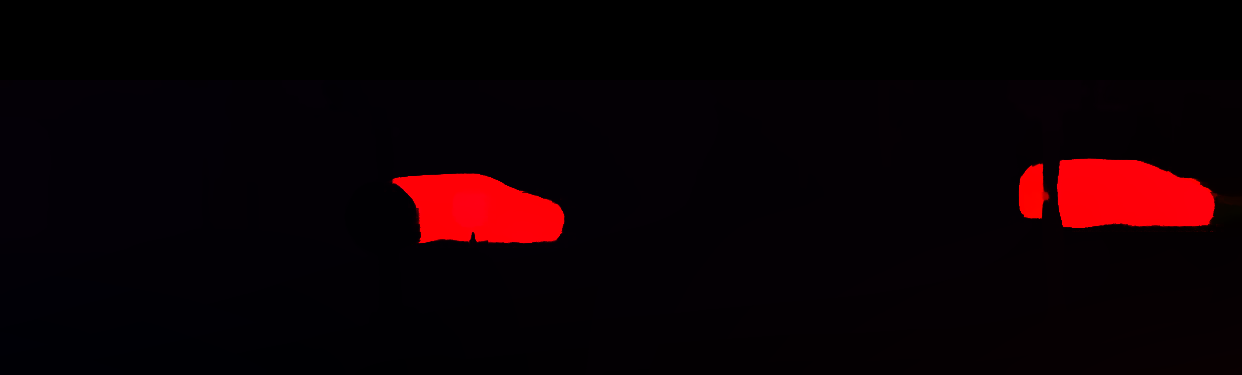} &
\begin{tikzpicture}
\draw (0, 0) node[imgstyle] (img) {
\includegraphics[trim={0 0 0 \imgtrimtop},clip,width=0.197\textwidth]{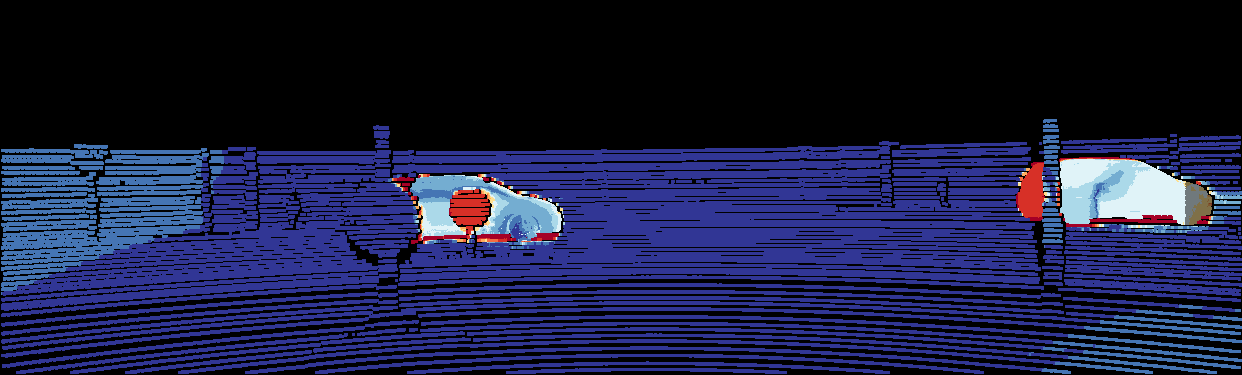}};
\draw (img.north west) node[labelstyle] {Fl: 3.64};
\end{tikzpicture} &
\begin{tikzpicture}
\draw (0, 0) node[imgstyle] (img) {
\includegraphics[trim={0 0 0 \imgtrimtop},clip,width=0.197\textwidth]{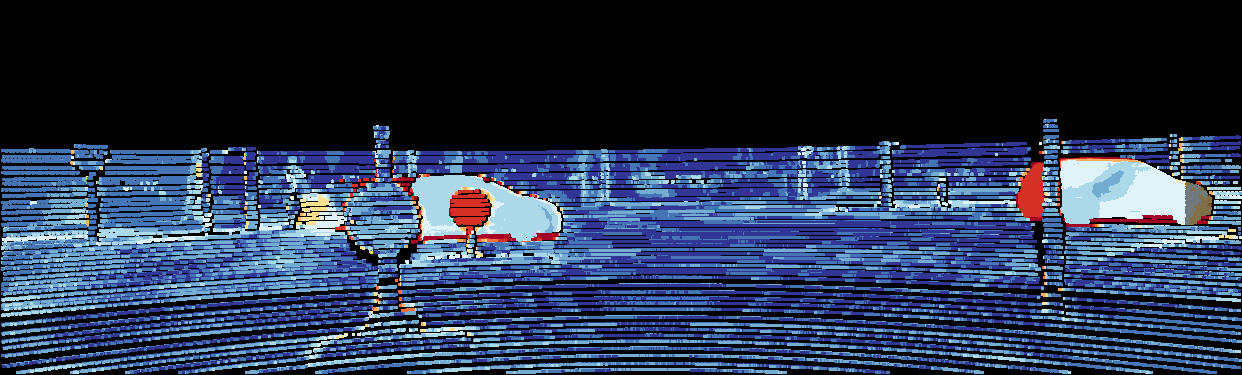}};
\draw (img.north west) node[labelstyle] {SF: 4.43};
\end{tikzpicture}
\\[-0.6mm]
\begin{tikzpicture}
\draw (0, 0) node[imgstyle] (img) {
\includegraphics[trim={0 0 0 \imgtrimtop},clip,width=0.197\textwidth]{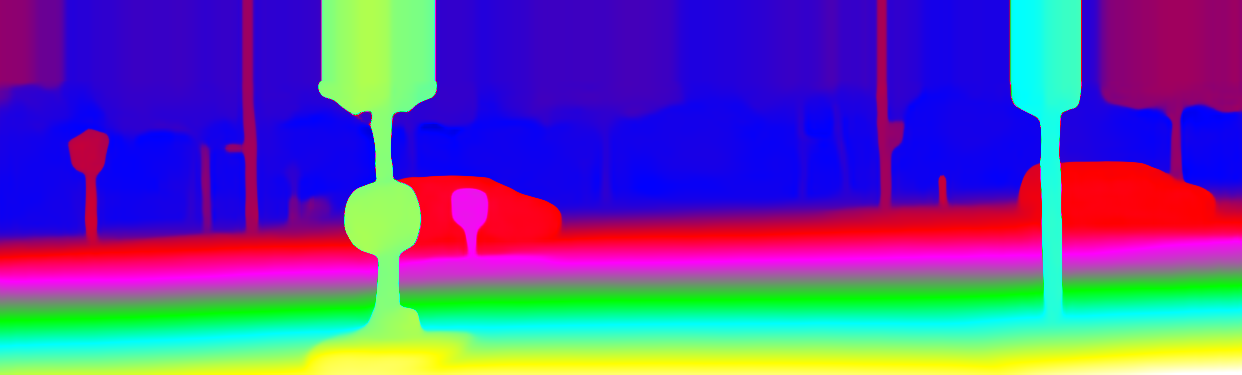}};
\draw (img.north west) node[labelstyle] {M-FUSE};
\end{tikzpicture} &
\begin{tikzpicture}
\draw (0, 0) node[imgstyle] (img) {
\includegraphics[trim={0 0 0 \imgtrimtop},clip,width=0.197\textwidth]{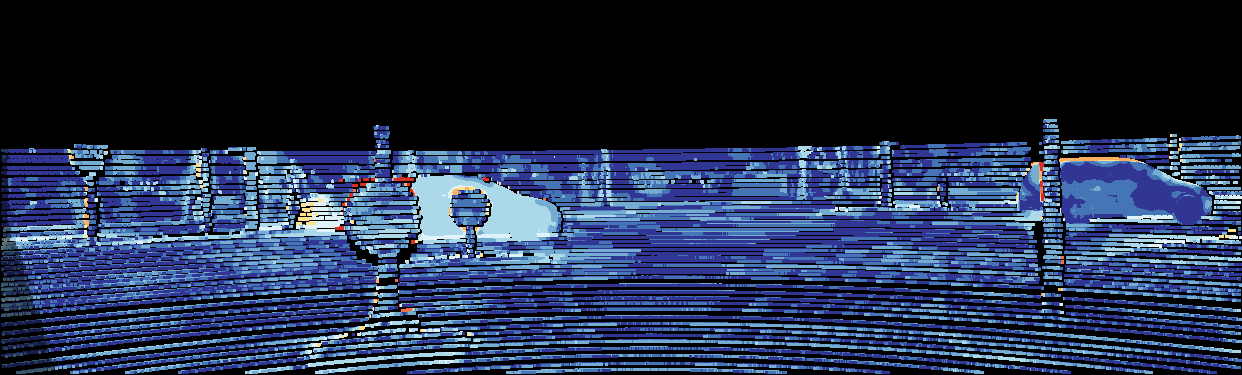}};
\draw (img.north west) node[labelstyle] {D2: 0.99};
\end{tikzpicture} &
\includegraphics[trim={0 0 0 \imgtrimtop},clip,width=0.197\textwidth]{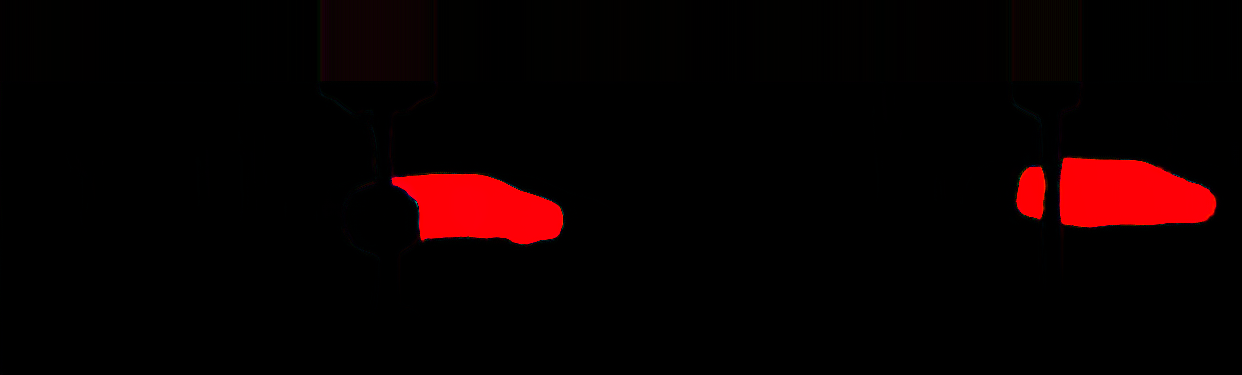} &
\begin{tikzpicture}
\draw (0, 0) node[imgstyle] (img) {
\includegraphics[trim={0 0 0 \imgtrimtop},clip,width=0.197\textwidth]{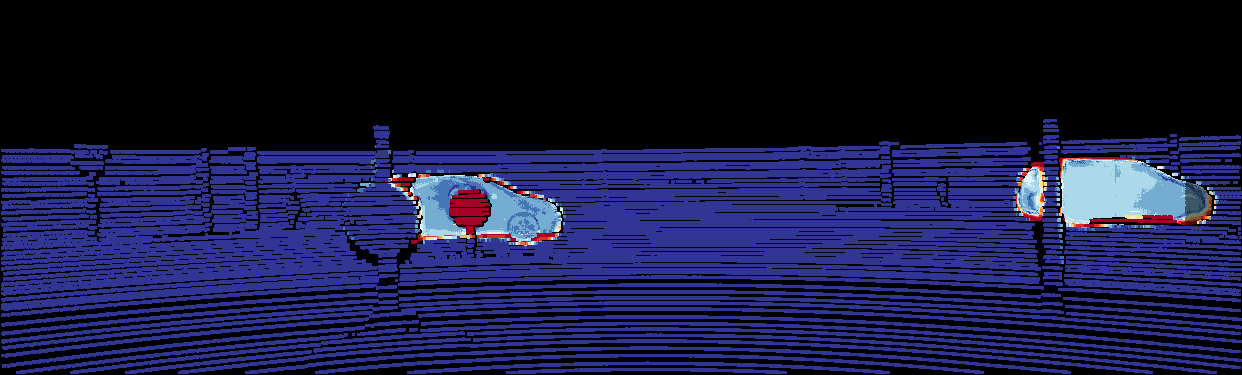}};
\draw (img.north west) node[labelstyle] {Fl: 2.02};
\end{tikzpicture} &
\begin{tikzpicture}
\draw (0, 0) node[imgstyle] (img) {
\includegraphics[trim={0 0 0 \imgtrimtop},clip,width=0.197\textwidth]{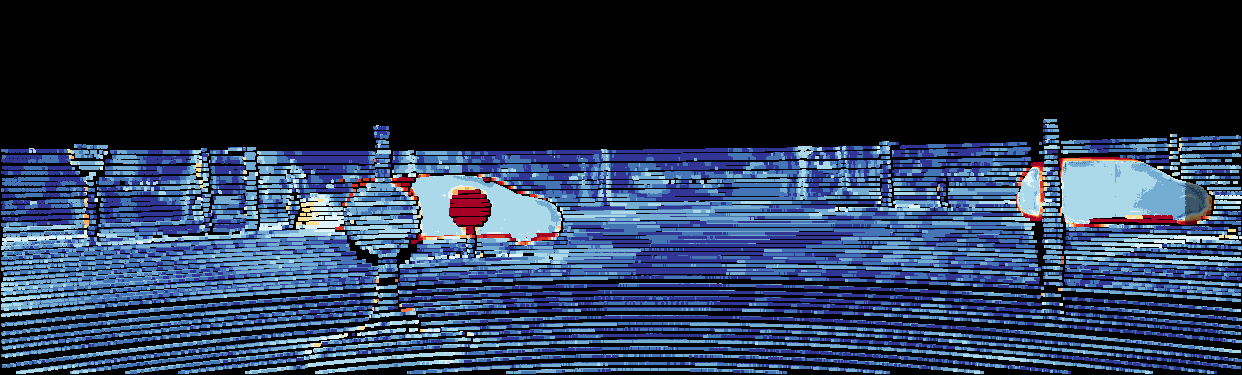}};
\draw (img.north west) node[labelstyle] {SF: 2.60};
\end{tikzpicture}
\end{tabular}
}
\caption{Qualitative comparison of our method, the original RAFT-3D, as well as the two top-performing approaches from the literature for two scenes using the visualizations provided by the KITTI benchmark~\cite{Menze2015_KITTI}. \emph{From left to right:} Target disparity visualization, corresponding \emph{D2} error plot, optical flow visualization, corresponding \emph{Fl} error plot, combined \emph{SF} error plot.}
\label{fig:addQualitative}
\end{figure*}

\subsection{Benchmark results}
In our first experiment, we compare the accuracy of our multi-frame scene flow method to that of other recent scene flow approaches from the literature. 
To this end, we computed the scene flow for the KITTI {\em test} split
both with our novel M-FUSE approach as well as with its underlying two-frame baseline and submitted the corresponding flow fields to the official %KITTI %scene flow 
benchmark~\cite{Menze2015_KITTI}.
To this end, we can not only show total improvements but also investigate the influence of the improved stereo method we employ.
Table~\ref{tab:KITTI_results} shows the obtained results together with the results of the ten top-ranked published scene flow methods.
Thereby, it lists the standard outlier rates 
\emph{D1} and \emph{D2} for the disparities at time $t$ and $t+1$, the optical flow error \emph{Fl} and the scene flow error \emph{SF}.
% $D1$, $D2$, $Fl$ and $SF$ for the disparities at time $t$ and $t+1$, the optical flow and the scene flow, respectively,
%$D1$ for disparity at time $t$, $D2$ for the disparity at time $t+1$, $Fl$ for the optical flow at the left frame pair as well as $SF$ for the scene flow 
% not only for \emph{all} pixels but also separately for static objects only moving due to camera motion (\emph{bg=background}) and for dynamic objects that move independently (\emph{fg=foreground});
These errors are evaluated on \emph{all} pixels, as well as separately for static \emph{background} (bg) objects only moving due to camera motion and for dynamic \emph{foreground} (fg) objects that move independently;
see~\cite{Menze2015_KITTI} for details.
%As in \cite{Menze2015_KITTI} the outlier rates are defined as the percentage of bad pixels in the dataset, where a pixel is deemed bad if it deviates by more than 3px and more than 5\% from the ground truth.
Additionally, for each outlier rate, the table shows relative improvements of the baseline and our method with respect to RAFT-3D as well as the relative improvements of our multi-frame approach compared to the two-frame baseline.

As one can see, already our baseline (RAFT-3D, with LEA\-Stereo) shows significantly improved results compared to the original RAFT-3D approach (with GANet).
In this context, the total gain of 9.9\% can be mainly attributed to strong improvements in the background region.
Our full \mbox{M-FUSE} approach then improves these results even further, outperforming RAFT-3D by 16.3\%.
Thereby, it also shows strong gains in the foreground, which are due to the consideration of multi-frame information (see M-FUSE vs.\ Baseline).
As a result, on the KITTI benchmark our method ranks second for all pixels, and first for foreground regions.

In Figure~\ref{fig:multframe_improv} we analyze the multi-frame improvements for four exemplary KITTI sequences.
In accordance with the numbers in Table~\ref{tab:KITTI_results}, we observe that (i) multi-frame improvements are strongest for the optical flow error compared to the disparity error and (ii) the improvements are most prominent on the individually moving foreground objects.
% We show visual results in Figure~\ref{fig:addQualitative}.
Figure~\ref{fig:addQualitative} visually shows improvements for background regions (\emph{top}) and foreground objects (\emph{bottom}).

\subsection{Ablations}

We ablate our model architecture in Table~\ref{tab:ablations}.
For these and all following experiments, we perform 4-fold cross validation on the KITTI \emph{train} split for more reliable evaluations with only limited data available.
Note that we omit the error measure for \emph{D1} in the tables since it is identical.

\vspace{1cm}
\medskip
\noindent
\textbf{Feature aggregation.}
Our U-Net computes additive increments for previous layers at the same resolution, which leads to a residual structure.
% adds the upsampled estimates to the previous layers at the same resolution giving it a residual structure where additive increments are estimated.
We compare this approach to the 
% original U-Net architecture~\cite{Ronneberger2015_UNet},
strategy presented in~\cite{Ronneberger2015_UNet},
where feature maps are concatenated and not summed up before a convolution.
% Employing additive residual connections consistently improves results likely since this allows the expanding part to predict incremental updates.
Using additive residual connections slightly improves results.

\medskip
\noindent
\textbf{Fusion module depth.}
In a second study, we ablate the depth of our fusion U-Net by comparing variants with two, three and four levels.
While a two-level U-Net still gives on-par results %with 3 levels 
in the \emph{D2} error, our three-level U-Net 
%still outperforms both it and the four-level U-Net
outperforms both the other networks 
in the \emph{Fl} and \emph{SF} errors.

\begin{table}
\caption{Ablation study. We show 4-fold cross validation results on KITTI \emph{train} in terms of the D2, Fl and SF errors~\cite{Menze2015_KITTI} as well as the number of parameters in millions.}
\label{tab:ablations}
\begin{center}
{\setlength\tabcolsep{2.7pt}
\begin{tabular}{lcccc}
\toprule
& D2 & Fl & SF & \#param\!
\\
\midrule
\midrule
two-frame & 1.81 & 3.67 & 4.07
\\
\midrule
\midrule
\multicolumn{4}{l}{\emph{Feature aggregation}}
\\
\midrule
concat. & 2.08 & 3.42 & 3.99 & 2.56
\\
add (ours) & \textbf{1.99} & \textbf{3.21} & \textbf{3.82} & 2.38
\\
\midrule
\midrule
\multicolumn{4}{l}{\emph{Fusion module depth}}
\\
\midrule
2 levels & \textbf{1.99} & 3.40 & 4.02 & 0.53
\\
3 levels (ours) & \textbf{1.99} & \textbf{3.21} & \textbf{3.82} & 2.38
\\
4 levels & 2.06 & 3.34 & 4.02 & 9.79
\\
\midrule
\midrule
\multicolumn{4}{l}{\emph{Additional fusion inputs}}
\\
\midrule
none & 2.72 & 3.33 & 4.62 & 2.36
\\
% corr
% \\
% dz
% \\
% ae
% \\
corrCost,dispRes & \textbf{1.87} & 3.36 & 3.83 & 2.36
\\
corrCost,embVec & 2.29 & 3.71 & 4.58 & 2.38
\\
dispRes,embVec & 1.99 & 3.43 & 3.97 & 2.38
\\
corrCost,dispRes,embVec (ours) & 1.99 & \textbf{3.21} & \textbf{3.82} & 2.38
\\
corrCost,dispRes,embVec, $I^{t}_l$ & 2.10 & 3.27 & 3.96 & 2.38
\\
\bottomrule
\end{tabular}
}
\end{center}
% \vspace{-1mm}
\end{table}

\medskip
\noindent
\textbf{Additional fusion inputs.}
Our fusion network makes use of forward and backward scene flow estimates.
In a larger set of experiments we determined which additional inputs to our fusion module are useful.
To this end, we compare our set of additional inputs (correlation costs, disparity residuals and rigid motion embedding vectors) to omitting all of them (none) and to omitting each of them individually, to assess their individual contribution.
The results show that omitting all additional features significantly worsens results, which indicates that valuable information is contained in our set of features.
% Further, we see that omitting the rigid motion embedding vectors gives on-par results to our method, with superior results in the \emph{D2} but a worse \emph{Fl} error.
Further, we see that omitting the rigid motion embedding vectors gives inconclusive results compared to our method, with superior \emph{D2} results but a worse \emph{Fl} error.
The disparity residuals seem most essential:
When removed, the resulting quality lowers significantly for all measures.
Further, removing correlation costs has a slight negative impact.
Additionally, we investigated if adding the reference frame $I^{t}_l$ to the set of inputs \cite{Ren2019_FlowTemporalFusion} is helpful.
However, this did not improve results any further, presumably because the correlation cost already provides sufficient information.

We show additional ablations with less conclusive results in the supplementary material.

\subsection{Scene flow parametrization}
Next, we investigate the influence of the underlying scene flow parametrization in our fusion module.
Table~\ref{tab:sf_param} compares the image-space parametrizations of optical flow and target disparity $(u,v,d')$ against optical flow and the change in disparity $(u,v,\Delta d)$.
Additionally, we investigate the world-space 3D motion vector parametrization $(x,y,z)$.

Evidently, the image-space parametrizations outperform the 3D vector parametrization by a large margin in the optical flow error \emph{Fl}.
We argue that this is due to the error measures that are employed in scene flow estimation, which also work in image space.
Among the image-space parametrizations, using the change in disparity instead of the target disparity yields better results.
Presumably, predicting the disparity change (motion) instead of the target disparity (structure) bears a greater resemblance to the optical flow, which renders this strategy superior for the joint prediction.
% more closely resembles the optical flow definition and thus in the joint prediction this strategy is superior.
\begin{table}
\caption{Influence of the scene flow parametrization.}
\label{tab:sf_param}
\begin{center}
\begin{tabular}{lccc}
\toprule
& D2 & Fl & SF
\\
\midrule
\midrule
% two-frame & 1.81 & 3.67 & 4.07
% \\
% \midrule
$(u,v,d')$ & 2.06 & 3.49 & 4.13
\\
$(u,v,\Delta d)$ (ours) & \textbf{1.99} & \textbf{3.21} & \textbf{3.82}
\\
$(x,y,z)$ & 2.04 & 8.50 & 8.79
\\
\bottomrule
\end{tabular}
\end{center}
\end{table}

\subsection{Comparison of multi-frame strategies}
\begin{table}
\setlength\tabcolsep{4pt}
\caption{Comparison of multi-frame strategies}
\label{tab:comparisonStrategies}
\begin{center}
\begin{tabular}{lccc}
\toprule
& D2 & Fl & SF
\\
\midrule
\midrule
two-frame & \textbf{1.81} & 3.67 & 4.07
\\
\midrule
warm-start (inv. backward) & 2.59 & 5.29 & 5.73
\\
warm-start (fw-warped prev.) & 2.23 & 4.48 & 4.88
\\
\midrule
learned inv + mask fusion & 2.06 & 3.96 & 4.39
\\
\midrule
specialized U-Net (bw-warped prev.) & 2.10 & 3.78 & 4.26
\\
specialized U-Net (inv. backward) & 2.01 & 3.59 & 4.05
\\
\midrule
M-FUSE & 1.99 & \textbf{3.21} & \textbf{3.82}
\\
\bottomrule
\end{tabular}
\end{center}
%
% ===================================================
%
% THIS WAS USED FOR ARXIV TO FIT ACKNOWLEDGEMENT
%
% ===================================================
%
% \vspace{-0.1cm}
%
% ===================================================
%
\end{table}

Finally, we compare our approach to three multi-frame strategies available in the literature:
Warm-starting the method, a learned inversion with mask-based fusion and a specialized U-Net with additional inputs; see Table~\ref{tab:comparisonStrategies}.

% First, the warm-start strategy has been shown to be highly successful~\cite{Teed2020_RAFT} in recent recurrent networks, where a forward-warped flow estimate from previous frames serves as an initialization for the recurrent structure.
First, the warm-start initialization strategy has been shown to be highly successful in recent recurrent networks~\cite{Teed2020_RAFT}.
We considered two variants for our baseline approach:
% Backward scene flow matrices that are inverted before initialization as well as previous scene flow forward-warped in the corresponding Lie algebra~\cite{Teed2021_Lie}.
For one, we used the matrix-inverted backward flow as an initialization, in contrast to the identity matrix initialization from~\cite{Teed2021_RAFT3D}.
For the other, we initialized with the previous forward scene flow that is forward-warped in the corresponding Lie algebra~\cite{Teed2021_Lie} using the estimated optical flow.
%
% One estimates the backward scene flow and inverts the transformation matrices before using them as initialization, in contrast to identity matrix initialization~\cite{Teed2021_RAFT3D}.
% The other variant follows the forward-warping strategy of~\cite{Teed2020_RAFT} estimating forward scene flow for the previous time step and uses the transformation matrix field as initialization after forward-warping it with the estimated optical flow.
% , where the optical flow is used to forward-warp the transformation matrix field to the current time step, which is used for initialization.
% Note that forward-warping is performed in the corresponding Lie algebra~\cite{Teed2021_Lie}.
In Table~\ref{tab:comparisonStrategies}, both approaches perform considerably worse than the two-frame baseline,
% even though the forward-warping variant yields better results than the inverted-backward variant.
even though forward-warping yields better results than inverted backward flow.
This is in line with previous studies, where warm-start on the KITTI dataset did not yield improvements~\cite{Teed2020_RAFT}.
% While the forward-warping variant yields better results than the inverted-backward variant, they both do not seem to bring any advantage.
% This could stem from the $SE(3)$ parametrization being more advanced than the one in \cite{Teed2020_RAFT} or from the specific motion in the KITTI dataset,
% on which the warm-start strategy was unsuccessful also in previous studies~\cite{Teed2020_RAFT}.
% on which the warm-start strategy was also previously unsuccessful~\cite{Teed2020_RAFT}.
% where also previously no improvements with the warmstart strategy were shown~\cite{Teed2020_RAFT}.

Second, we considered a recent strategy that relies on a learned backward-to-forward inverter~\cite{Maurer2018_ProFlow,Schuster2021_DTF} followed by a predicted fusion mask that linearly combines forward and backward estimates~\cite{Schuster2021_DTF}.
% This strategy implicitly learns a motion model relating backward and forward flow in the inversion.
% In contrast, our matrix inversion is parameter-free and a motion model is implicitly learned in our fusion module.
We reimplemented the inversion and fusion module from~\cite{Schuster2021_DTF} and pretrained the former, before using these modules in our method.
For comparability, we adapted the modules to our three-channel prediction case, keeping $D^t$ fixed.
% For the comparison, we reimplement the inversion module from~\cite{Schuster2021_DTF}, but keep $d$ fixed for comparability to our method yielding a 3-channel prediction by the inversion network and a 3-channel component-wise mask-based fusion.
% As in~\cite{Schuster2021_DTF}, we first pretrain the inversion module separately before training it end-to-end in our network.
In Table~\ref{tab:comparisonStrategies}, this strategy clearly yields worse results than our approach.
We attribute this to the simplistic structure of the motion model and the restrictive convex combination of flow inputs.
%which we attribute to two reasons:
%First, the motion model contained inside the inversion module may not be capable of representing the full complexity of the scenes due to its simplistic structure.
%Second, a linear combination of forward and inverted backward flow might be too restrictive, not allowing the model to deviate from its input and thus limiting its error correction capabilities.

Third, we investigated a strategy that employs the specialized fusion U-Net from FlowNet2~\cite{Ilg2017_Flownet2} for fusing optical flow estimates~\cite{Ren2019_FlowTemporalFusion} guided by a brightness constancy map and the reference image.
To this end, we extended this fusion module to the scene flow setting and embedded it in our approach.
For a fair comparison, we also added disparity residuals to its fusion inputs.
% To allow for a fair comparison with our M-FUSE approach, we propose a direct extension of their fusion module to our setting and use the brightness constancy error, reference image and disparity residuals as additional inputs to the network.
% In Table~\ref{tab:comparisonStrategies}, their strategy shows the best results so far, also slightly outperforming the two-frame baseline.
% However, in comparison our approach still yields significantly better results. 
We evaluated two variants, one with backward-warping the previous flow estimate as in~\cite{Ren2019_FlowTemporalFusion}, and one with inverted backward flow, as in our approach.
While only the approach with inversion is able to reach results on-par with the two-frame baseline, both cannot keep up with the results achieved by our method.

\subsection{Timing and parameter counts}
% \begin{table}
% \caption{Parameter Count}
% \label{tab:paramcount}
% \begin{center}
% \begin{tabular}{lccc}
% \toprule
% & stereo & scene flow & total
% \\
% \midrule
% %GANet + R3D
% RAFT-3D & 6.58M & \textbf{44.85M} & 51.43M
% \\
% Baseline & \textbf{1.81M} & \textbf{44.85M} & \textbf{46.66M}
% \\
% M-FUSE (ours) & \textbf{1.81M} & 47.22M & \textbf{49.03M}
% \\
% \bottomrule
% \end{tabular}
% \end{center}
% \end{table}

\begin{table}
\setlength\tabcolsep{3.5pt}
\caption{Parameter Count and Timing}
\label{tab:paramcount}
\begin{center}
\begin{tabular}{lcccccc}
\toprule
& \multicolumn{2}{c}{stereo} & \multicolumn{2}{c}{scene flow} & \multicolumn{2}{c}{total}
\\
\midrule
%GANet + R3D
RAFT-3D\;\; & 6.6M & 4.0s\;\; & 44.9M & 0.4s\;\; & 51.4M & 4.4s
\\
Baseline & 1.8M & 0.8s\;\; & 44.9M & 0.4s\;\; & 46.7M & 1.2s
\\
M-FUSE & 1.8M & 1.2s\;\; & 47.2M & 1.3s\;\; & 49.0M & 2.5s
\\
\bottomrule
\end{tabular}
\end{center}
\end{table}
% Table \ref{tab:paramcount} shows that our method takes a total of around 2.5s per frame for inference on a NVIDIA GeForce RTX 2080 Ti, %, compared to 0.4s for RAFT-3D. 
% consisting of 1.2s for stereo (3$\times$0.4s LEAStereo) and 1.3s for scene flow (2$\times$0.4s RAFT-3D baseline + 0.5s fusion).
% Naturally, our method yields longer inference times than the baseline while still being significantly faster than the original RAFT-3D approach employing GANet.
% %The parameter counts are given in Table~\ref{tab:paramcount}.
% %Our multi-frame fusion approach increases the required parameters by 5\% compared to the two-frame baseline in terms of the scene flow estimator.
% %However, when comparing to the full RAFT-3D approach~\cite{Teed2021_RAFT3D}, we still need less parameters as a result of the more parameter-efficient stereo approach~\cite{Cheng2020_LEAStereo}.
% Parameter-wise our multi-frame approach increases the count by 5\% compared to the two-frame baseline, while still having less parameters than RAFT-3D due to the %parameter-efficient
% lightweight LEAStereo method~\cite{Cheng2020_LEAStereo}; see Table~\ref{tab:paramcount}.

Table \ref{tab:paramcount} shows that our method takes a total of around 2.5s per frame for inference on a NVIDIA GeForce RTX 2080 Ti with 51.4M parameters.
The runtime is composed of 1.2s for stereo (3$\times$0.4s LEAStereo) and 1.3s for scene flow (2$\times$0.4s RAFT-3D baseline + 0.5s fusion).
While runtime and parameter count is increased compared to the two-frame baseline, our method is still faster and more parameter-efficient than the original RAFT-3D approach due to the fast and lightweight LEAStereo method~\cite{Cheng2020_LEAStereo}.

% Naturally, our method yields longer inference times than the baseline while still being significantly faster than the original RAFT-3D approach employing GANet.
%The parameter counts are given in Table~\ref{tab:paramcount}.

\section{Conclusion}
We proposed a novel multi-frame scene flow approach that leverages the performance of recent high accuracy two-frame methods. To this end, we developed an improved RAFT-3D baseline and embedded it into a U-Net-based fusion approach that adaptively integrates temporal information by combining an $SE(3)$-based extrapolation of the backward flow with the jointly estimated forward flow. 
The achieved results clearly demonstrate that our strategy of explicitly tailoring our architecture towards the underlying baseline pays off.
With more than 16\% improvements compared to the original RAFT-3D approach, they show significantly larger improvements than other multi-frame networks in the literature. Moreover, in absolute accuracy  
our method ranks second in the public KITTI benchmark, clearly outperforming all other multi-frame approaches. 

%We proposed a novel approach for scene flow estimation that makes use of stereo frames from three time instances.
%First, we show that utilizing a recent stereo estimation method already significantly improves the RAFT-3D scene flow approach.
%Second, we propose a U-Net for the fusion of forward and inverted backward scene flow estimates which again yields substantial improvements, especially for the challenging independently-moving foreground objects.
%With total improvements of 16\% we currently rank second on the public KITTI benchmark.

\medskip
\noindent
\textbf{Acknowledgements.}
Funded by the Deutsche Forschungsgemeinschaft (DFG, German Research Foundation) -- Project-ID 251654672 -- TRR 161 (B04).
A.J.\ and J.S.\ acknowledge support from the International Max Planck Research School for Intelligent Systems (IMPRS-IS).
% \cleardoublepage

{\small
\bibliographystyle{ieee_fullname}
\bibliography{references}
}

\clearpage
\appendix

In this supplementary material, we provide a visualization of our fusion U-Net, additional ablations and additional qualitative results.

\section{Architecture of the U-Net}
\begin{figure}[h!]
    \centering
    \includegraphics[width=0.8\linewidth]{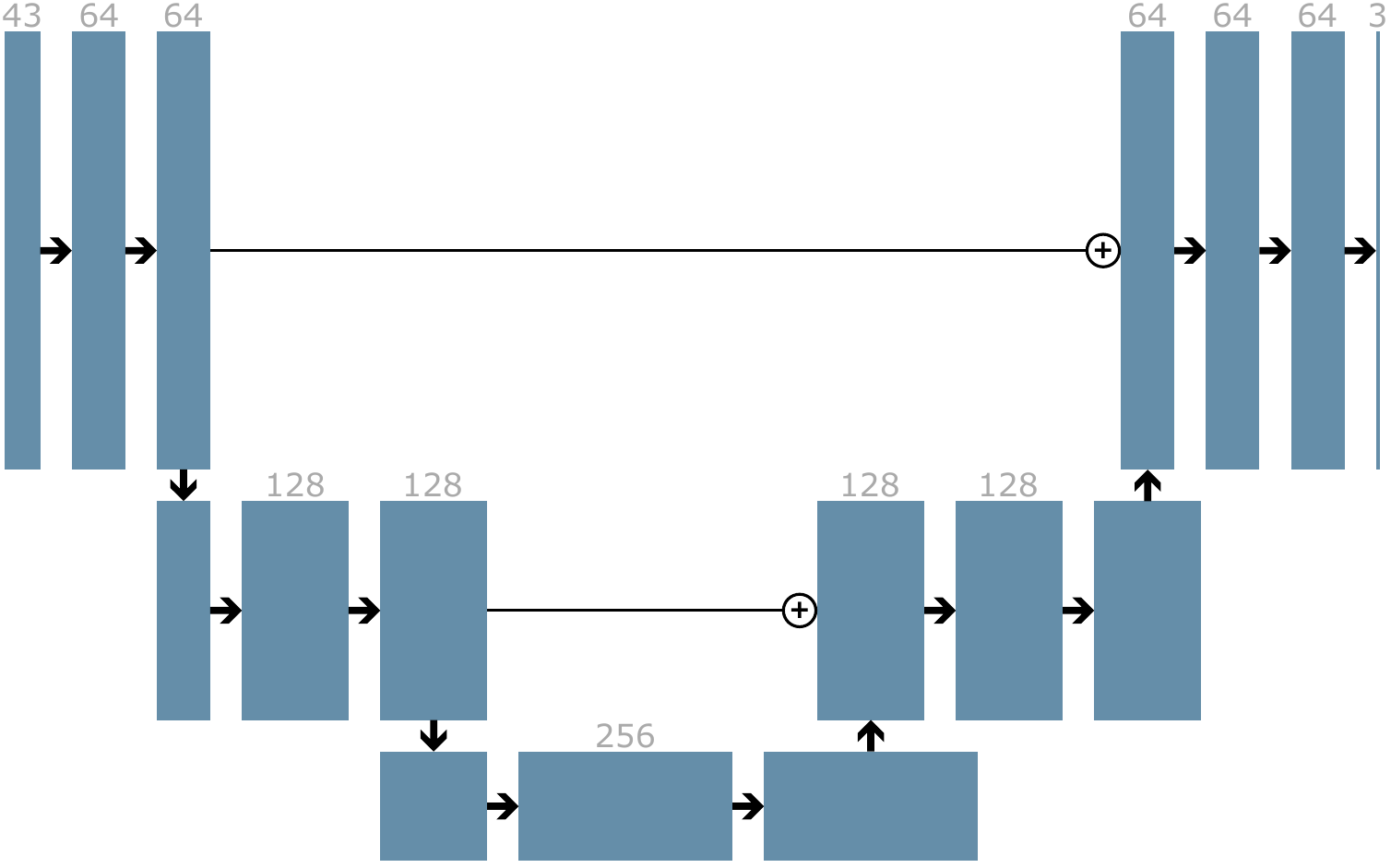}
    \caption{Architecture of our fusion U-Net.}
    \label{fig:UNet}
\end{figure}

Figure~\ref{fig:UNet} shows the architecture of our 3-level U-Net with residual connections.

\section{Additional ablations}
In addition to the ablations in the main paper, we conducted two more experiments as shown in Table~\ref{tab:addablations}.

\medskip
\noindent
\textbf{In-between convolutions.}
As can be seen in Figure~\ref{fig:UNet}, in every depth level for the contracting as well as the expanding part one additional in-between convolutional layer is used to process information.
Thus, we performed an ablation over several options: completely omitting this layer (none), having one (1 conv) or two (2 convs) convolutions, or using a residual block~\cite{He2016_ResNet} (resblock).
%consisting of two convolutions with residual skip connection.
% The results for none, one or two convolutional layers are very close to each other, where no convolutional layer results in on-par \emph{D2} results.
% Two convolutions outperform a single one in the \emph{Fl} error -- however at the cost of decreased \emph{D2} quality.
% Thus, in the combined \emph{SF} error, one convolution slightly outperforms the other choices.
The results for none, one or two convolutional layers are inconclusive, with no significant best option.
As a compromise, we chose one convolutional layer for our method since it is most similar to other U-Nets in the literature.
Finally, despite being most closely related to the two convolutions, the residual block slightly decreases the quality compared to all other cases.
% Finally, investigating the introduction of a residual block, which is most closely related to the 2 convs case, we can see that the quality further decreases compared to the 2 convs case and subsequently compared to our architecture.

\medskip
\noindent
\textbf{Image features.}
Finally, we compare two options to encode image-related features guiding our fusion module.
The first option is to utilize the learned correlation cost from our baseline, which is upsampled from 1/8th of the resolution.
% However, they are computed on 1/8th of the resolution and upsampled by the network.
The second option is a full-resolution brightness constancy error map~\cite{Ren2019_FlowTemporalFusion} %, which is calculated
as the $L2$ distance between the warped and original image.
% This option is available on the full image resolution.
As one can see, the learned correlation features outperform the brightness constancy maps slightly -- although the former are upsampled from lower resolution.
% This is probably due to the fact some learned preprocessing is needed instead of directly working with the image pixel values.
%We hypothesize that a learned preprocessing helps to increase the contained information compared to directly using raw pixel values.

\begin{table}
\caption{Additional ablations. We show 4-fold cross validation results on KITTI \emph{train} in terms of the D2, Fl and SF errors as well as the number of parameters in millions.}
\label{tab:addablations}
\begin{center}
{\setlength\tabcolsep{2.7pt}
\begin{tabular}{lcccc}
\toprule
& D2 & Fl & SF & \#param\!
\\
\midrule
\midrule
two-frame & 1.81 & 3.67 & 4.07
\\
\midrule
\midrule
\multicolumn{4}{l}{\emph{In-between convs}}
\\
\midrule
none & \textbf{1.99} & 3.33 & 3.89 & 1.42
\\
1 conv (ours) & \textbf{1.99} & 3.21 & \textbf{3.82} & 2.38
\\
2 convs & 2.06 & \textbf{3.19} & 3.84 & 3.34
\\
resblock & 2.08 & 3.47 & 3.99 & 3.34
\\
\midrule
\midrule
\multicolumn{4}{l}{\emph{Image features}}
\\
\midrule
corrCost (ours) & \textbf{1.99} & \textbf{3.21} & \textbf{3.82} & 2.38
\\
BCE & 2.00 & 3.33 & 3.96 & 2.38
\\
\bottomrule
\end{tabular}
}
\end{center}
\end{table}

\section{Additional qualitative results}
\begin{figure*}
\centering{
\tikzset{labelstyle/.style={anchor=north west, text=white, inner sep=2, text opacity=1, scale=0.7, yshift=-1, xshift=1, fill=black, opacity=0.6}}
\tikzset{imgstyle/.style={inner sep=0,anchor=north west,outer sep=0,draw=none,line width=0}}
\setlength\tabcolsep{1pt}
\begin{tabular}{ccccc}
\begin{tikzpicture}
\draw (0, 0) node[imgstyle] (img) {
\includegraphics[trim={0 0 0 2.85cm},clip,width=0.197\textwidth]{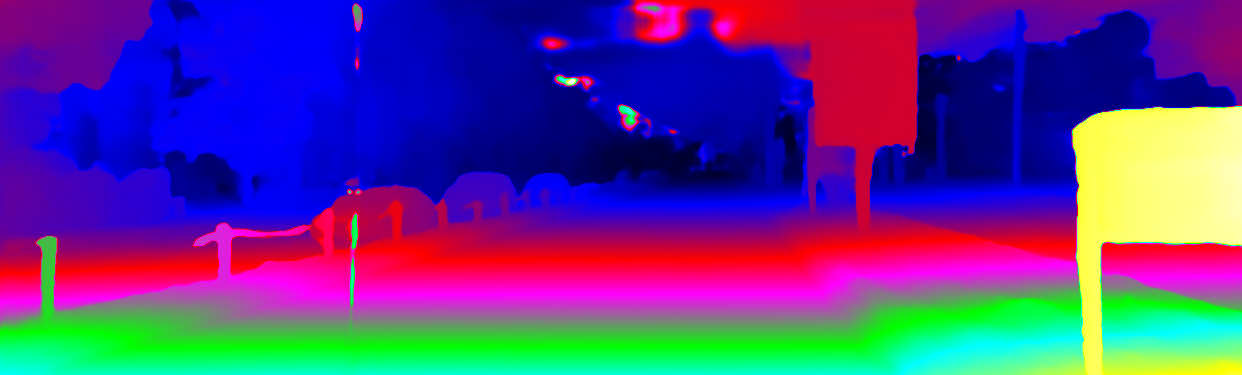}};
\draw (img.north west) node[labelstyle] {RigidMask+ISF};
\end{tikzpicture} &
\begin{tikzpicture}
\draw (0, 0) node[imgstyle] (img) {
\includegraphics[trim={0 0 0 2.85cm},clip,width=0.197\textwidth]{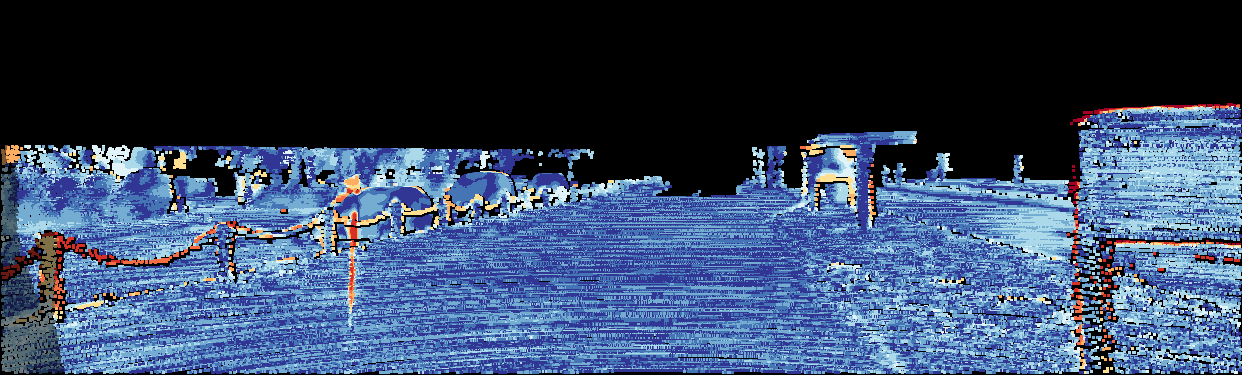}};
\draw (img.north west) node[labelstyle] {D2: 2.26};
\end{tikzpicture} &
\includegraphics[trim={0 0 0 2.85cm},clip,width=0.197\textwidth]{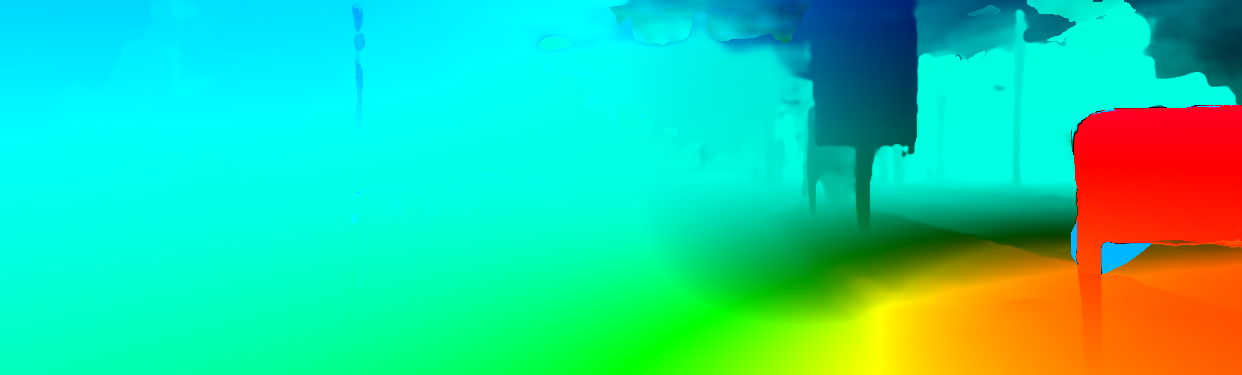} &
\begin{tikzpicture}
\draw (0, 0) node[imgstyle] (img) {
\includegraphics[trim={0 0 0 2.85cm},clip,width=0.197\textwidth]{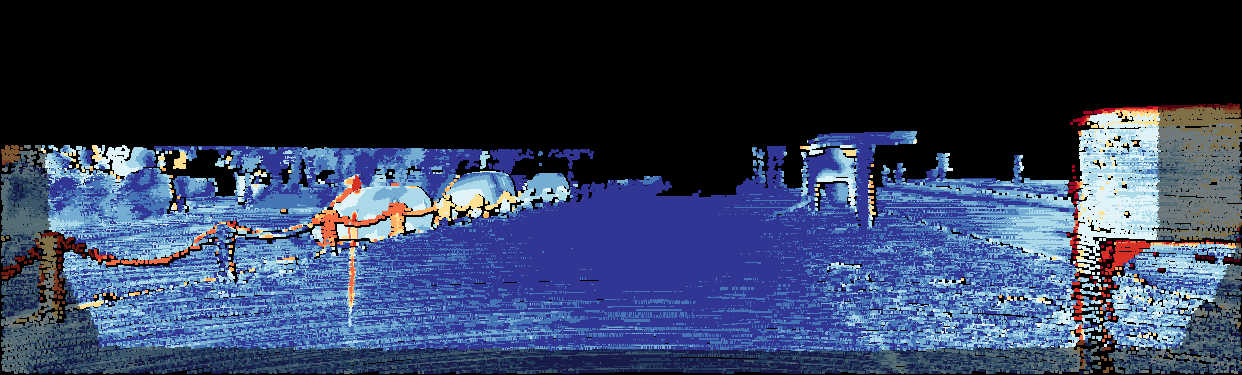}};
\draw (img.north west) node[labelstyle] {Fl: 3.24};
\end{tikzpicture} &
\begin{tikzpicture}
\draw (0, 0) node[imgstyle] (img) {
\includegraphics[trim={0 0 0 2.85cm},clip,width=0.197\textwidth]{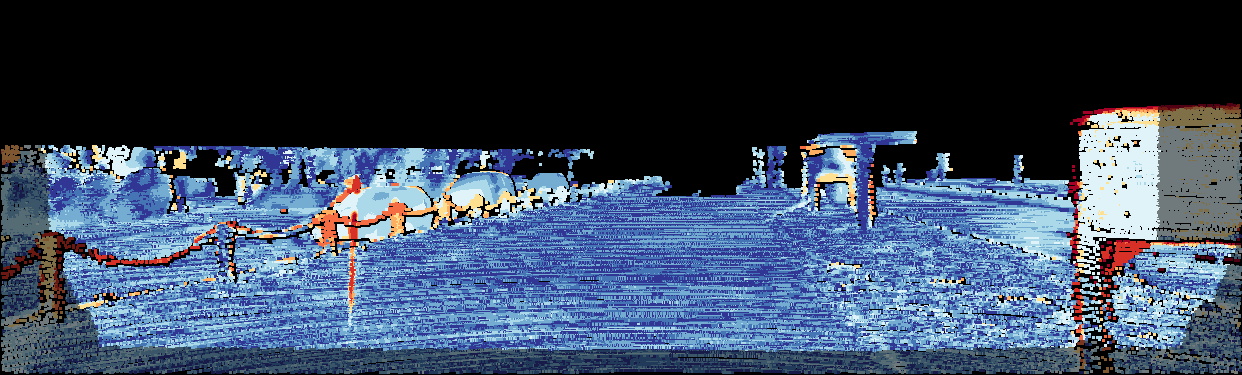}};
\draw (img.north west) node[labelstyle] {SF: 3.66};
\end{tikzpicture}
\\
\end{tabular}
}
\caption{Qualitative comparison of our method, the original RAFT-3D, as well as the two top-performing approaches from the literature using the visualizations provided by the KITTI benchmark. \emph{From left to right:} Target disparity visualization, corresponding \emph{D2} error plot, optical flow visualization, corresponding \emph{Fl} error plot, combined \emph{SF} error plot.}
\label{fig:begin}
\end{figure*}

\begin{figure*}
\centering{
\tikzset{labelstyle/.style={anchor=north west, text=white, inner sep=2, text opacity=1, scale=0.7, yshift=-1, xshift=1, fill=black, opacity=0.6}}
\tikzset{imgstyle/.style={inner sep=0,anchor=north west,outer sep=0,draw=none,line width=0}}
\setlength\tabcolsep{1pt}
\begin{tabular}{ccccc}
\begin{tikzpicture}
\draw (0, 0) node[imgstyle] (img) {
\includegraphics[trim={0 0 0 2.85cm},clip,width=0.197\textwidth]{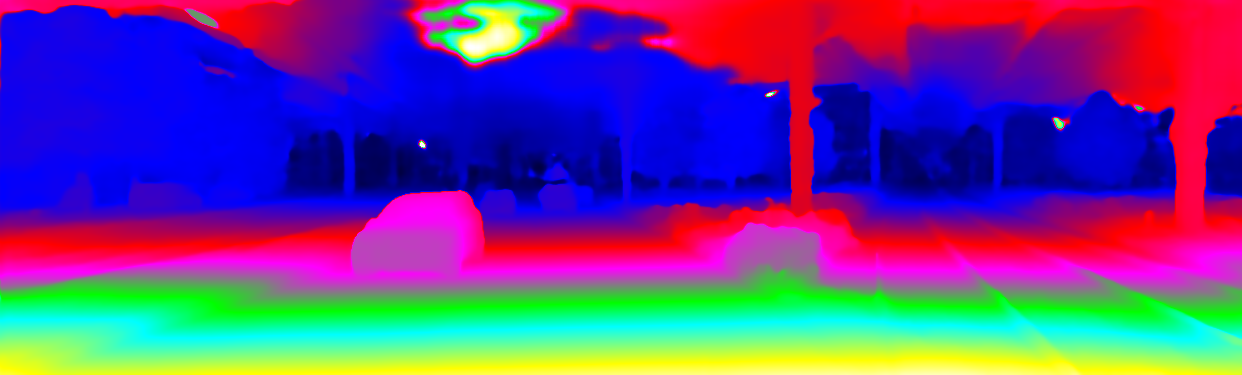}};
\draw (img.north west) node[labelstyle] {RigidMask+ISF};
\end{tikzpicture} &
\begin{tikzpicture}
\draw (0, 0) node[imgstyle] (img) {
\includegraphics[trim={0 0 0 2.85cm},clip,width=0.197\textwidth]{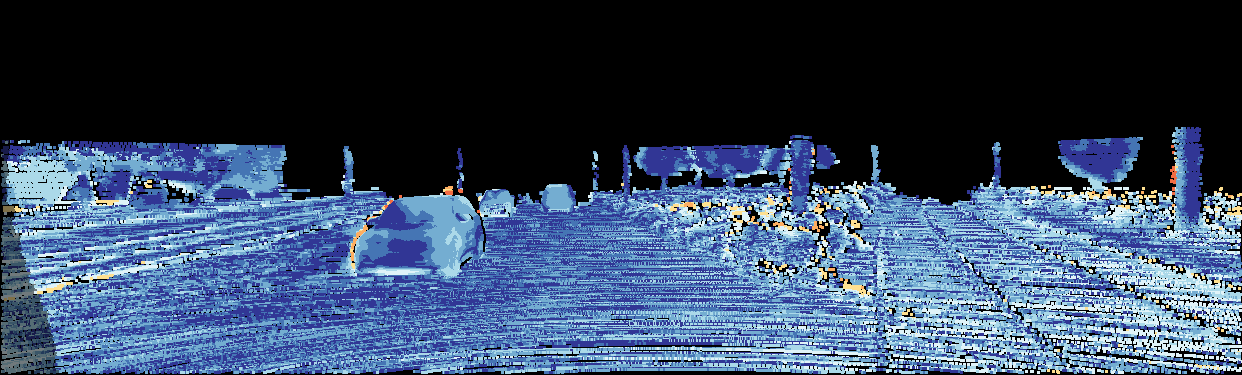}};
\draw (img.north west) node[labelstyle] {D2: 1.65};
\end{tikzpicture} &
\includegraphics[trim={0 0 0 2.85cm},clip,width=0.197\textwidth]{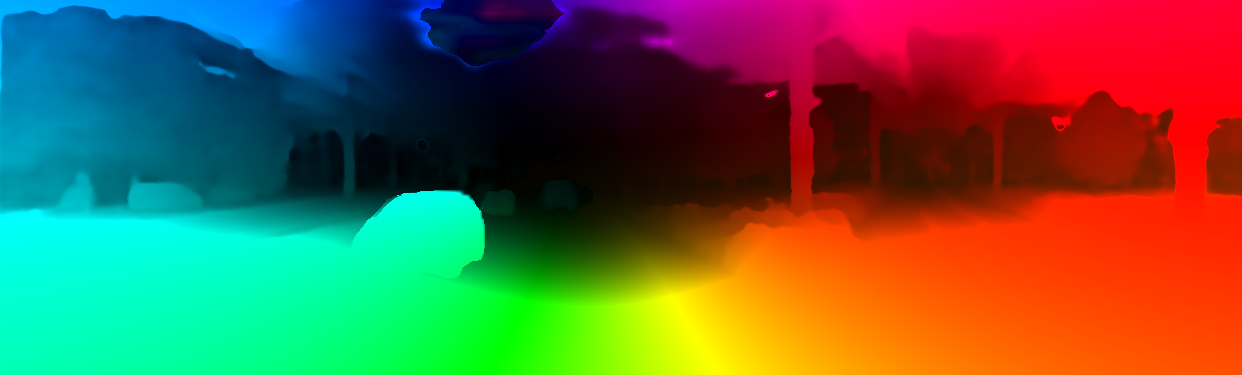} &
\begin{tikzpicture}
\draw (0, 0) node[imgstyle] (img) {
\includegraphics[trim={0 0 0 2.85cm},clip,width=0.197\textwidth]{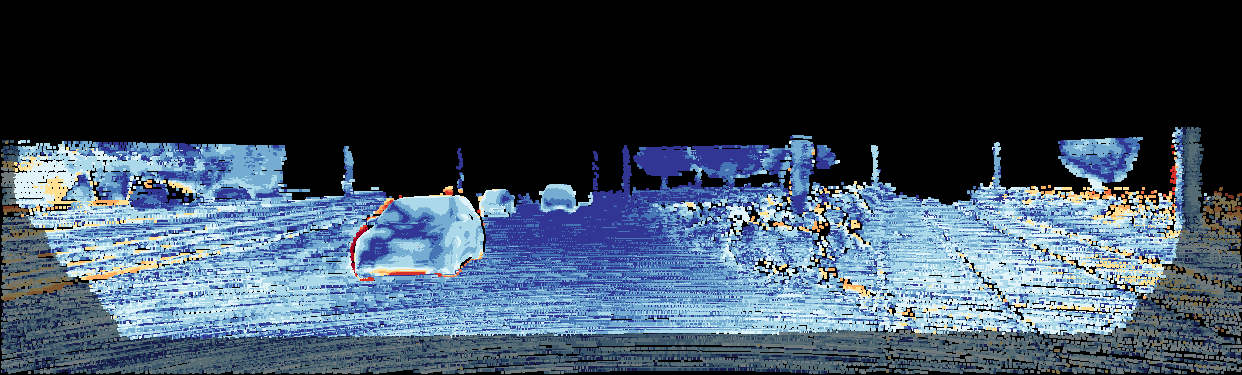}};
\draw (img.north west) node[labelstyle] {Fl: 2.87};
\end{tikzpicture} &
\begin{tikzpicture}
\draw (0, 0) node[imgstyle] (img) {
\includegraphics[trim={0 0 0 2.85cm},clip,width=0.197\textwidth]{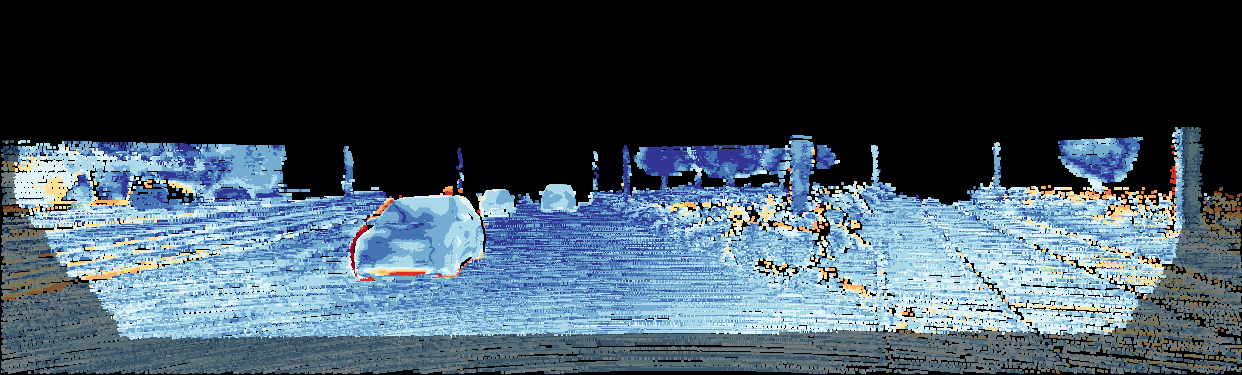}};
\draw (img.north west) node[labelstyle] {SF: 3.36};
\end{tikzpicture}
\\[-0.6mm]
\begin{tikzpicture}
\draw (0, 0) node[imgstyle] (img) {
\includegraphics[trim={0 0 0 2.85cm},clip,width=0.197\textwidth]{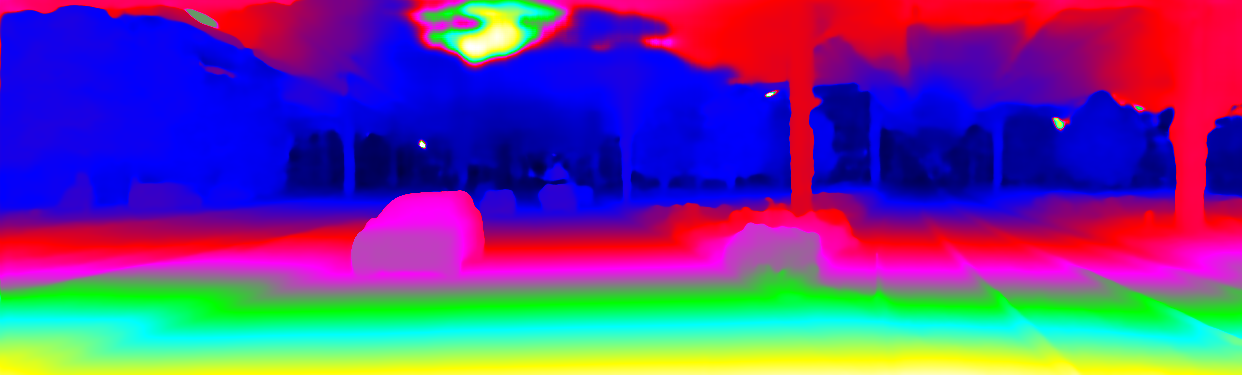}};
\draw (img.north west) node[labelstyle] {CamLiFlow};
\end{tikzpicture} &
\begin{tikzpicture}
\draw (0, 0) node[imgstyle] (img) {
\includegraphics[trim={0 0 0 2.85cm},clip,width=0.197\textwidth]{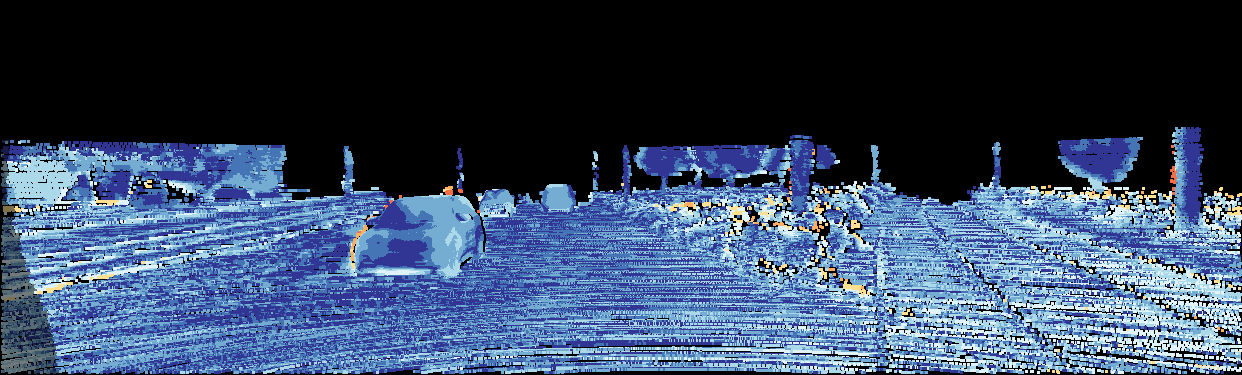}};
\draw (img.north west) node[labelstyle] {D2: 2.01};
\end{tikzpicture} &
\includegraphics[trim={0 0 0 2.85cm},clip,width=0.197\textwidth]{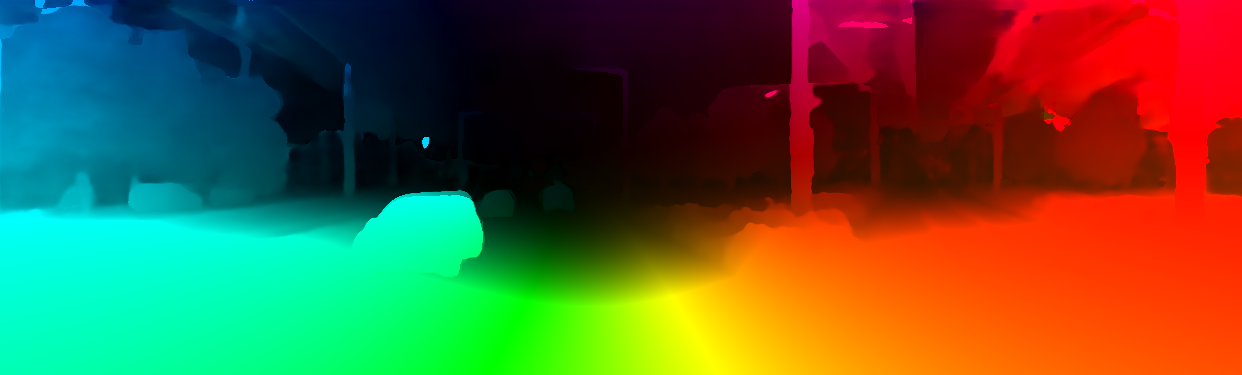} &
\begin{tikzpicture}
\draw (0, 0) node[imgstyle] (img) {
\includegraphics[trim={0 0 0 2.85cm},clip,width=0.197\textwidth]{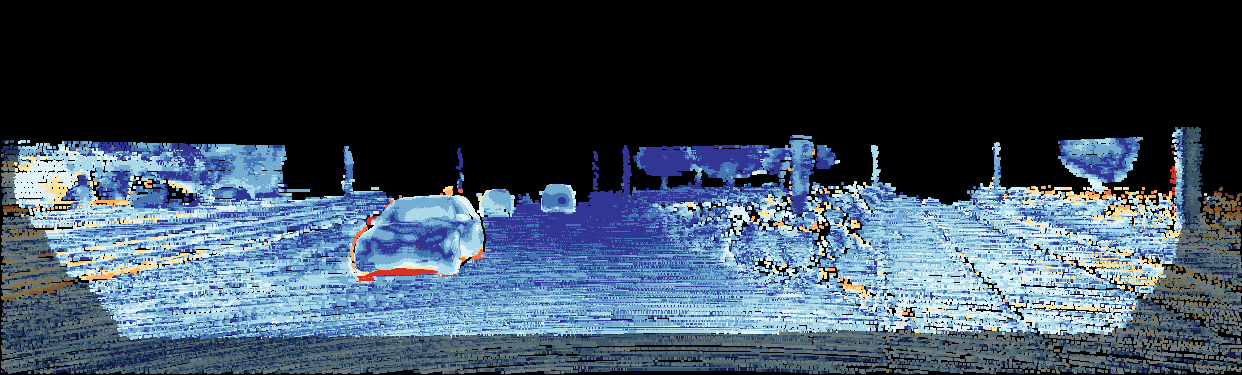}};
\draw (img.north west) node[labelstyle] {Fl: 3.12};
\end{tikzpicture} &
\begin{tikzpicture}
\draw (0, 0) node[imgstyle] (img) {
\includegraphics[trim={0 0 0 2.85cm},clip,width=0.197\textwidth]{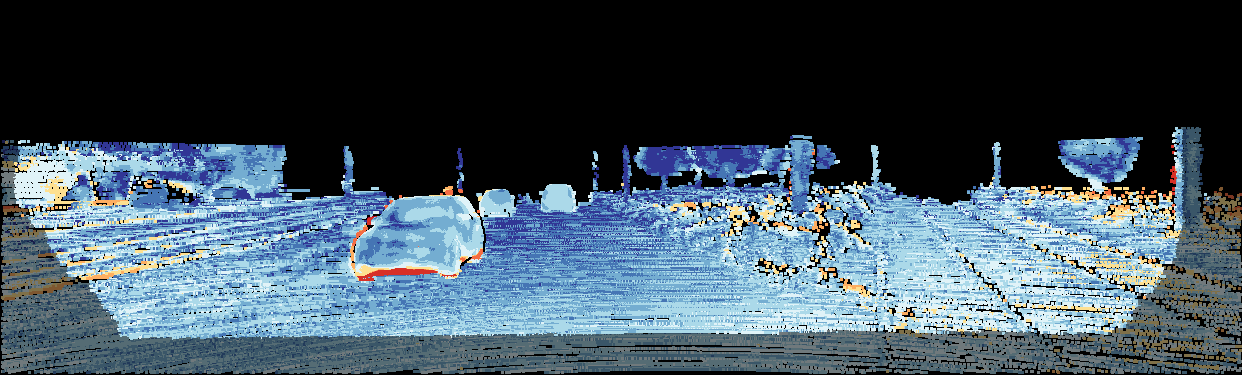}};
\draw (img.north west) node[labelstyle] {SF: 3.74};
\end{tikzpicture}
\\[-0.6mm]
\begin{tikzpicture}
\draw (0, 0) node[imgstyle] (img) {
\includegraphics[trim={0 0 0 2.85cm},clip,width=0.197\textwidth]{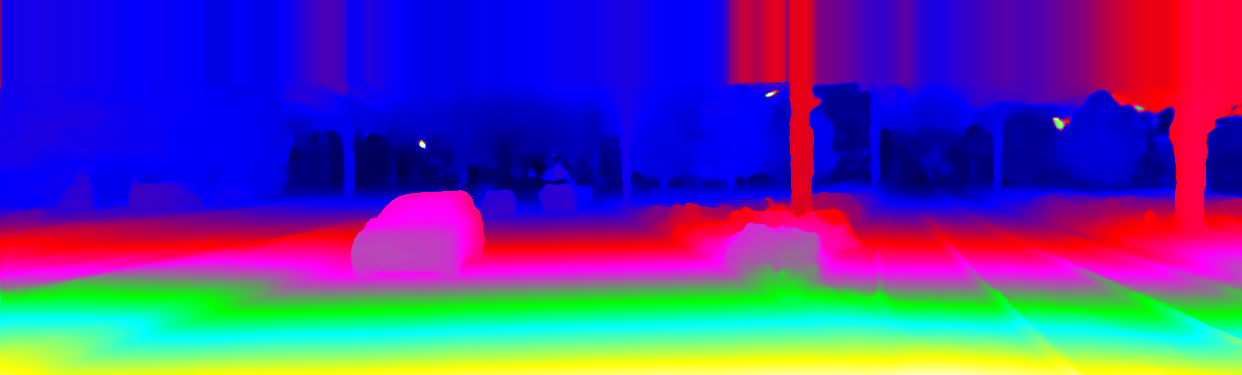}};
\draw (img.north west) node[labelstyle] {RAFT-3D};
\end{tikzpicture} &
\begin{tikzpicture}
\draw (0, 0) node[imgstyle] (img) {
\includegraphics[trim={0 0 0 2.85cm},clip,width=0.197\textwidth]{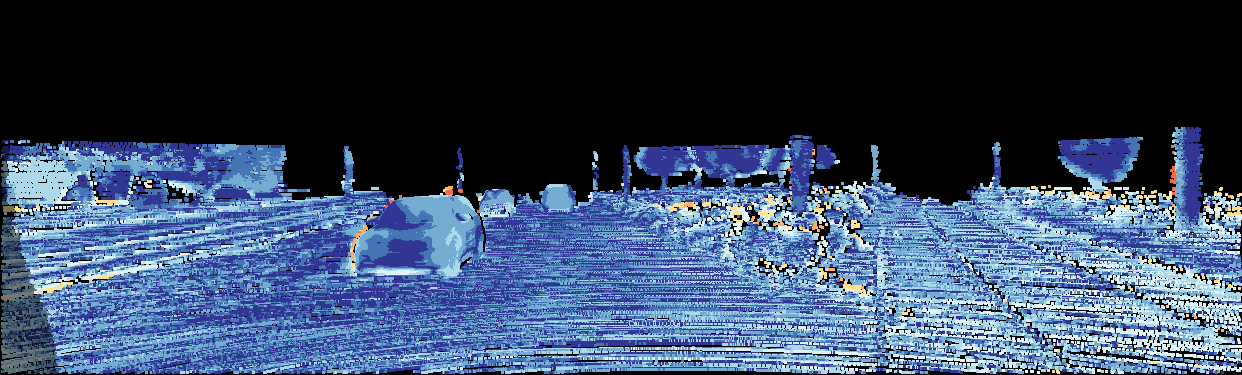}};
\draw (img.north west) node[labelstyle] {D2: 1.31};
\end{tikzpicture} &
\includegraphics[trim={0 0 0 2.85cm},clip,width=0.197\textwidth]{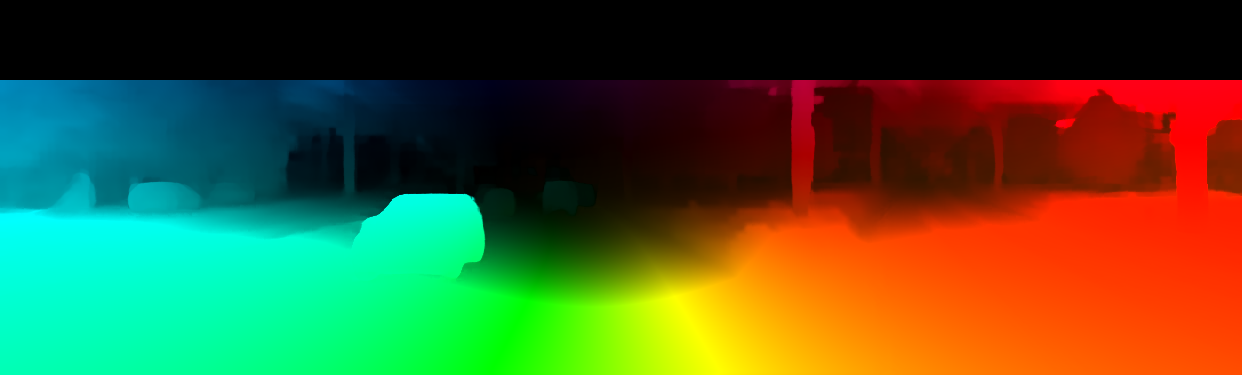} &
\begin{tikzpicture}
\draw (0, 0) node[imgstyle] (img) {
\includegraphics[trim={0 0 0 2.85cm},clip,width=0.197\textwidth]{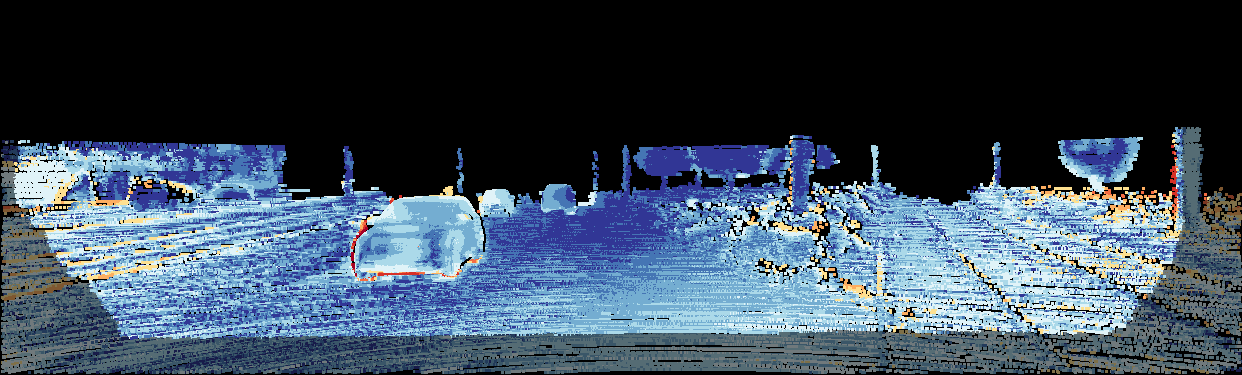}};
\draw (img.north west) node[labelstyle] {Fl: 2.92};
\end{tikzpicture} &
\begin{tikzpicture}
\draw (0, 0) node[imgstyle] (img) {
\includegraphics[trim={0 0 0 2.85cm},clip,width=0.197\textwidth]{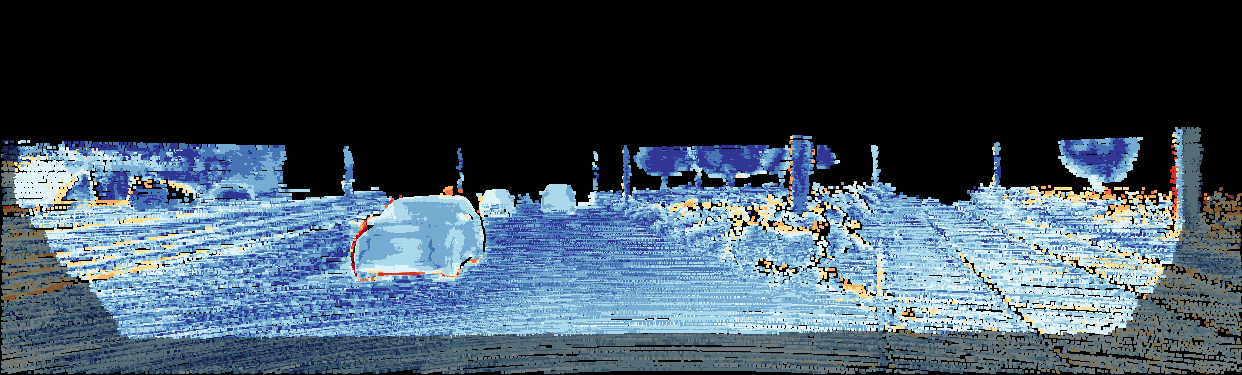}};
\draw (img.north west) node[labelstyle] {SF: 3.30};
\end{tikzpicture}
\\[-0.6mm]
\begin{tikzpicture}
\draw (0, 0) node[imgstyle] (img) {
\includegraphics[trim={0 0 0 2.85cm},clip,width=0.197\textwidth]{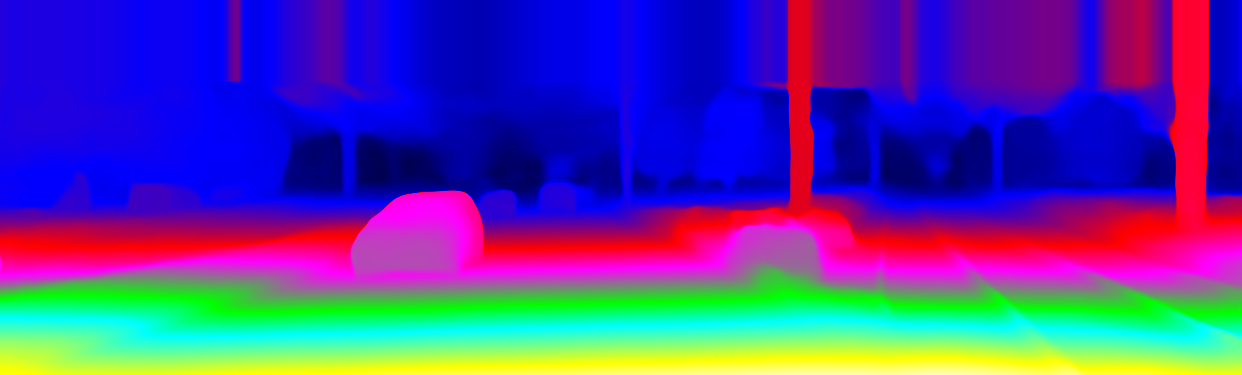}};
\draw (img.north west) node[labelstyle] {M-FUSE};
\end{tikzpicture} &
\begin{tikzpicture}
\draw (0, 0) node[imgstyle] (img) {
\includegraphics[trim={0 0 0 2.85cm},clip,width=0.197\textwidth]{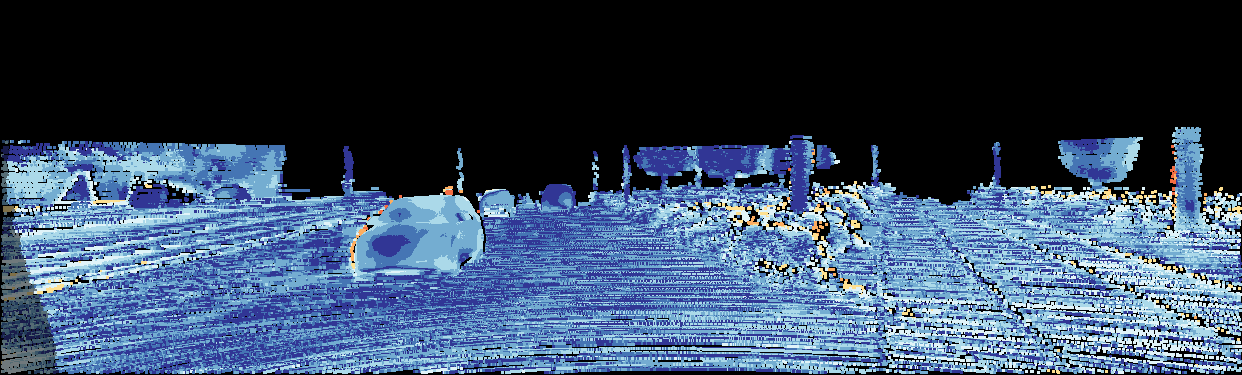}};
\draw (img.north west) node[labelstyle] {D2: 1.39};
\end{tikzpicture} &
\includegraphics[trim={0 0 0 2.85cm},clip,width=0.197\textwidth]{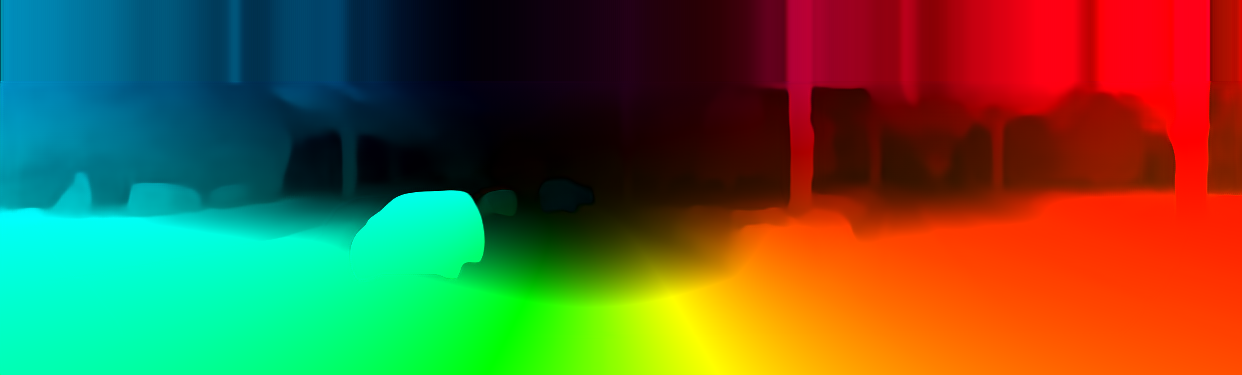} &
\begin{tikzpicture}
\draw (0, 0) node[imgstyle] (img) {
\includegraphics[trim={0 0 0 2.85cm},clip,width=0.197\textwidth]{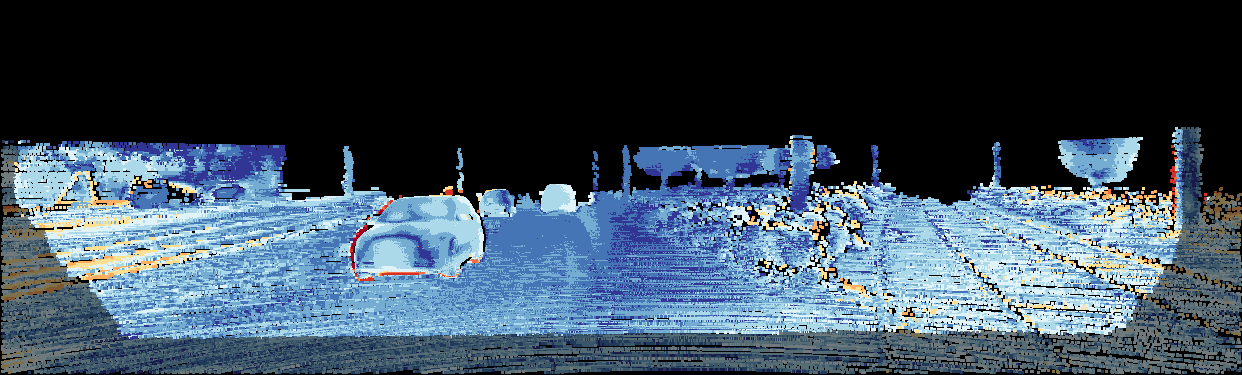}};
\draw (img.north west) node[labelstyle] {Fl: 2.84};
\end{tikzpicture} &
\begin{tikzpicture}
\draw (0, 0) node[imgstyle] (img) {
\includegraphics[trim={0 0 0 2.85cm},clip,width=0.197\textwidth]{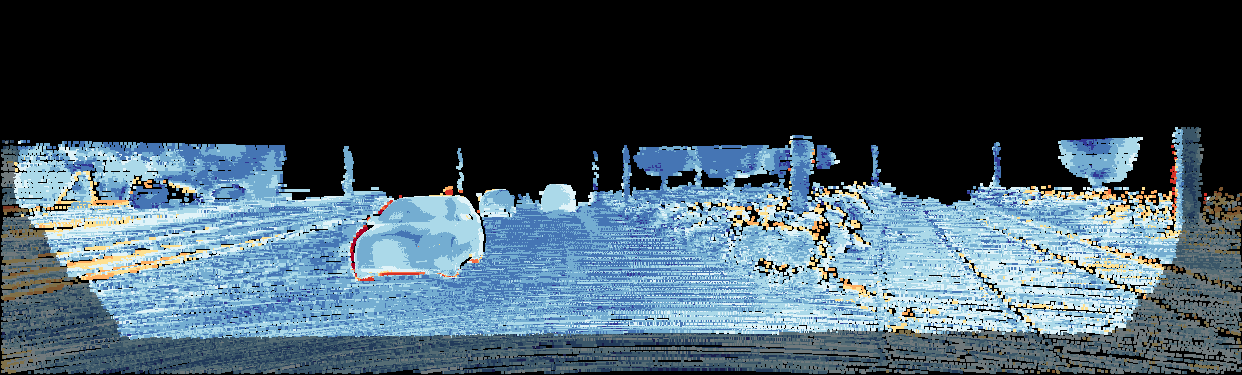}};
\draw (img.north west) node[labelstyle] {SF: 3.11};
\end{tikzpicture}
\\
\begin{tikzpicture}
\draw (0, 0) node[imgstyle] (img) {
\includegraphics[trim={0 0 0 2.85cm},clip,width=0.197\textwidth]{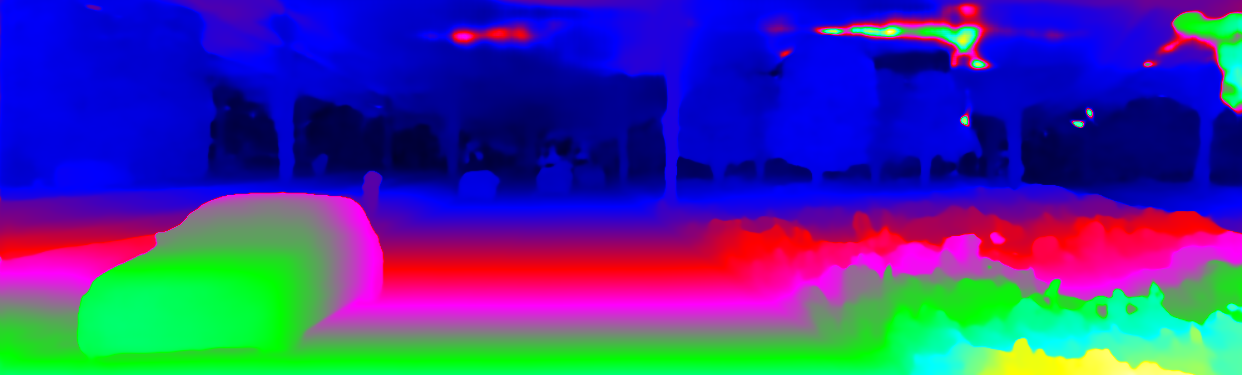}};
\draw (img.north west) node[labelstyle] {RigidMask+ISF};
\end{tikzpicture} &
\begin{tikzpicture}
\draw (0, 0) node[imgstyle] (img) {
\includegraphics[trim={0 0 0 2.85cm},clip,width=0.197\textwidth]{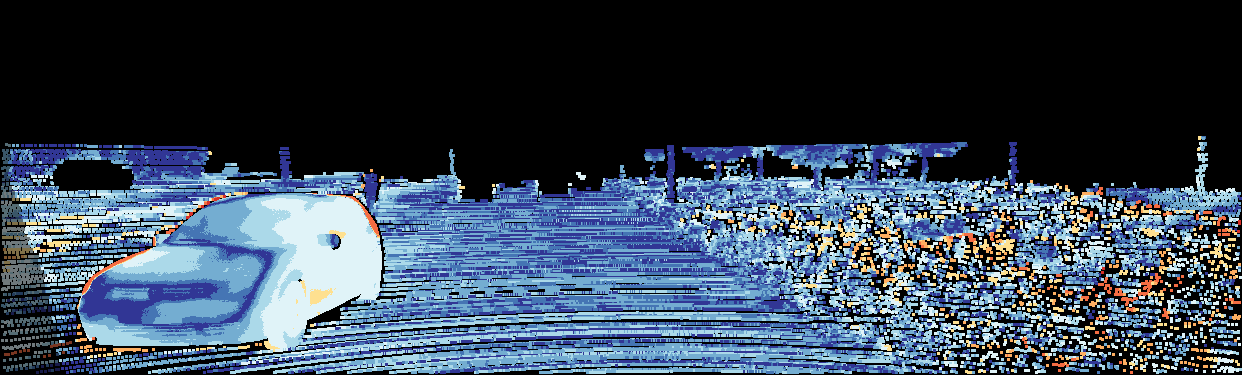}};
\draw (img.north west) node[labelstyle] {D2: 47.41};
\end{tikzpicture} &
\includegraphics[trim={0 0 0 2.85cm},clip,width=0.197\textwidth]{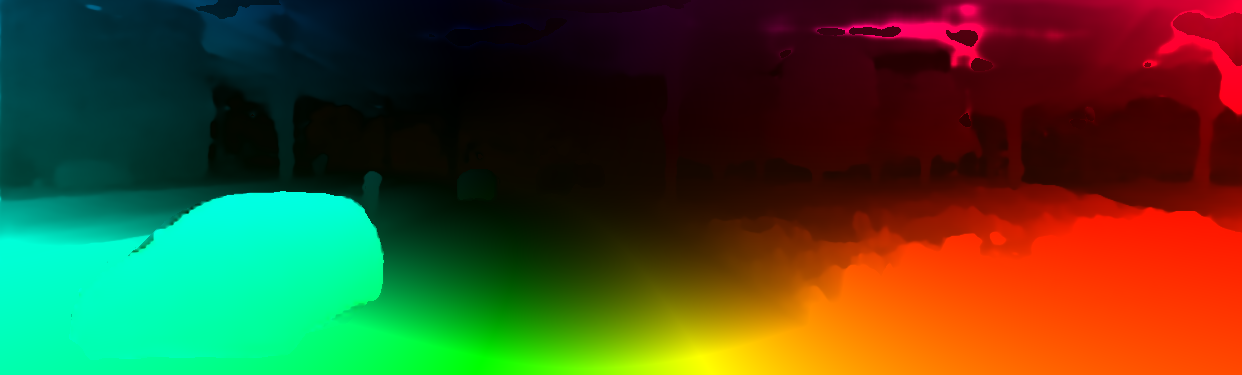} &
\begin{tikzpicture}
\draw (0, 0) node[imgstyle] (img) {
\includegraphics[trim={0 0 0 2.85cm},clip,width=0.197\textwidth]{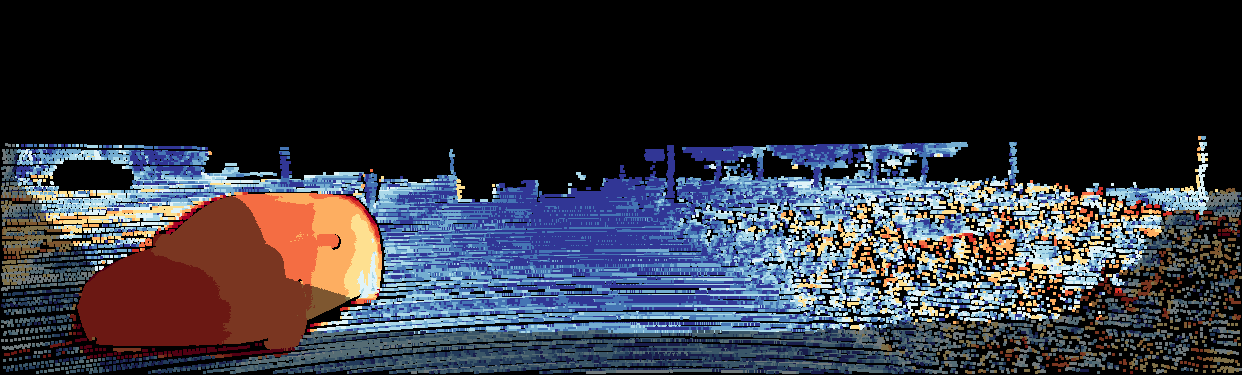}};
\draw (img.north west) node[labelstyle] {Fl: 51.39};
\end{tikzpicture} &
\begin{tikzpicture}
\draw (0, 0) node[imgstyle] (img) {
\includegraphics[trim={0 0 0 2.85cm},clip,width=0.197\textwidth]{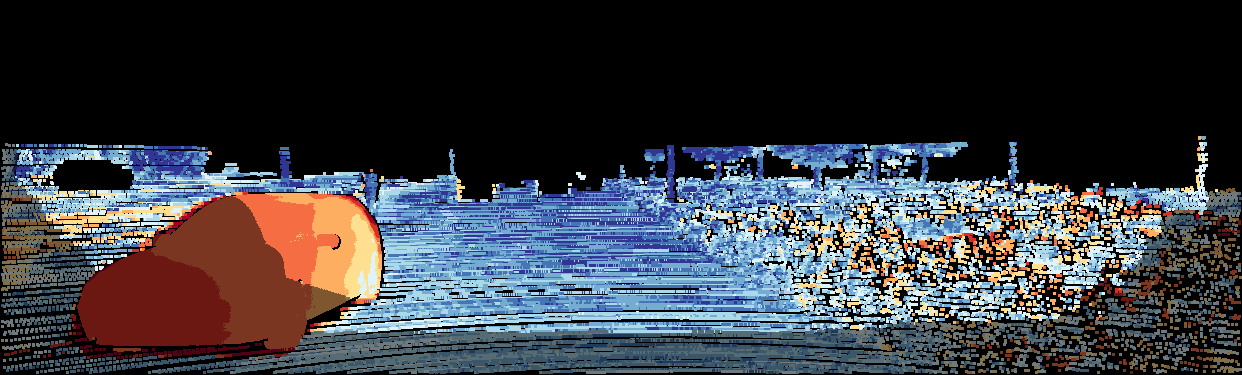}};
\draw (img.north west) node[labelstyle] {SF: 51.70};
\end{tikzpicture}
\\[-0.6mm]
\begin{tikzpicture}
\draw (0, 0) node[imgstyle] (img) {
\includegraphics[trim={0 0 0 2.85cm},clip,width=0.197\textwidth]{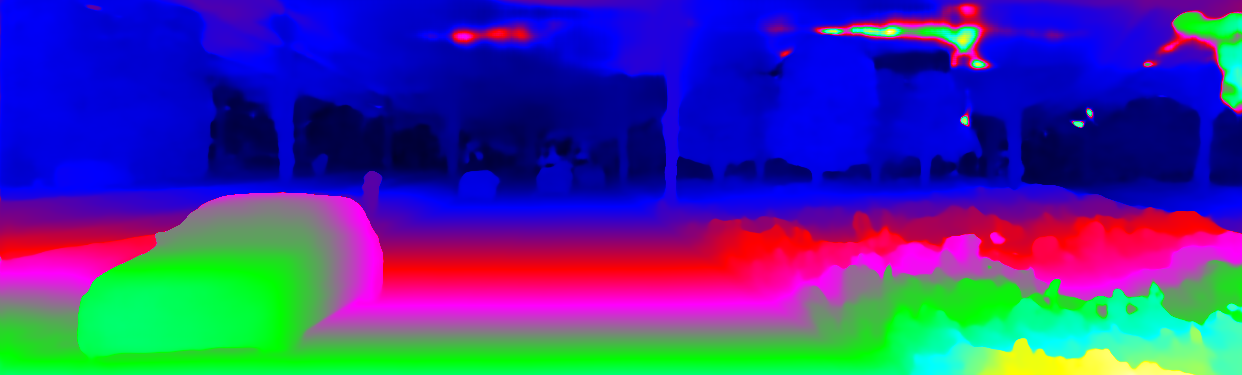}};
\draw (img.north west) node[labelstyle] {CamLiFlow};
\end{tikzpicture} &
\begin{tikzpicture}
\draw (0, 0) node[imgstyle] (img) {
\includegraphics[trim={0 0 0 2.85cm},clip,width=0.197\textwidth]{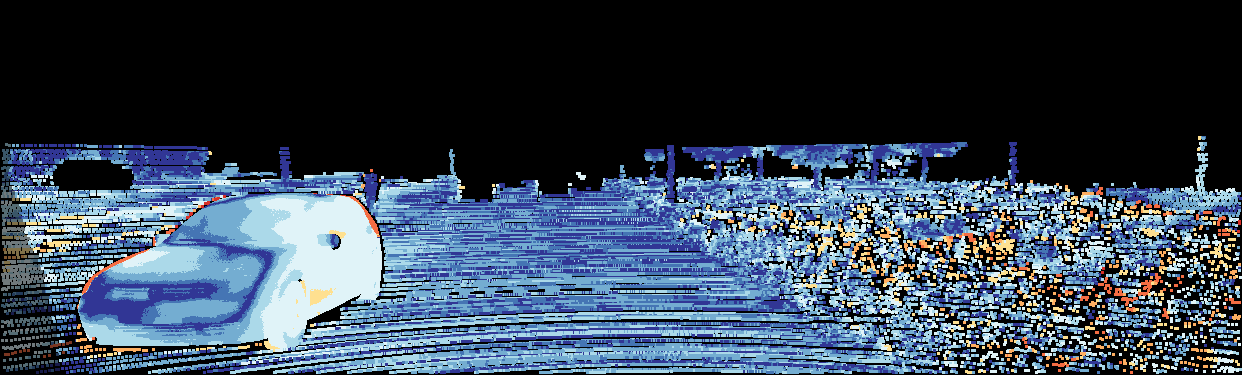}};
\draw (img.north west) node[labelstyle] {D2: 33.24};
\end{tikzpicture} &
\includegraphics[trim={0 0 0 2.85cm},clip,width=0.197\textwidth]{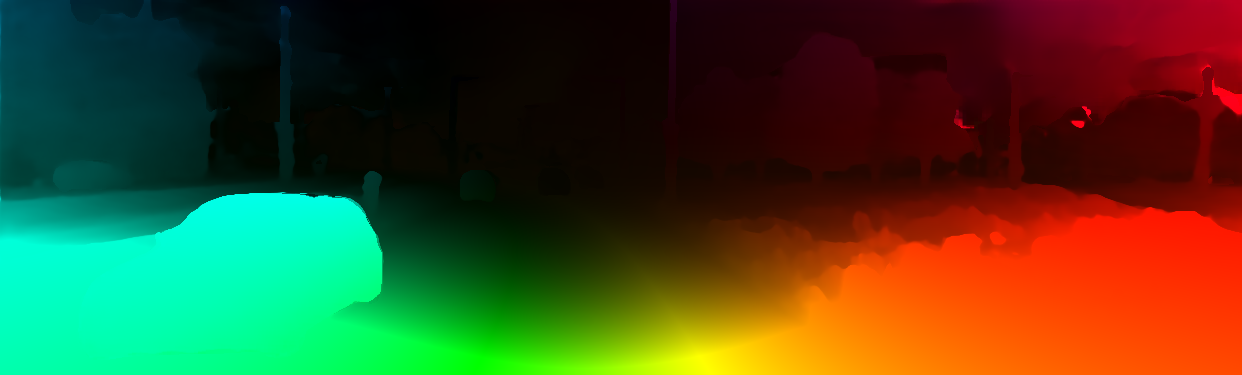} &
\begin{tikzpicture}
\draw (0, 0) node[imgstyle] (img) {
\includegraphics[trim={0 0 0 2.85cm},clip,width=0.197\textwidth]{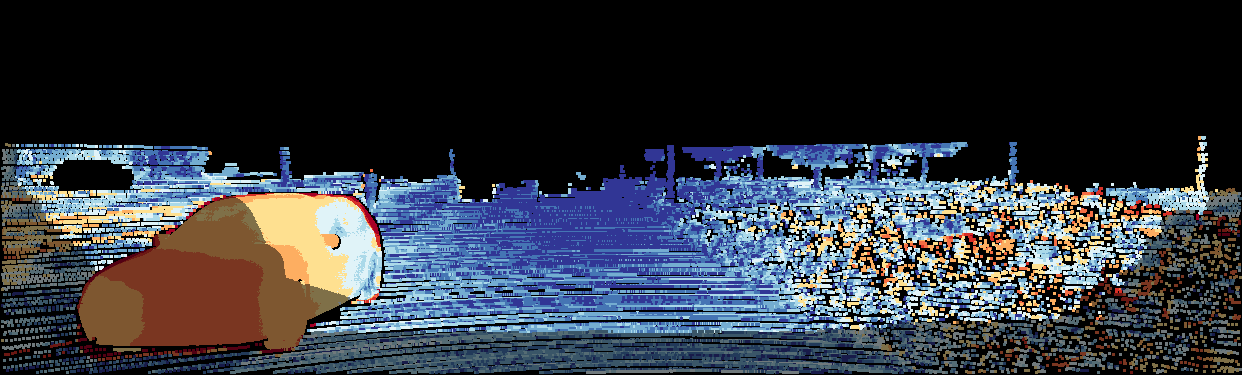}};
\draw (img.north west) node[labelstyle] {Fl: 46.64};
\end{tikzpicture} &
\begin{tikzpicture}
\draw (0, 0) node[imgstyle] (img) {
\includegraphics[trim={0 0 0 2.85cm},clip,width=0.197\textwidth]{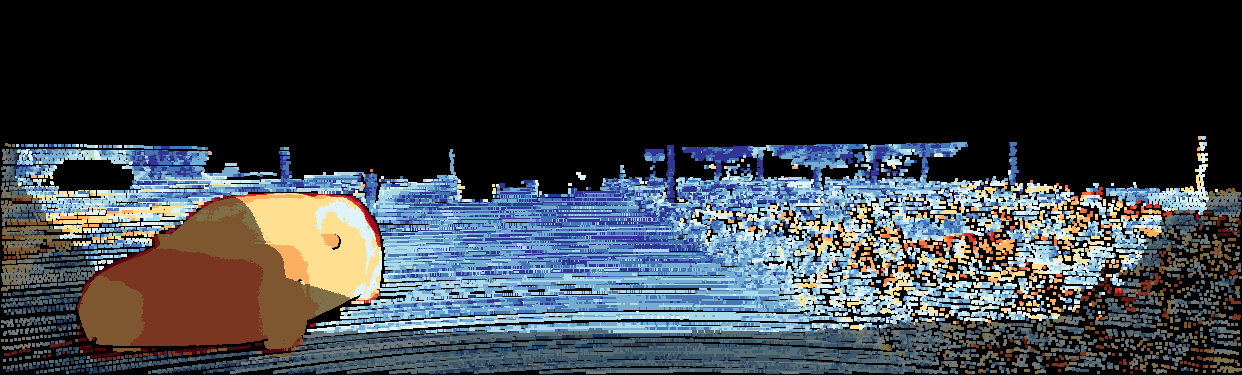}};
\draw (img.north west) node[labelstyle] {SF: 49.97};
\end{tikzpicture}
\\[-0.6mm]
\begin{tikzpicture}
\draw (0, 0) node[imgstyle] (img) {
\includegraphics[trim={0 0 0 2.85cm},clip,width=0.197\textwidth]{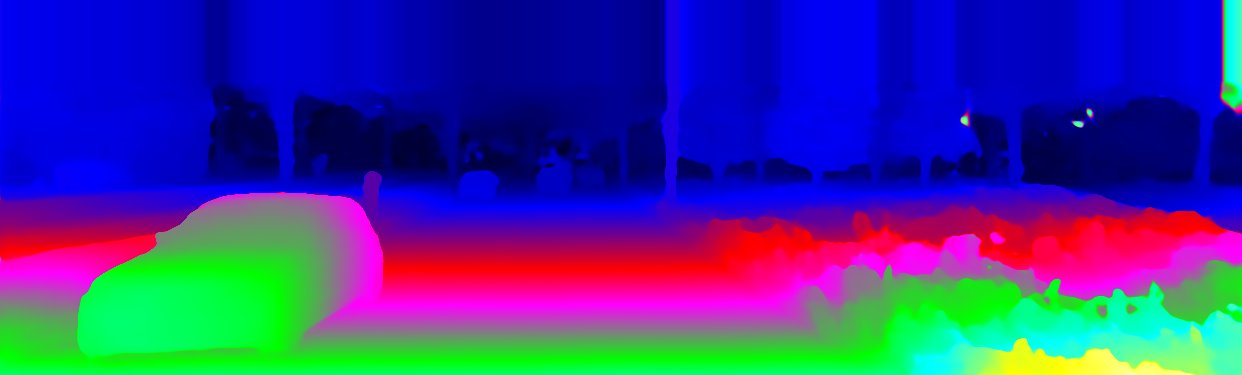}};
\draw (img.north west) node[labelstyle] {RAFT-3D};
\end{tikzpicture} &
\begin{tikzpicture}
\draw (0, 0) node[imgstyle] (img) {
\includegraphics[trim={0 0 0 2.85cm},clip,width=0.197\textwidth]{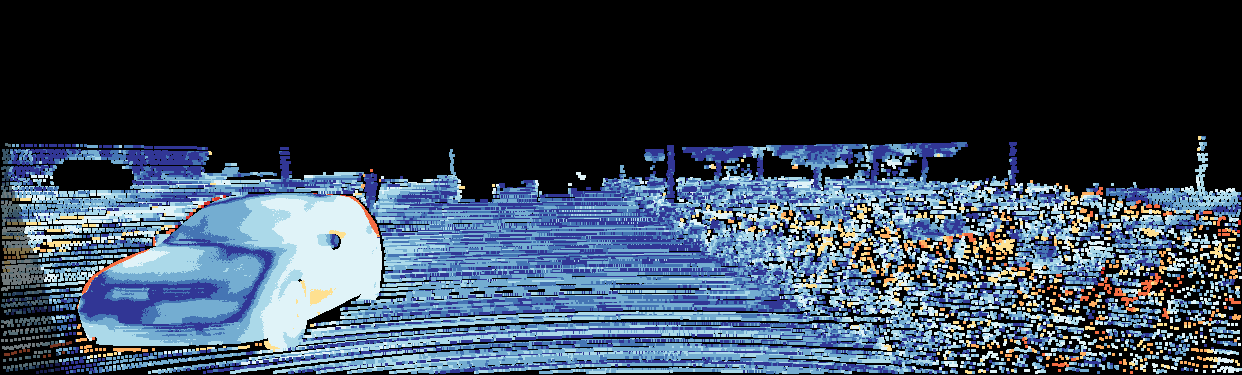}};
\draw (img.north west) node[labelstyle] {D2: 46.10};
\end{tikzpicture} &
\includegraphics[trim={0 0 0 2.85cm},clip,width=0.197\textwidth]{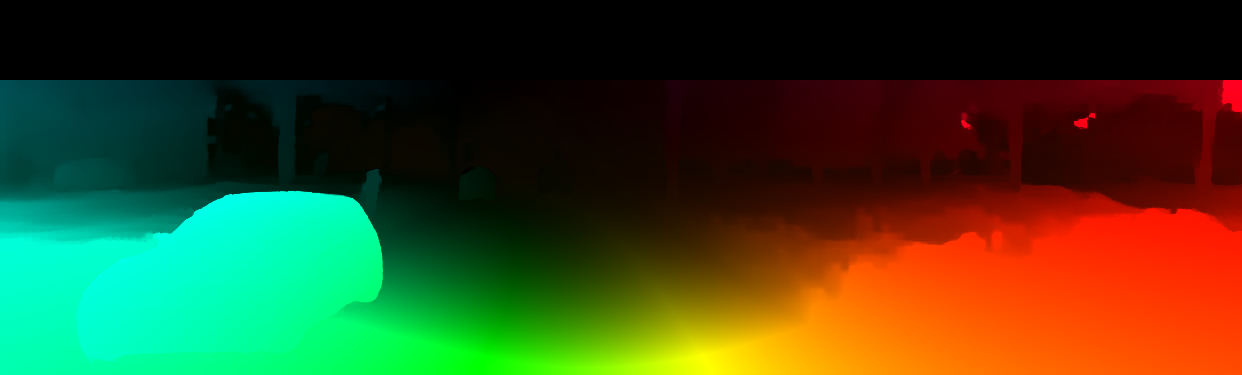} &
\begin{tikzpicture}
\draw (0, 0) node[imgstyle] (img) {
\includegraphics[trim={0 0 0 2.85cm},clip,width=0.197\textwidth]{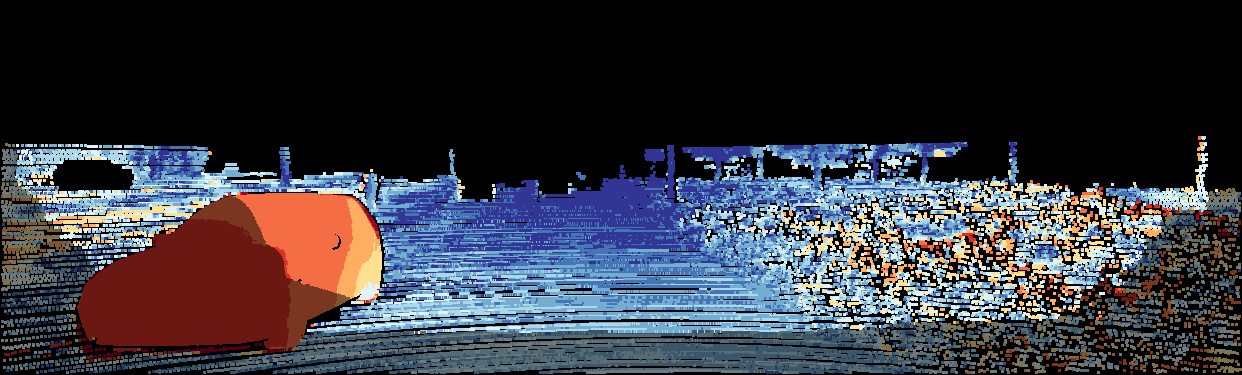}};
\draw (img.north west) node[labelstyle] {Fl: 51.76};
\end{tikzpicture} &
\begin{tikzpicture}
\draw (0, 0) node[imgstyle] (img) {
\includegraphics[trim={0 0 0 2.85cm},clip,width=0.197\textwidth]{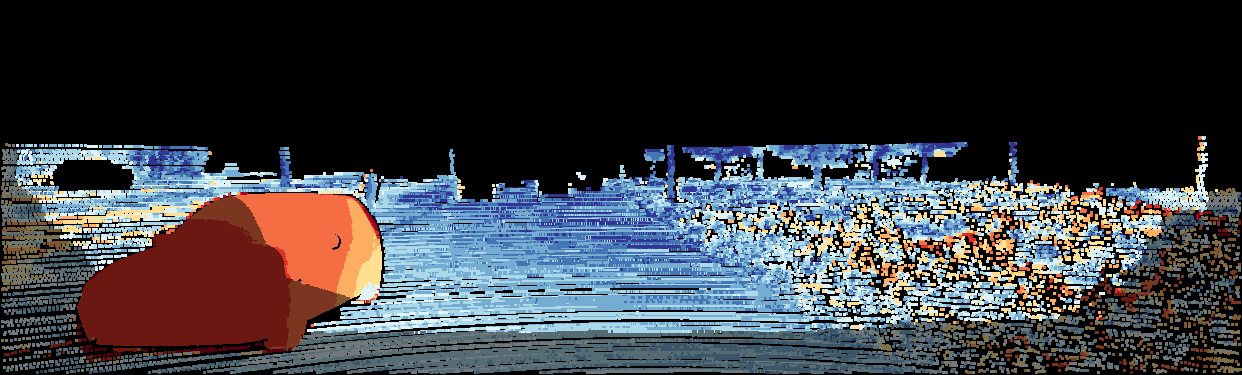}};
\draw (img.north west) node[labelstyle] {SF: 52.14};
\end{tikzpicture}
\\[-0.6mm]
\begin{tikzpicture}
\draw (0, 0) node[imgstyle] (img) {
\includegraphics[trim={0 0 0 2.85cm},clip,width=0.197\textwidth]{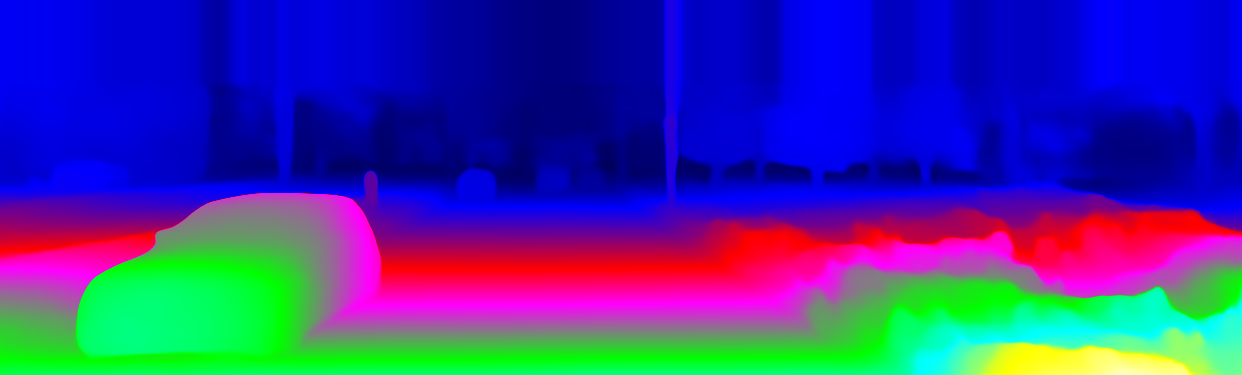}};
\draw (img.north west) node[labelstyle] {M-FUSE};
\end{tikzpicture} &
\begin{tikzpicture}
\draw (0, 0) node[imgstyle] (img) {
\includegraphics[trim={0 0 0 2.85cm},clip,width=0.197\textwidth]{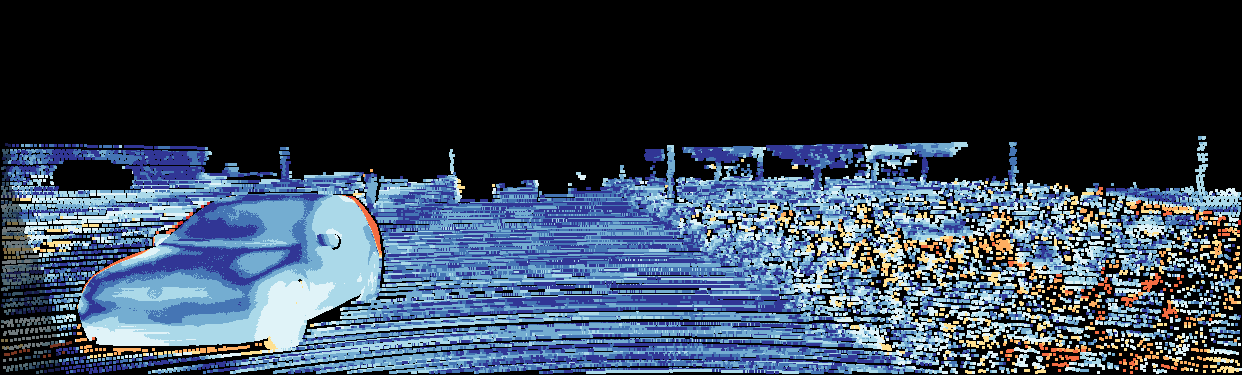}};
\draw (img.north west) node[labelstyle] {D2: 30.93};
\end{tikzpicture} &
\includegraphics[trim={0 0 0 2.85cm},clip,width=0.197\textwidth]{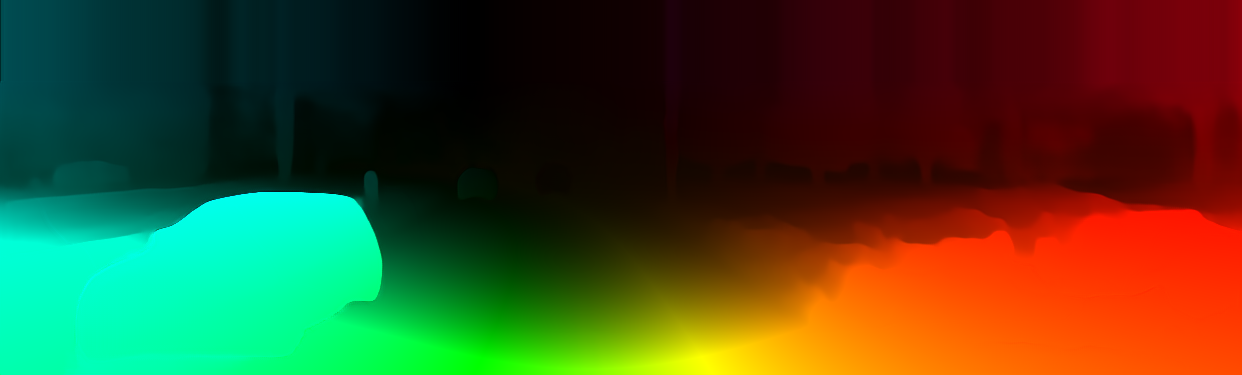} &
\begin{tikzpicture}
\draw (0, 0) node[imgstyle] (img) {
\includegraphics[trim={0 0 0 2.85cm},clip,width=0.197\textwidth]{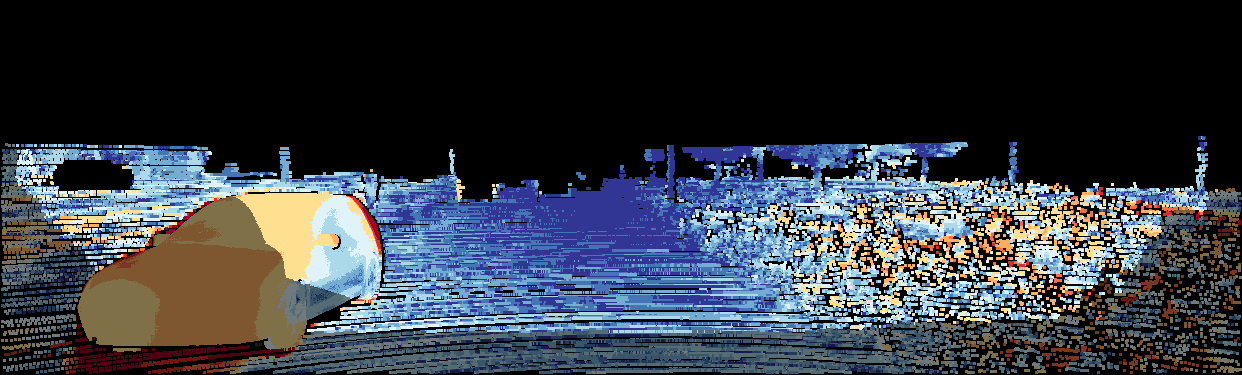}};
\draw (img.north west) node[labelstyle] {Fl: 42.31};
\end{tikzpicture} &
\begin{tikzpicture}
\draw (0, 0) node[imgstyle] (img) {
\includegraphics[trim={0 0 0 2.85cm},clip,width=0.197\textwidth]{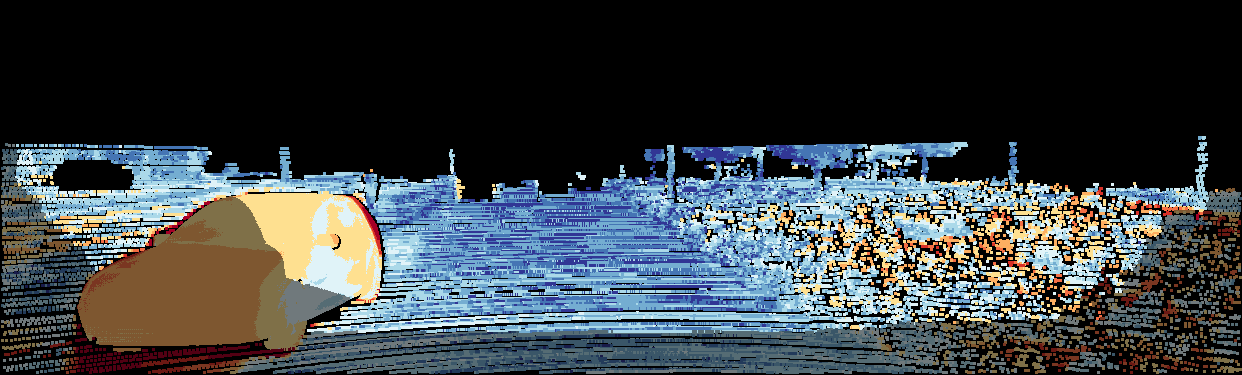}};
\draw (img.north west) node[labelstyle] {SF: 46.16};
\end{tikzpicture}
\\
\begin{tikzpicture}
\draw (0, 0) node[imgstyle] (img) {
\includegraphics[trim={0 0 0 2.85cm},clip,width=0.197\textwidth]{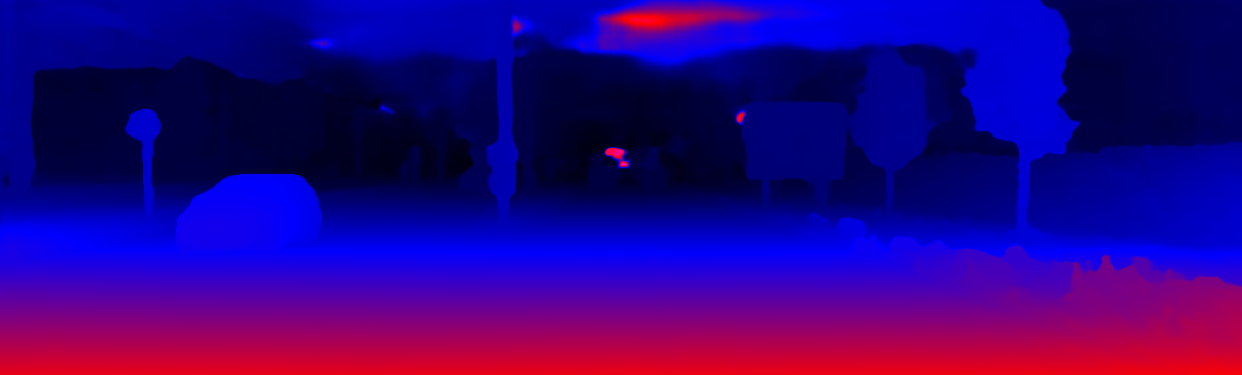}};
\draw (img.north west) node[labelstyle] {RigidMask+ISF};
\end{tikzpicture} &
\begin{tikzpicture}
\draw (0, 0) node[imgstyle] (img) {
\includegraphics[trim={0 0 0 2.85cm},clip,width=0.197\textwidth]{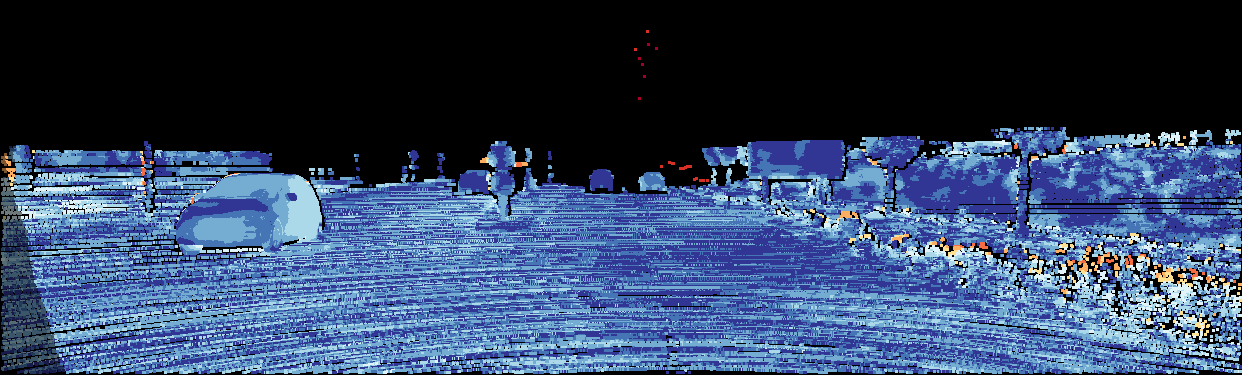}};
\draw (img.north west) node[labelstyle] {D2: 1.36};
\end{tikzpicture} &
\includegraphics[trim={0 0 0 2.85cm},clip,width=0.197\textwidth]{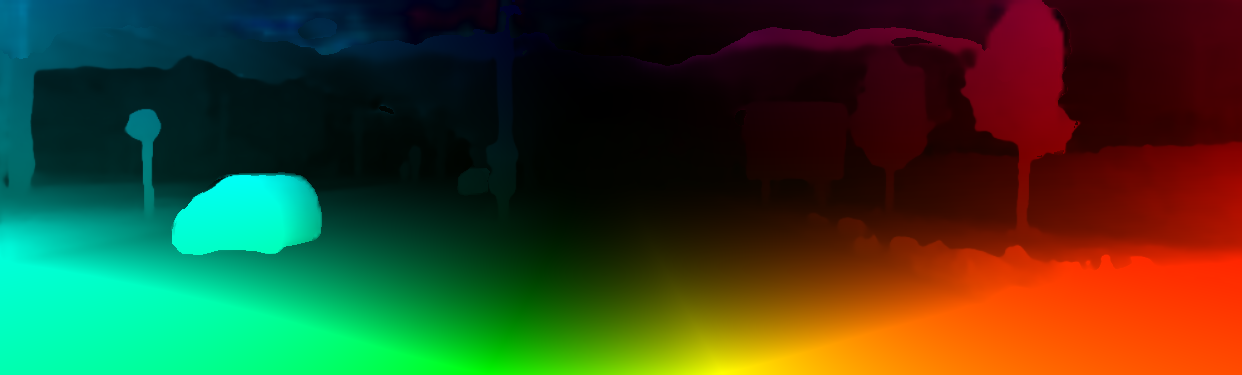} &
\begin{tikzpicture}
\draw (0, 0) node[imgstyle] (img) {
\includegraphics[trim={0 0 0 2.85cm},clip,width=0.197\textwidth]{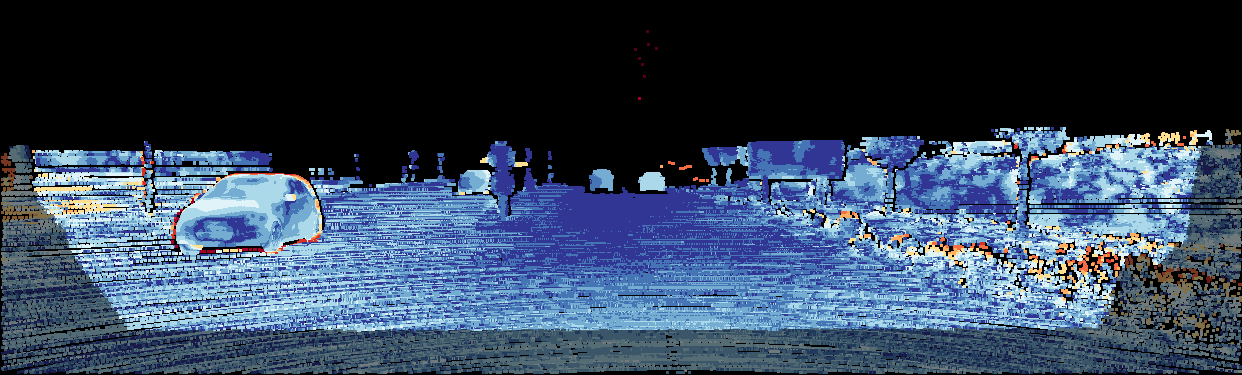}};
\draw (img.north west) node[labelstyle] {Fl: 1.94};
\end{tikzpicture} &
\begin{tikzpicture}
\draw (0, 0) node[imgstyle] (img) {
\includegraphics[trim={0 0 0 2.85cm},clip,width=0.197\textwidth]{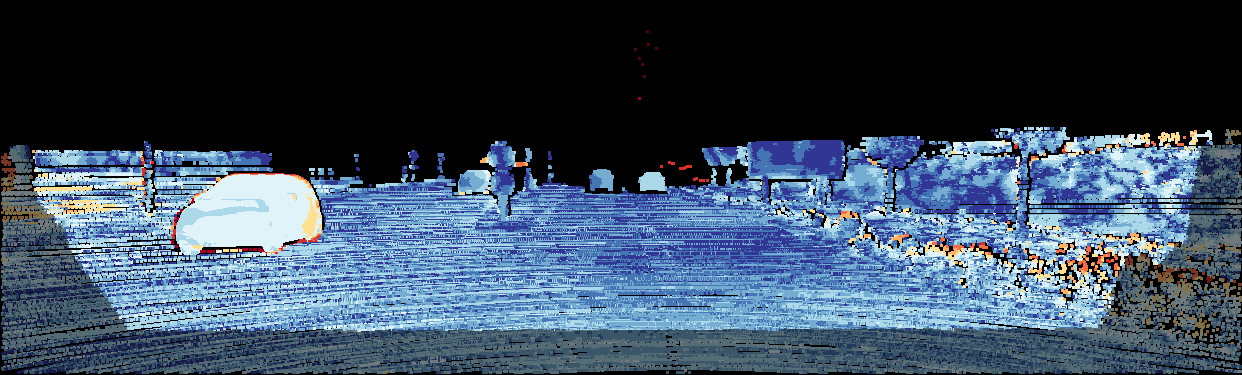}};
\draw (img.north west) node[labelstyle] {SF: 2.34};
\end{tikzpicture}
\\[-0.6mm]
\begin{tikzpicture}
\draw (0, 0) node[imgstyle] (img) {
\includegraphics[trim={0 0 0 2.85cm},clip,width=0.197\textwidth]{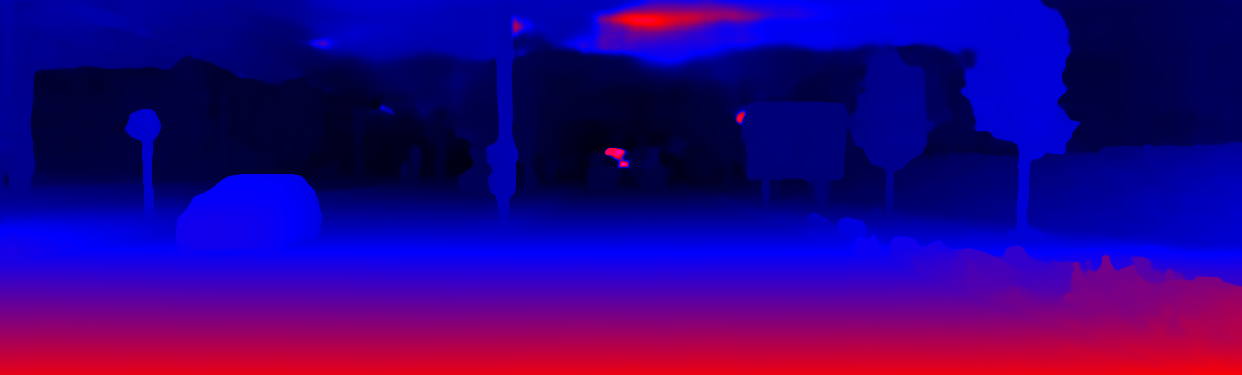}};
\draw (img.north west) node[labelstyle] {CamLiFlow};
\end{tikzpicture} &
\begin{tikzpicture}
\draw (0, 0) node[imgstyle] (img) {
\includegraphics[trim={0 0 0 2.85cm},clip,width=0.197\textwidth]{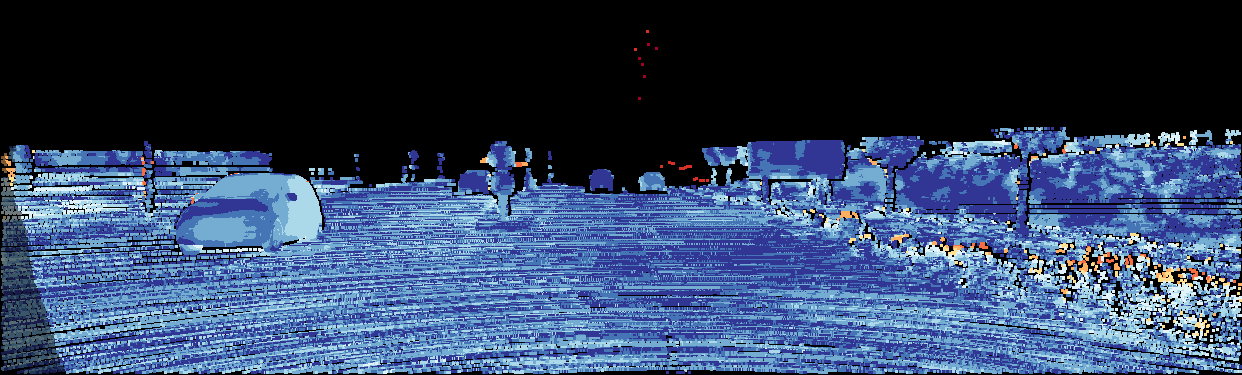}};
\draw (img.north west) node[labelstyle] {D2: 0.98};
\end{tikzpicture} &
\includegraphics[trim={0 0 0 2.85cm},clip,width=0.197\textwidth]{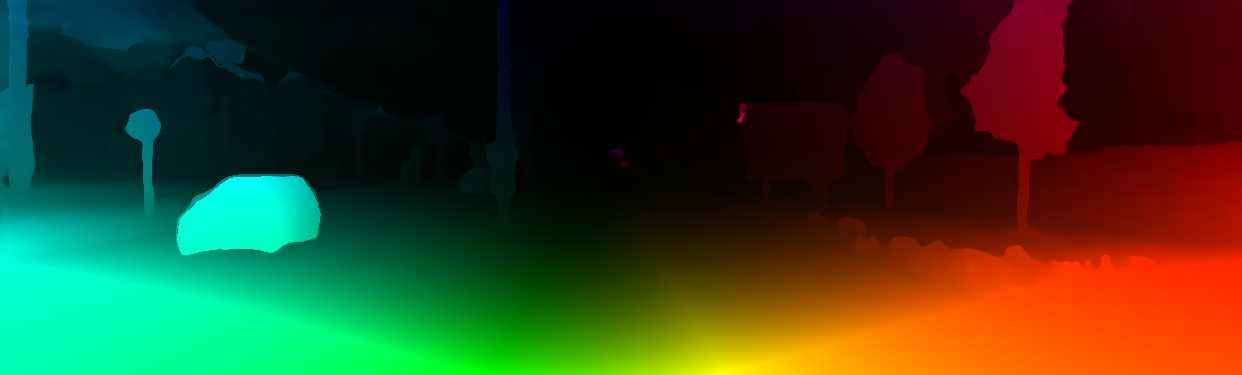} &
\begin{tikzpicture}
\draw (0, 0) node[imgstyle] (img) {
\includegraphics[trim={0 0 0 2.85cm},clip,width=0.197\textwidth]{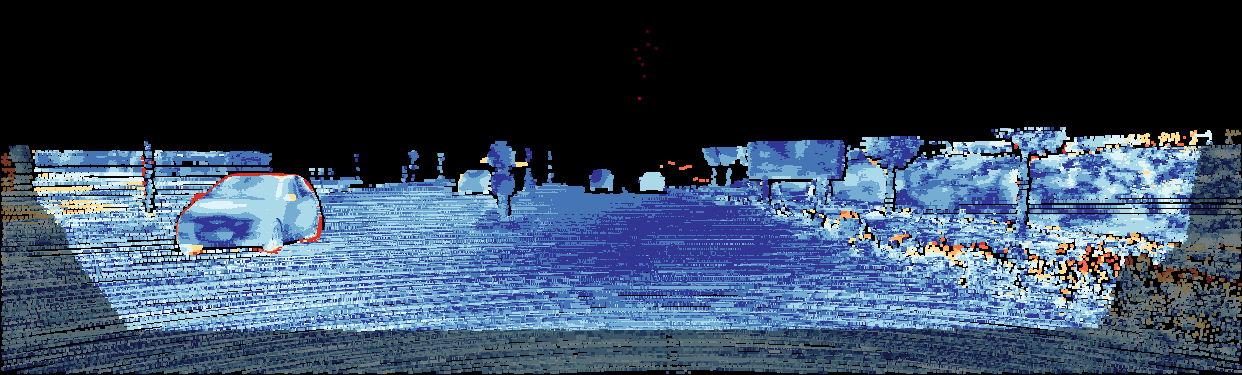}};
\draw (img.north west) node[labelstyle] {Fl: 2.24};
\end{tikzpicture} &
\begin{tikzpicture}
\draw (0, 0) node[imgstyle] (img) {
\includegraphics[trim={0 0 0 2.85cm},clip,width=0.197\textwidth]{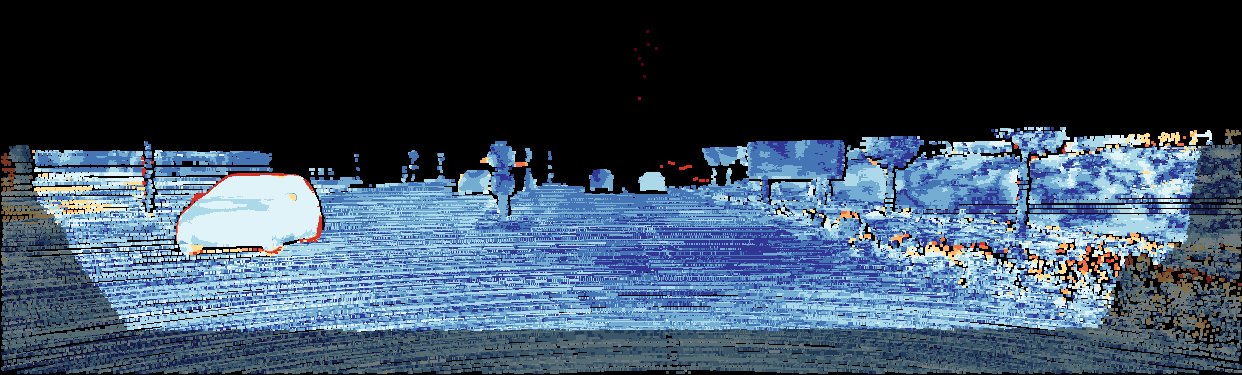}};
\draw (img.north west) node[labelstyle] {SF: 2.29};
\end{tikzpicture}
\\[-0.6mm]
\begin{tikzpicture}
\draw (0, 0) node[imgstyle] (img) {
\includegraphics[trim={0 0 0 2.85cm},clip,width=0.197\textwidth]{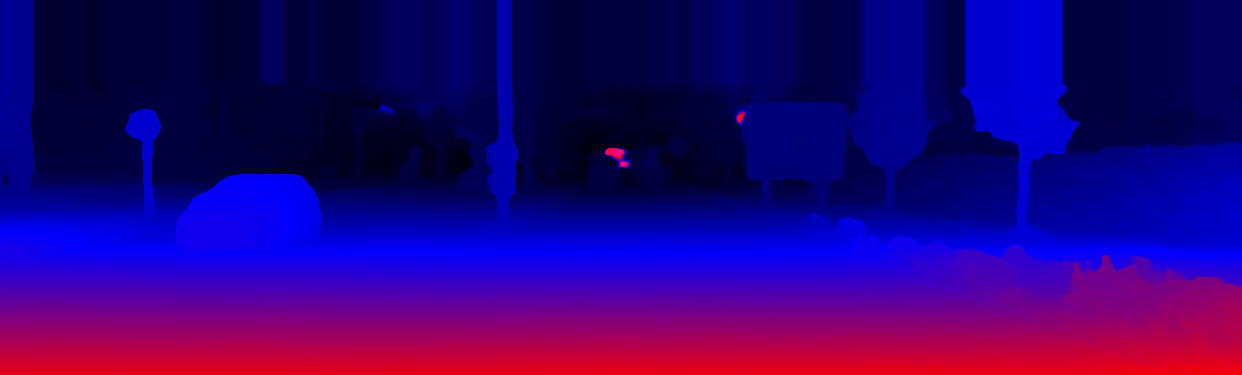}};
\draw (img.north west) node[labelstyle] {RAFT-3D};
\end{tikzpicture} &
\begin{tikzpicture}
\draw (0, 0) node[imgstyle] (img) {
\includegraphics[trim={0 0 0 2.85cm},clip,width=0.197\textwidth]{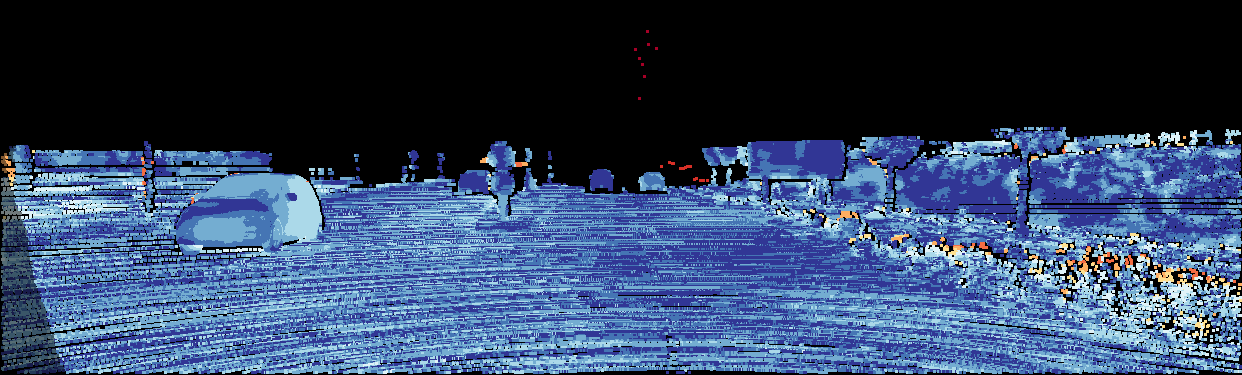}};
\draw (img.north west) node[labelstyle] {D2: 1.31};
\end{tikzpicture} &
\includegraphics[trim={0 0 0 2.85cm},clip,width=0.197\textwidth]{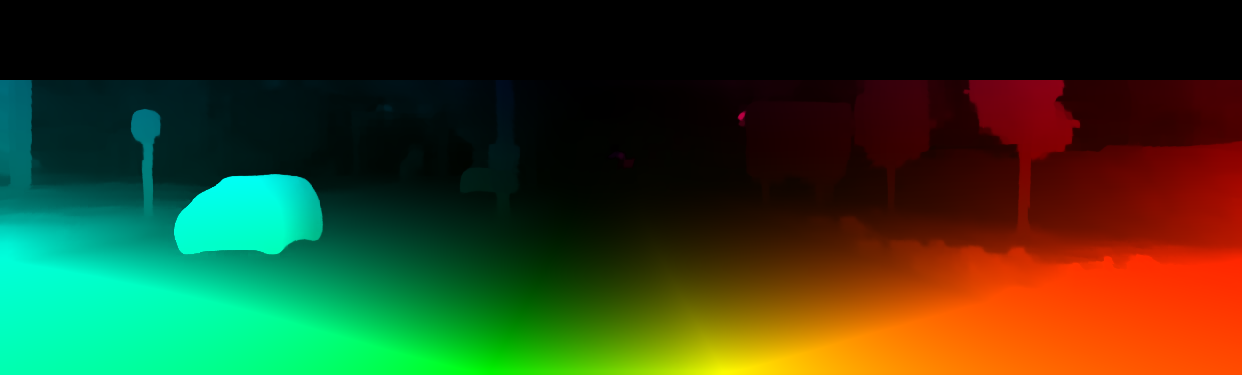} &
\begin{tikzpicture}
\draw (0, 0) node[imgstyle] (img) {
\includegraphics[trim={0 0 0 2.85cm},clip,width=0.197\textwidth]{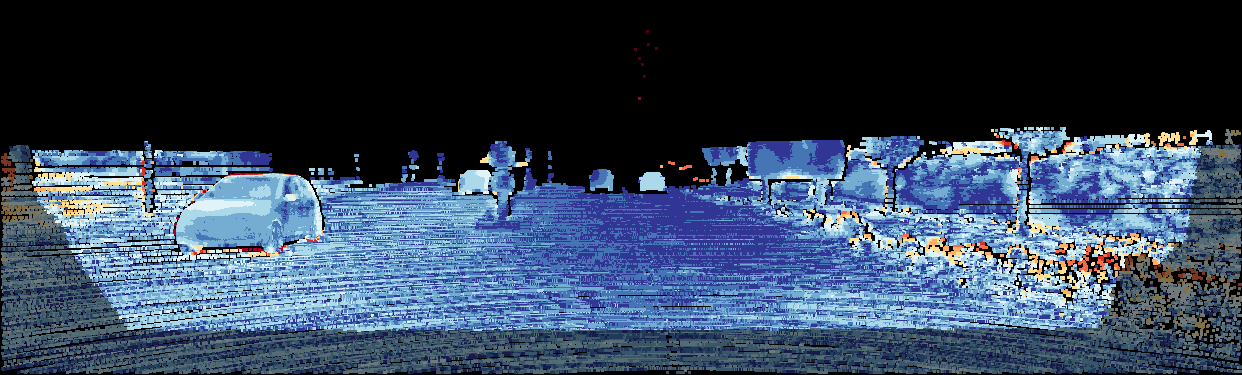}};
\draw (img.north west) node[labelstyle] {Fl: 1.97};
\end{tikzpicture} &
\begin{tikzpicture}
\draw (0, 0) node[imgstyle] (img) {
\includegraphics[trim={0 0 0 2.85cm},clip,width=0.197\textwidth]{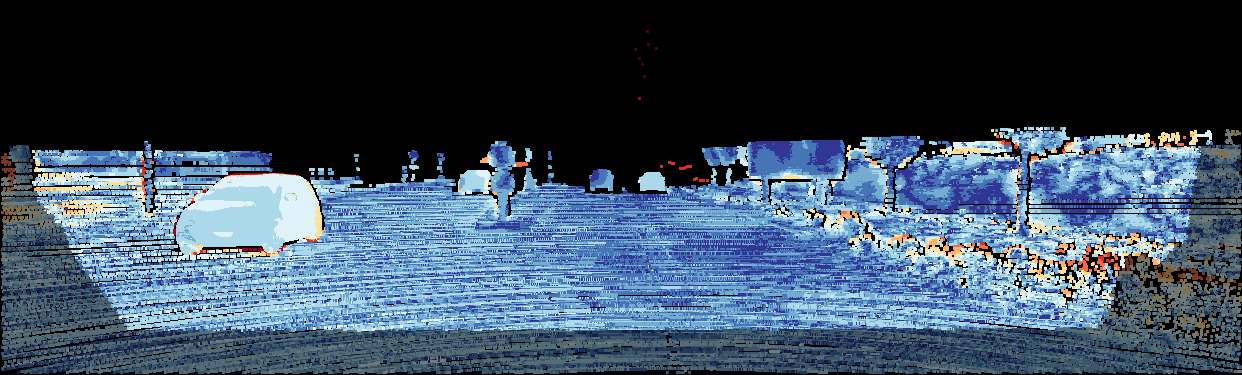}};
\draw (img.north west) node[labelstyle] {SF: 2.21};
\end{tikzpicture}
\\[-0.6mm]
\begin{tikzpicture}
\draw (0, 0) node[imgstyle] (img) {
\includegraphics[trim={0 0 0 2.85cm},clip,width=0.197\textwidth]{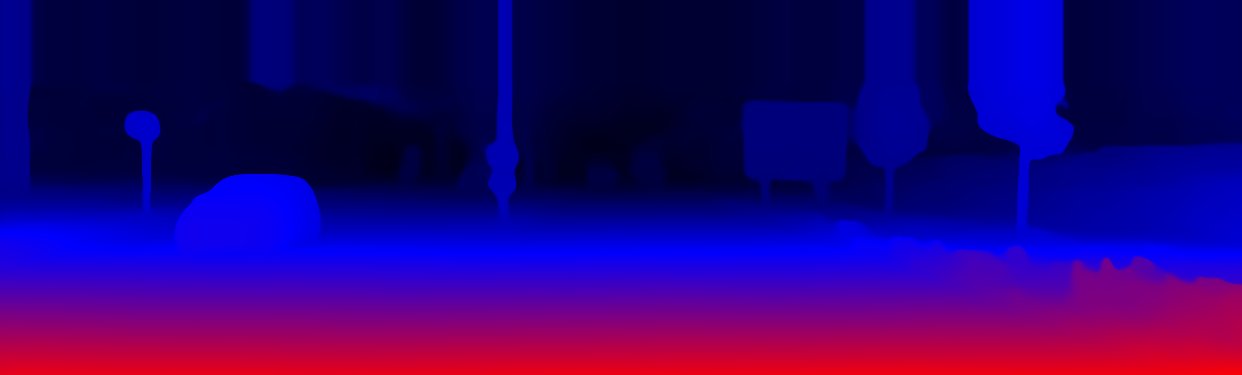}};
\draw (img.north west) node[labelstyle] {M-FUSE};
\end{tikzpicture} &
\begin{tikzpicture}
\draw (0, 0) node[imgstyle] (img) {
\includegraphics[trim={0 0 0 2.85cm},clip,width=0.197\textwidth]{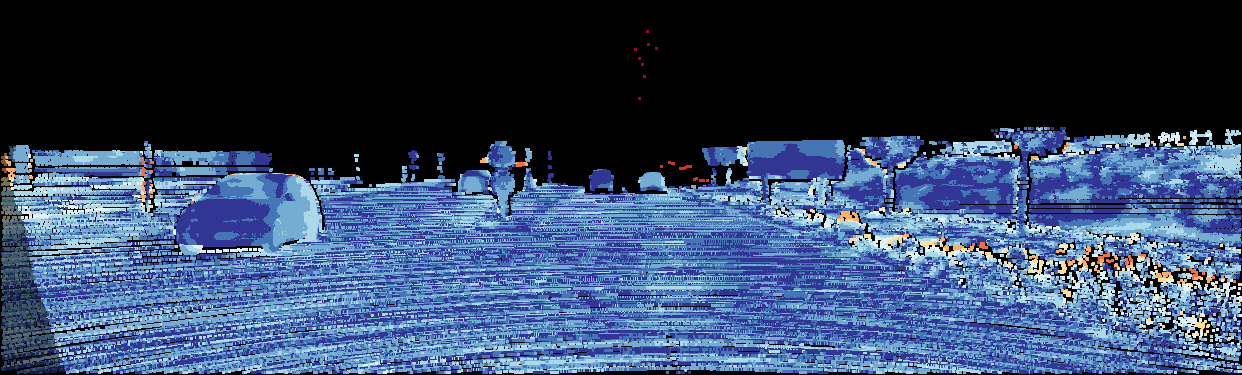}};
\draw (img.north west) node[labelstyle] {D2: 1.11};
\end{tikzpicture} &
\includegraphics[trim={0 0 0 2.85cm},clip,width=0.197\textwidth]{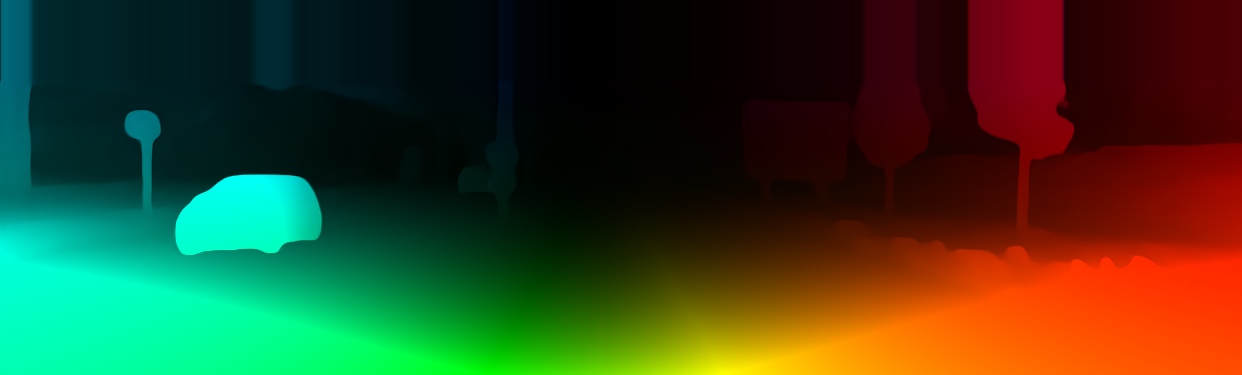} &
\begin{tikzpicture}
\draw (0, 0) node[imgstyle] (img) {
\includegraphics[trim={0 0 0 2.85cm},clip,width=0.197\textwidth]{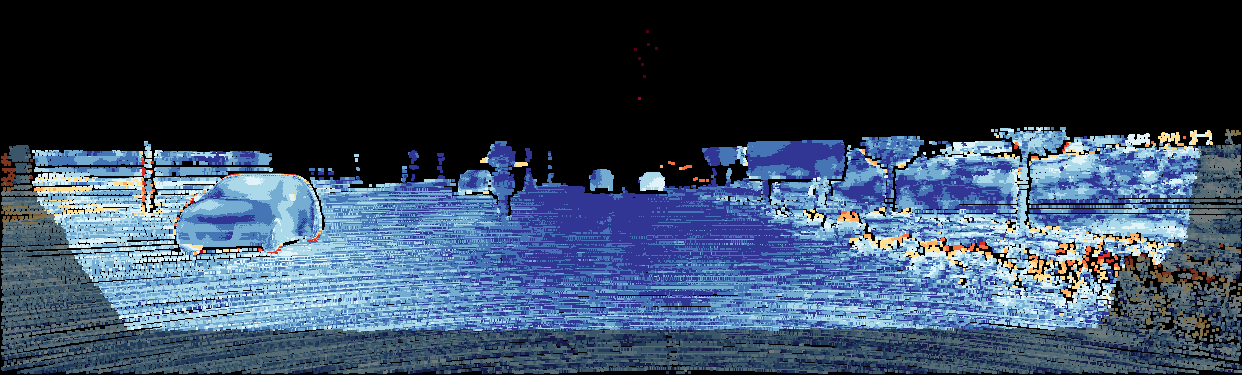}};
\draw (img.north west) node[labelstyle] {Fl: 1.91};
\end{tikzpicture} &
\begin{tikzpicture}
\draw (0, 0) node[imgstyle] (img) {
\includegraphics[trim={0 0 0 2.85cm},clip,width=0.197\textwidth]{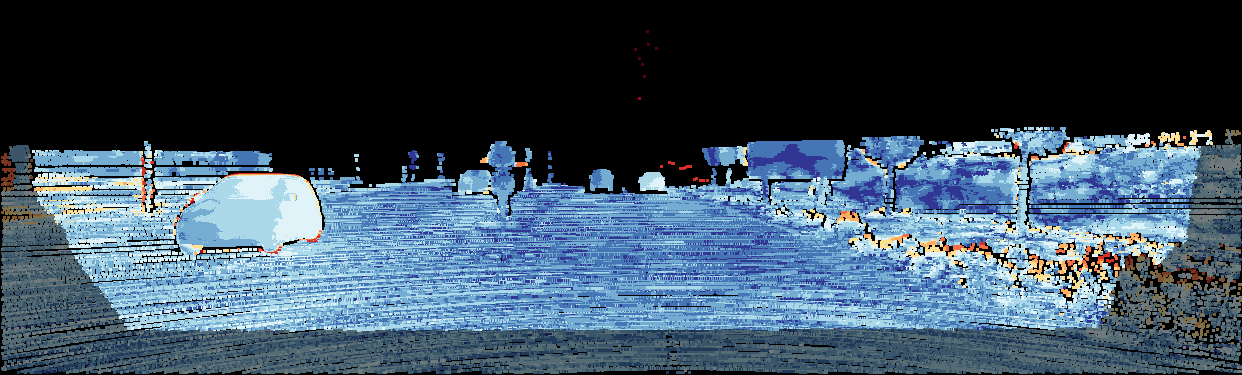}};
\draw (img.north west) node[labelstyle] {SF: 2.04};
\end{tikzpicture}
\end{tabular}
}
\caption{Qualitative comparison of our method, the original RAFT-3D, as well as the two top-performing approaches from the literature using the visualizations provided by the KITTI benchmark. \emph{From left to right:} Target disparity visualization, corresponding \emph{D2} error plot, optical flow visualization, corresponding \emph{Fl} error plot, combined \emph{SF} error plot.}
\label{fig:end}
\end{figure*}

We show additional visual results from the KITTI benchmark in Figures~\ref{fig:begin}--\ref{fig:end}.

\begin{figure*}
\centering{
\tikzset{labelstyle/.style={anchor=north west, text=white, inner sep=2, text opacity=1, scale=0.7, yshift=-1, xshift=1, fill=black, opacity=0.6}}
\tikzset{imgstyle/.style={inner sep=0,anchor=north west,outer sep=0,draw=none,line width=0}}
\setlength\tabcolsep{1pt}
\begin{tabular}{ccccc}
\begin{tikzpicture}
\draw (0, 0) node[imgstyle] (img) {
\includegraphics[trim={0 0 0 2.85cm},clip,width=0.197\textwidth]{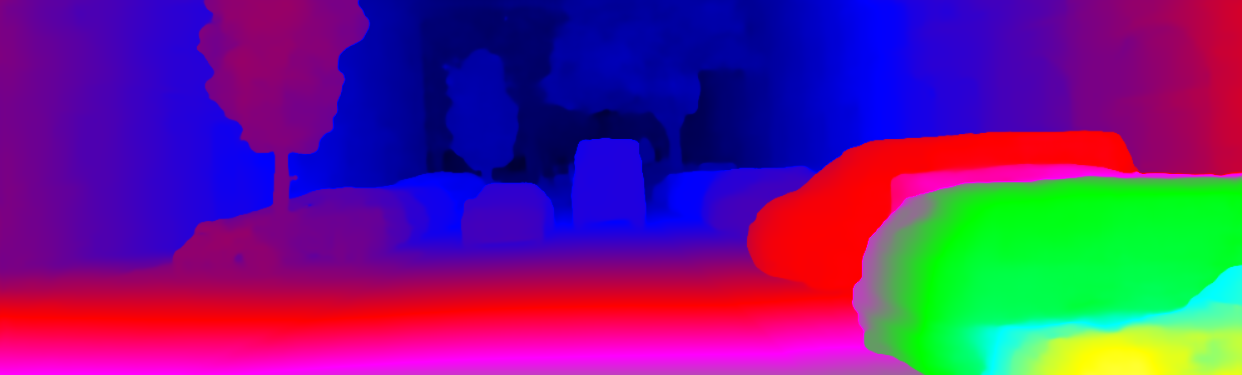}};
\draw (img.north west) node[labelstyle] {RigidMask+ISF};
\end{tikzpicture} &
\begin{tikzpicture}
\draw (0, 0) node[imgstyle] (img) {
\includegraphics[trim={0 0 0 2.85cm},clip,width=0.197\textwidth]{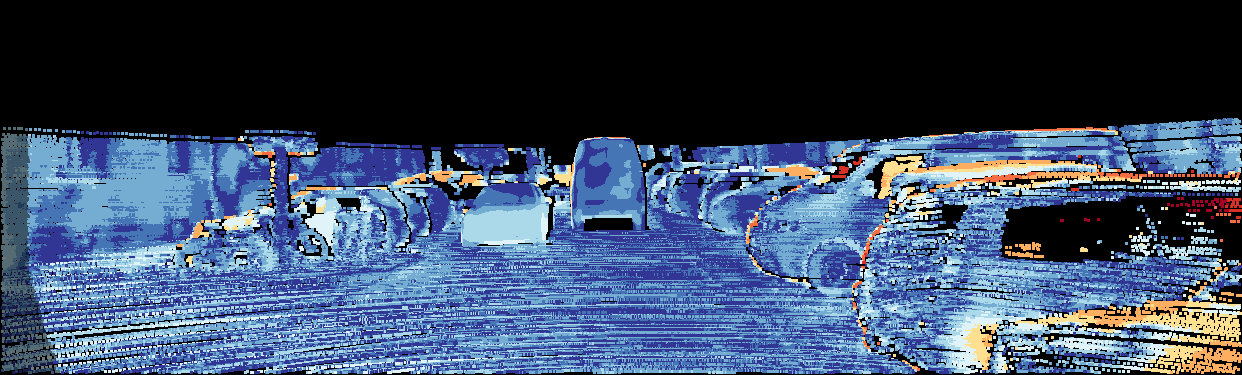}};
\draw (img.north west) node[labelstyle] {D2: 3.96};
\end{tikzpicture} &
\includegraphics[trim={0 0 0 2.85cm},clip,width=0.197\textwidth]{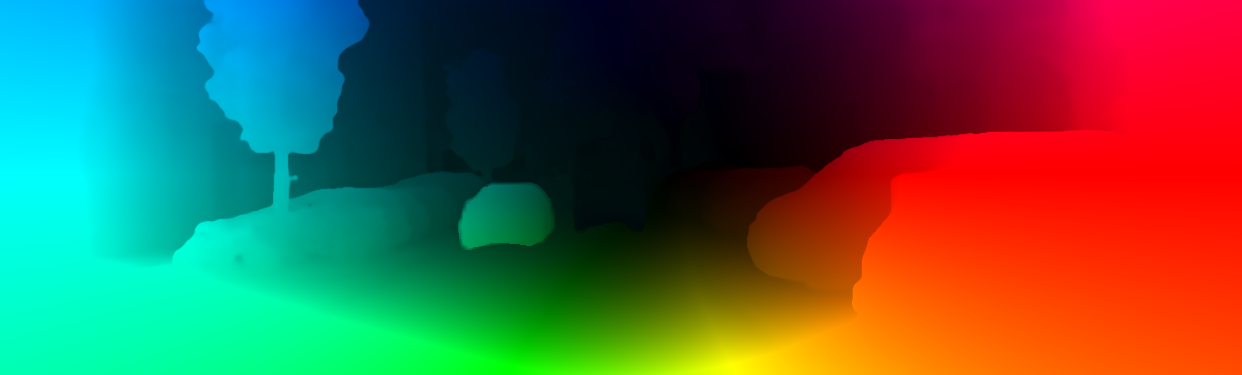} &
\begin{tikzpicture}
\draw (0, 0) node[imgstyle] (img) {
\includegraphics[trim={0 0 0 2.85cm},clip,width=0.197\textwidth]{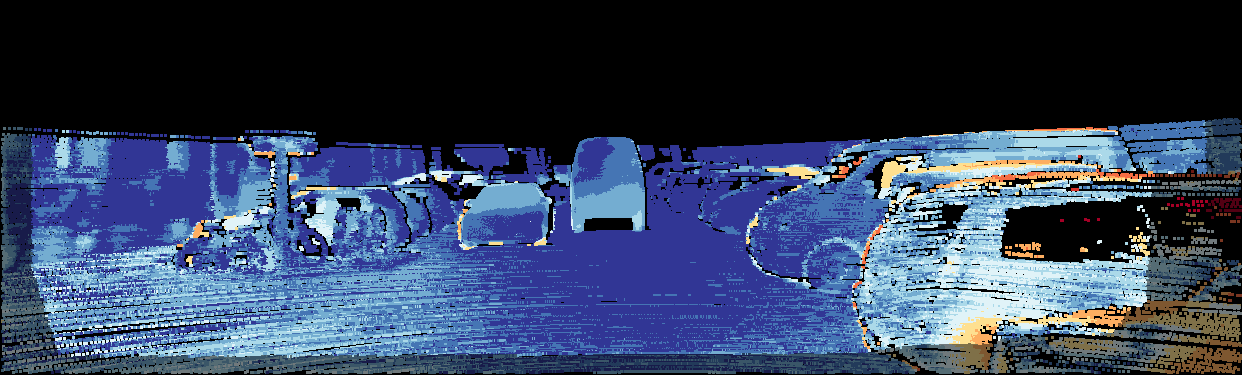}};
\draw (img.north west) node[labelstyle] {Fl: 3.24};
\end{tikzpicture} &
\begin{tikzpicture}
\draw (0, 0) node[imgstyle] (img) {
\includegraphics[trim={0 0 0 2.85cm},clip,width=0.197\textwidth]{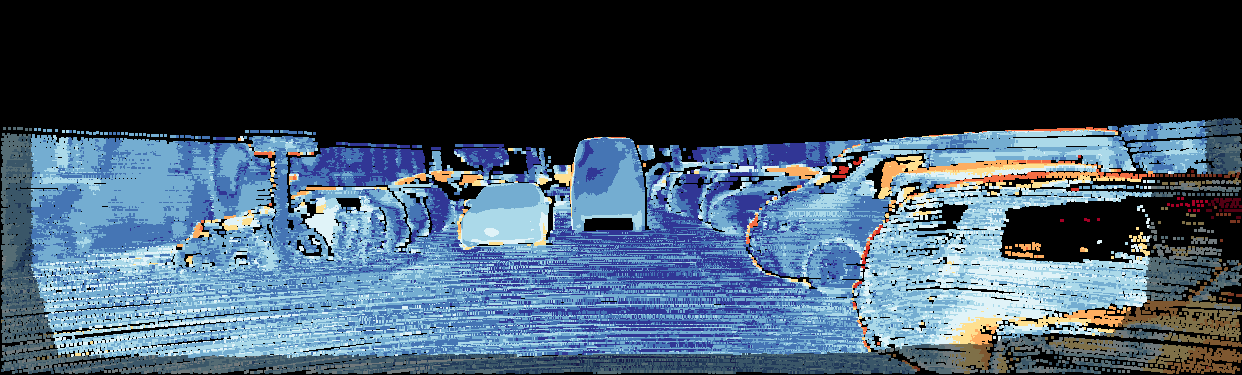}};
\draw (img.north west) node[labelstyle] {SF: 4.41};
\end{tikzpicture}
\\
\end{tabular}
}
\caption{Qualitative comparison of our method, the original RAFT-3D, as well as the two top-performing approaches from the literature using the visualizations provided by the KITTI benchmark. \emph{From left to right:} Target disparity visualization, corresponding \emph{D2} error plot, optical flow visualization, corresponding \emph{Fl} error plot, combined \emph{SF} error plot.}
\end{figure*}

\begin{figure*}
\centering{
\tikzset{labelstyle/.style={anchor=north west, text=white, inner sep=2, text opacity=1, scale=0.7, yshift=-1, xshift=1, fill=black, opacity=0.6}}
\tikzset{imgstyle/.style={inner sep=0,anchor=north west,outer sep=0,draw=none,line width=0}}
\setlength\tabcolsep{1pt}
\begin{tabular}{ccccc}
\begin{tikzpicture}
\draw (0, 0) node[imgstyle] (img) {
\includegraphics[trim={0 0 0 2.85cm},clip,width=0.197\textwidth]{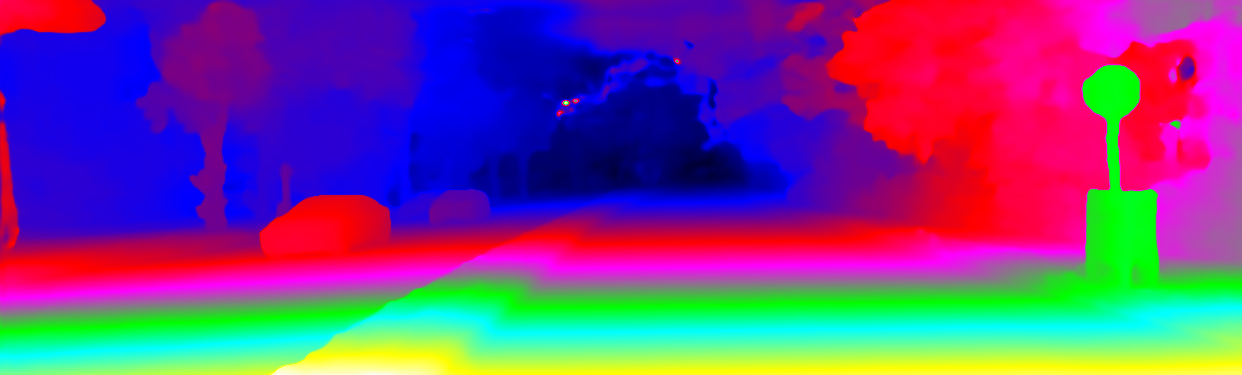}};
\draw (img.north west) node[labelstyle] {RigidMask+ISF};
\end{tikzpicture} &
\begin{tikzpicture}
\draw (0, 0) node[imgstyle] (img) {
\includegraphics[trim={0 0 0 2.85cm},clip,width=0.197\textwidth]{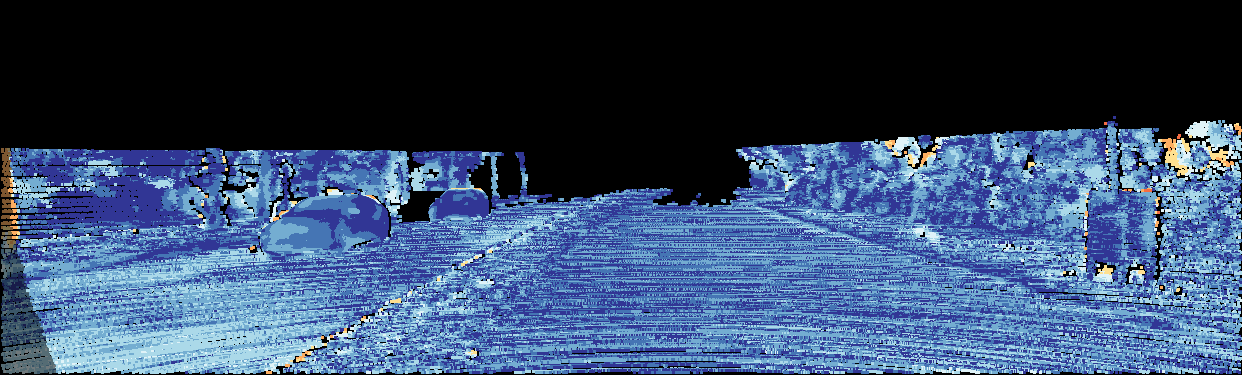}};
\draw (img.north west) node[labelstyle] {D2: 0.68};
\end{tikzpicture} &
\includegraphics[trim={0 0 0 2.85cm},clip,width=0.197\textwidth]{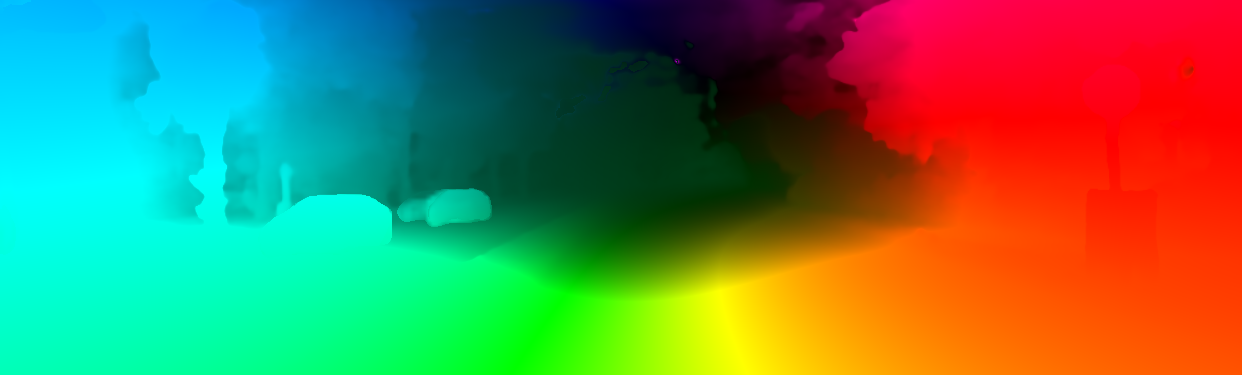} &
\begin{tikzpicture}
\draw (0, 0) node[imgstyle] (img) {
\includegraphics[trim={0 0 0 2.85cm},clip,width=0.197\textwidth]{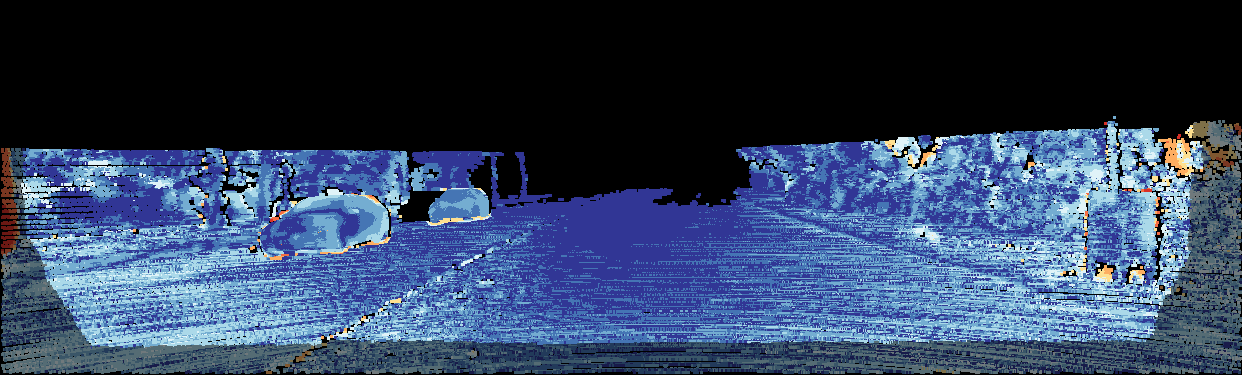}};
\draw (img.north west) node[labelstyle] {Fl: 1.02};
\end{tikzpicture} &
\begin{tikzpicture}
\draw (0, 0) node[imgstyle] (img) {
\includegraphics[trim={0 0 0 2.85cm},clip,width=0.197\textwidth]{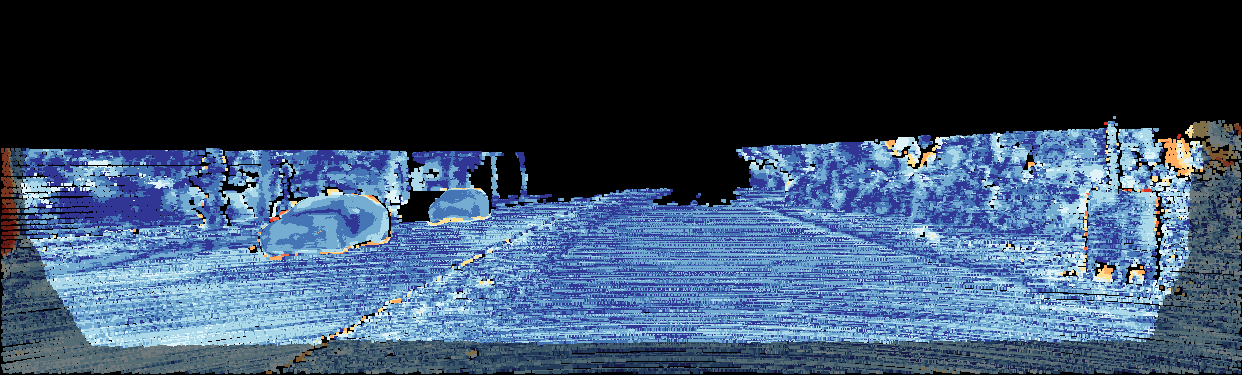}};
\draw (img.north west) node[labelstyle] {SF: 1.10};
\end{tikzpicture}
\\
\end{tabular}
}
\caption{Qualitative comparison of our method, the original RAFT-3D, as well as the two top-performing approaches from the literature using the visualizations provided by the KITTI benchmark. \emph{From left to right:} Target disparity visualization, corresponding \emph{D2} error plot, optical flow visualization, corresponding \emph{Fl} error plot, combined \emph{SF} error plot.}
\end{figure*}

\end{document}